\documentclass[lettersize,journal]{IEEEtran}
\usepackage{amsmath,amsfonts}
\usepackage{algorithmic}
\usepackage{array}
\usepackage{textcomp}
\usepackage{stfloats}
\usepackage{url}
\usepackage{verbatim}
\usepackage{graphicx}
\usepackage{color}
\usepackage{subfig}
\usepackage{lipsum}
\usepackage{float}
\usepackage{siunitx}
\usepackage{amssymb}
\usepackage{cite}
\usepackage{multirow}
\usepackage{xcolor}
\usepackage{hyperref}
\usepackage[ruled, lined, linesnumbered, commentsnumbered, longend]{algorithm2e}

\def\BibTeX{{\rm B\kern-.05em{\sc i\kern-.025em b}\kern-.08em
    T\kern-.1667em\lower.7ex\hbox{E}\kern-.125emX}}
\usepackage{balance}

\begin{document}

\twocolumn

\title{Investigation of Hierarchical Spectral Vision Transformer Architecture for Classification of Hyperspectral Imagery}

\author{Wei Liu, \emph{Member, IEEE}, Saurabh Prasad, \emph{Senior Member, IEEE}, Melba Crawford, \emph{Life Fellow, IEEE}
% <-this % stops a space
\thanks{© 2024 IEEE. Personal use of this material is permitted. Permission from IEEE must be obtained for all other uses, in any current or future media, including reprinting/republishing this material for advertising or promotional purposes, creating new collective works, for resale or redistribution to servers or lists, or reuse of any copyrighted component of this work in other works.}

\thanks{Wei Liu is with the School of Electrical and Computer Engineering, Purdue University, West Lafayette, IN 47907 USA (e-mail: liu3044@purdue.edu).

Saurabh Prasad is with the Department of Electrical and Computer Engineering, University of Houston, Houston, TX 77204-4005 USA (e-mail: saurabh.prasad@ieee.org).

Melba Crawford is with the Lyles School of Civil Engineering and School of Electrical and Computer Engineering, Purdue University, West Lafayette, IN 47907 USA (e-mail: mcrawford@purdue.edu).}}

% \markboth{Journal of \LaTeX\ Class Files,~Vol.~18, No.~9, September~2020}%
% {How to Use the IEEEtran \LaTeX \ Templates}

\maketitle

\begin{abstract}

In the past three years, there has been significant interest in hyperspectral imagery (HSI) classification using vision Transformers for analysis of remotely sensed data. Previous research predominantly focused on the empirical integration of convolutional neural networks (CNNs) to augment the network's capability to extract local feature information. Yet, the theoretical justification for vision Transformers out-performing CNN architectures in HSI classification remains a question. To address this issue, a unified hierarchical spectral vision Transformer architecture, specifically tailored for HSI classification, is investigated. In this streamlined yet effective vision Transformer architecture, multiple mixer modules are strategically integrated separately. These include the CNN-mixer, which executes convolution operations; the spatial self-attention (SSA)-mixer and channel self-attention (CSA)-mixer, both of which are adaptations of classical self-attention blocks; and hybrid models such as the SSA+CNN-mixer and CSA+CNN-mixer, which merge convolution with self-attention operations. This integration facilitates the development of a broad spectrum of vision Transformer-based models tailored for HSI classification. In terms of the training process, a comprehensive analysis is performed, contrasting classical CNN models and vision Transformer-based counterparts, with particular attention to disturbance robustness and the distribution of the largest eigenvalue of the Hessian. From the evaluations conducted on various mixer models rooted in the unified architecture, it is concluded that the unique strength of vision Transformers can be attributed to their overarching architecture, rather than being exclusively reliant on individual multi-head self-attention (MSA) components.
Extensive experiments demonstrate that the derived vision Transformer models, based on the unified architecture, surpass the classical methods when applied to multiple hyperspectral benchmark datasets. 
\end{abstract}


\section{Introduction}
 \label{Introduction}
\IEEEPARstart{H}{yperspectral} imagery (HSI) enables detailed material identification by representing the reflectance spectra of objects via hundreds of contiguous bands. HSI data are used in diverse applications including environmental monitoring, precision agriculture, geology, urban mapping, and defense \cite{li2019deep, chen2023spectraldiff, fang2023towards, xu2023multiscale}. 
Owing to the rapid advancements in deep learning \cite{cheng2020s, li2021metasaug, xiong2023stereoflowgan, lu2020rskdd, wu2023querying, wu2023efficient}, CNN architectures have emerged as the predominant standard for HSI classification in recent years. In \cite{song2018hyperspectral}, a deep feature fusion CNN is utilized to categorize each pixel of HSI data. To bolster extraction of spectrally-based features, \cite{hamida20183,he2017multi} introduce 3D-CNNs for HSI classification. Additionally, an attention mechanism can be integrated into the CNN framework to facilitate band selection for HSI data, as demonstrated in \cite{zhu2020residual}.
The efficacy of CNN-based HSI classification faces two significant limitations: 1) CNNs often struggle to adequately capture long-range dependencies; 2) The adoption of small input image window patches serves as a compromise between the high dimensionality of HSI data and its corresponding lower spatial resolution. This restricts the design possibilities of the network, impacting its depth and width.
In the past three years, the appeal of using vision Transformers for HSI classification has grown \cite{liu2023cnn, yang2022hyperspectral, wang2023dcn}. This is attributed to the understanding that the spectral dimension of HSI parallels sequence data, irrespective of whether analysis is conducted at the pixel or patch level. In \cite{hong2021spectralformer}, group-wise spectral embedding is employed for HSI classification. Similarly, \cite{mei2022hyperspectral} introduces a group-aware hierarchical vision Transformer to strengthen HSI classification. Furthermore, the LESSFormer design, as presented in \cite{zou2022lessformer}, aims to increase the capture of local information using adaptive spectral-spatial tokens. However, some have suggested that this configuration compromises the inductive bias inherent in CNNs \cite{qi2023global, yang2022hyperspectral}. To address this, some have integrated vision Transformer and CNN modules, either in parallel or sequentially, to harness the advantages of both \cite{tu2022maxvit, ouyang2023multigranularity}. Owing to the scalability of vision Transformers, they typically have a higher number of parameters compared to traditional CNNs. Incorporating an additional CNN branch on top of the multi-head self-attention (MSA) typically leads to a further increase in the model's parameter size. At the same time, it has been noted that the overarching structure of vision Transformers, rather than just the MSA mixer, is pivotal to delivering top-tier performance \cite{tolstikhin2021mlp}. This notion is further emphasized in studies where MSAs are substituted for multi-layer perceptrons (MLPs), as highlighted in \cite{shao2022spatial, guo2022hire, yu2023metaformer, chen2021pre}.
While vision Transformer-based network architectures presently have a pronounced edge in HSI classification-based metrics relative to CNNs, the associated exploration predominantly remains empirical. Thus, this field continues to struggle with pivotal questions:
(1) Does MSA serve as the \emph{critical component} in vision Transformers that enhances HSI classification?
(2) What fundamental differences exist relative to the training process for vision Transformer-based models and CNNs in analysis of hyperspectral datasets?

To address these questions, this paper proposes a unified hierarchical spectral vision Transformer architecture designed to integrate discriminative features for HSI classification. Notably, the simple yet effective unified architecture can be seamlessly integrated with any type of mixer block to construct a novel vision Transformer model. In this paper, various mixer modules, including the CNN-mixer, spatial self-attention (SSA)-mixer, channel self-attention (CSA)-mixer, SSA+CNN-mixer, and CSA+CNN-mixer, are independently integrated into the unified architecture, resulting in multiple vision Transformer models. A comprehensive analysis is conducted on these derived vision Transformer models and classical models, considering both the macroscopic aspect of the disturbance robustness and the microscopic aspect of the distribution of the maximum eigenvalue of the Hessian after the training process. In this paper, the term 'Hessian' specifically refers to the Hessian of the loss function relative to the parameters of the network. A key goal of this study is to explore the influence of different mixers on training vision Transformer-based models. A comprehensive comparison is conducted to explore and highlight fundamental differences between the vision Transformer and CNN models. To the best of our knowledge, this is the first paper to thoroughly investigate the key factors behind the superior performance of vision Transformers in HSI classification. Other contributions include: a) Conducting a rigorous evaluation of the training process for both CNNs and vision Transformers; b) Demonstrating that the unified architecture, rather than the MSA modules contribute to the superior performance observed with vision Transformers in HSI classification.

The remainder of this paper is structured as follows: 
%A brief review of related works is provided in Section \ref{Related_Work_hyper}. 
Related work is summarized in Section \ref{Related_Work_hyper}. The proposed method is detailed in Section \ref{Proposed_method_hyper}. The experimental setup and results are presented in Section \ref{sec:result}. Section \ref{conlucison} includes the conclusion.

\section{Related Work} \label{Related_Work_hyper}

\textbf{CNN-based HSI classification}: 
As highlighted in the first review of deep learning-based HSI classification \cite{li2019deep}, traditional machine learning techniques often fall short in addressing the unique challenges inherent in HSI classification, and particularly the significant spatial variability of spectral signatures. Over the past decade, the application of CNN models has advanced significantly, both in terms of enhanced performance and efficiency in HSI classification. Compared to traditional machine learning techniques, CNN-based methods excel in their ability to capture localized and discriminative spatial information, all while exhibiting resilience to translations and other variations.
In \cite{yu2017convolutional}, a streamlined, end-to-end CNN structure utilizing 1 $\times$ 1 convolutional layers is adopted for HSI classification. \cite{yang2016hyperspectral} introduces a dual-channel CNN, crafted to jointly exploit spectral-spatial features from HSI. \cite{yue2022spectral} develops a spectral-spatial latent reconstruction framework that concurrently reconstructs spectral and spatial features, while also performing pixel-wise classification with high accuracy. \cite{yang2021enhanced} formulates a novel
enhanced multiscale feature fusion network to extract sufficiently
multiscale features from the parallel multipath architecture of three stages for
HSI classification. Additionally, \cite{yang2023can} implements a novel online spectral information compensation network for HSI classification. However, conventional 1D and 2D-CNNs often fall short in concurrently leveraging both spatial and spectral discriminative information. Recognizing this gap, researchers pioneered 3D-CNN architectures. For instance, \cite{paoletti2018new} investigates an enhanced 3D deep CNN encompassing five layers. Furthermore, \cite{yang2018hyperspectral} proposes a distinctive recurrent 3D-CNN, designed to refine the 3D-CNN model by progressively diminishing the patch size. \cite{ahmad2020fast} formulates a streamlined 3D-CNN model with minimal parameters, resulting in a notable reduction in duration to convergence, while boosting accuracy. However, it should be noted that 3D-CNN models may encounter challenges such as overfitting and substantial computational demands. Aiming to alleviate such issues, \cite{yu2020simplified} suggests a synergistic methodology that intertwines 2D-CNN and 3D-CNN. In this approach, the 2D-CNN is employed to extract spatial features, while the 3D-CNN, using small kernels, focuses on inter-band correlations. Complementing this, \cite{ge2020hyperspectral} proposes a 2D-3D CNN that incorporates a multi-branch feature fusion architecture. Some researchers specifically design CNN variants to efficiently extract feature representations \cite{xia2022vision, meng2024toward}. Notably, \cite{meng2024toward} proposes a novel geometry-aware convolutional foundation model that excels in learning unique geometry- and category-aware features and is informed by vehicle kinematics information to significantly enhance inclusive object detection and extend the perception range. Additionally, HSI shows category imbalance and complex spatial-spectral distributions, limiting adaptation performance. To address these issues, \cite{feng2024class} proposes a class-aligned and class-balancing generative domain adaptation method for HSI classification. Similarly, \cite{10225489} presents a novel framework with multigranularity generators and discriminators that uses adversarial and contrastive learning to continuously improve discriminator classification performance with diverse generated samples.

Recently, attention modules have gained widespread popularity in the field of deep learning, owing to their plug-and-play capability and their effectiveness in enhancing neural network performance \cite{lu2023hregnet, fang2023hyperspectral, cao2023ghostvit}. In \cite{lu2023hregnet}, a hierarchical network for efficient and accurate outdoor LiDAR point cloud registration is proposed by introducing an attention-based neighbor encoding module to gather neighborhood information. In pioneering work in instance-level HSI classification, \cite{fang2023hyperspectral} proposes a novel spectral–spatial feature pyramid network, which integrates multi-scale spectral and spatial information for instance segmentation in HSI. In \cite{cao2023ghostvit}, a ghost attention mechanism is proposed to significantly reduce both the parameters and FLOPs of the vision Transformer while achieving similar or better accuracy. The introduction of attention modules offers an alternative approach to boost HSI classification accuracy. These modules, by selectively emphasizing the most discriminative regions of an input small window patch or feature map, guide the network to focus on pivotal areas. Through the allocation of differential weights to various pixels, the attention mechanism captures essential details, ignoring extraneous information. This refinement contributes to the network's more accurate predictions. In \cite{dong2022weighted}, a pixel classification CNN is complemented with a superpixel-based graph attention network. The work in \cite{zhan2022enhanced} melds a spectral-spatial attention network with ResNet for HSI classification. Recognizing the potential of harnessing long-range semantic information, \cite{shang2022simplified} introduces an adaptive projection attention technique. Concurrently, several studies corroborate that the integration of attention modules significantly improves HSI classification accuracies, as evidenced by \cite{xu2022residual, zhai2022double, liu2022hyperspectral}.

\textbf{Transformer-based HSI classification}:
Over the last three years, the vision Transformer has excelled in the realm of HSI classification, showcasing its distinctive advantage in handling data sequences. The work in \cite{hong2021spectralformer} introduces SpectralFormer, which integrates group-wise spectral embedding and cross-layer adaptive fusion modules. Specifically, the group-wise spectral embedding is adept at capturing feature embeddings from adjacent spectral bands. This combination facilitates the capture of detailed local spectral representations and promotes the transmission of memory-like components from superficial to deeper layers. Meanwhile, \cite{xu2023multiscale} presents a multiscale and cross-level attention learning network designed to holistically harness both global and local multiscale features of pixels for enhanced classification. In \cite{mei2022hyperspectral}, a technique is introduced that employs grouped pixel embedding to better represent local representations. \cite{sun2022spectral} proposes the spectral-spatial feature tokenization Transformer (SSFTT) approach, crafted to efficiently encapsulate HSI's low, mid, and high-level semantic features. Aiming to optimize classification and reduce computational overhead, \cite{tu2022local} devises a neighborhood-centric representation of multi-scale HSI features. In \cite{xue2022local}, a novel local vision Transformer, complemented by a spatial partition restore network, is introduced for HSI classification. \cite{zou2022lessformer} details LESSFormer, a design for HSI classification that converts HSI into adaptable spectral-spatial tokens. These tokens are then enriched to capture both local and extensive data nuances. Addressing the vision Transformer's predominant focus on global data, \cite{gao2022fusion} integrates it with a CNN, aiming to extract local features and thereby enhance classification. \cite{ouyang2023multigranularity} develops a hybrid Transformer, merging multi-granularity tokens with spatial-spectral attention to model spatial-spectral information. Additionally, \cite{yan2023hybrid} implements a dual-branch architecture, combining the CNN and vision Transformer to seamlessly fuse spectral and spatial features. \cite{zhao2022joint} proposes a novel hybrid deep learning network that systematically combines hierarchical CNNs and Transformers for feature extraction and fusion. This approach effectively learns spatial-spectral features in HSIs and elevation information in LiDAR, significantly enhancing the accuracy of the joint classification. Similarly, \cite{chen2023shallow} introduces a novel layered architecture that integrates Transformer with CNN, utilizing a feature dimensionality reduction module and a Transformer-style CNN module to extract shallow features and enforce texture constraints, while employing the original Transformer encoder to extract deep features. Inspired by the observation that high-frequency information captures local details and low-frequency information provides global smooth variations, \cite{qiao2023dual} develops a frequency domain feature extraction vision Transformer network for HSI classification. The work in \cite{yang2023overcoming} puts forward three essential elements for efficient HSI classification through the integration of vision Transformer and CNN networks: extensive exploration of available features, effective reuse of representative features, and differentiated fusion of multi-domain features. Utilizing masked autoencoders' self-supervised training paradigm \cite{he2022masked}, some researchers adopt a masked image modeling strategy for remote sensing image classification \cite{hong2024spectralgpt, chen2024lfsmim, liu2022band}. \cite{hong2024spectralgpt} develops a novel 3D generative pretrained Transformer architecture based on masked autoencoders for remote sensing applications. \cite{chen2024lfsmim} introduces LFSMIM, a self-supervised network for HSI classification that employs low-pass filtering to construct the target domain within the masked image modeling framework. \cite{liu2022band} proposes an unsupervised band selection framework that captures nonlinear relationships between bands and leverages spatial information in HSI. From these studies, it is evident that while vision Transformers have advanced HSI classification accuracy compared to CNN models, the majority of research has concentrated on empirical modifications to the self-attention modules, such as integrating CNN modules or altering feature embeddings. Specifically, the current analysis of model enhancements relies heavily on metric outcomes and prediction maps, without a thorough exploration of the variations throughout the training phase. This overlooks a critical element that contributes to the superior performance of vision Transformer architecture in HSI classification.

To bridge these gaps, a unified architecture for HSI classification built upon the vision Transformer is proposed in this paper. Based on the unified architecture, the attributes of the vision Transformer equipped with multiple mixers are investigated from a model training perspective, and the influence of different mixer modules on model performance is explored.

\section{Proposed Method} \label{Proposed_method_hyper}

\begin{figure*}[ht]
\centering
\includegraphics[width=1\linewidth]{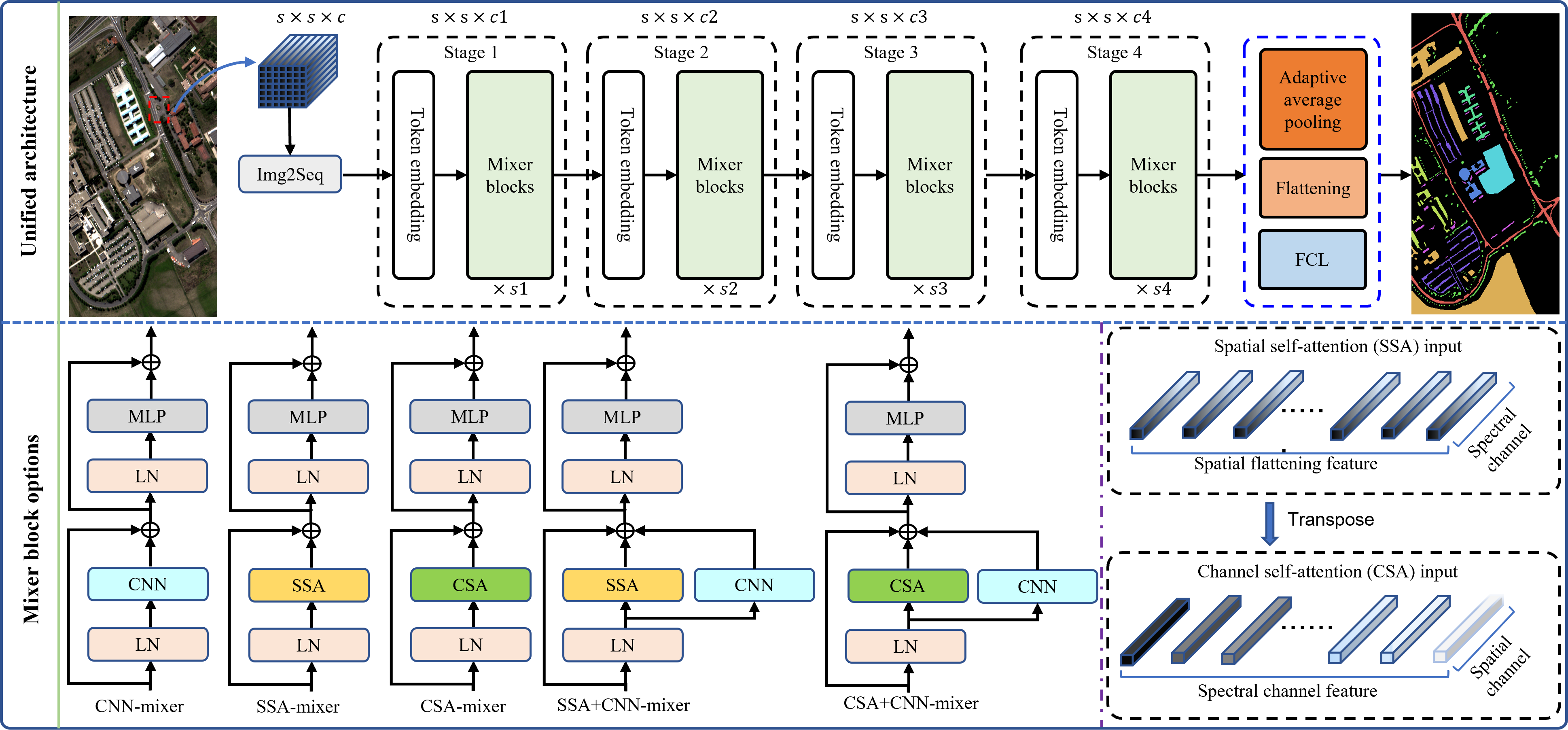}
\caption{Overall framework for HSI classification. The model consists of a unified architecture and mixer block options. The unified architecture is based on a novel hierarchical spectral vision Transformer, specifically tailored for HSI classification. Mixer block options include five common mixer blocks. When different mixers are individually chosen by the mixer blocks, it results in the creation of five unique Transformer models. The visualization in the bottom right corner demonstrates how the SSA-mixer and CSA-mixer can be easily converted on sequence inputs using the transpose operator. Img2Seq: transfer the image to sequence. LN: linear normalization. MLP: multilayer perceptron. CNN: convolutional neural network. SSA: spatial self-attention. CSA: channel self-attention. FCL: fully connected layer.}
\label{fig: overall_framework_hyper}
\end{figure*}
\subsection{Overall architecture construction.}
\label{ssec:overall_architecture}

The HSI classification model based on the vision Transformer primarily consists of two main modules, as depicted in Fig.~\ref{fig: overall_framework_hyper}: 1) a unified architecture and 2) mixer block options. Subsequent sections provide detailed descriptions of these modules. Additionally, rather than relying exclusively on prediction accuracy and empirical analysis of various models, this paper evaluates disturbance robustness and the distribution of the largest eigenvalue of the Hessian. This evaluation provides insights into the model training process from both macro and micro perspectives.

The original HSI data are denoted as $\textbf{\textit{I}} \in \mathbb{R} ^ {h \times w\times c}$, where $\textit{h}$ and $\textit{w}$ represent the spatial height and width, respectively, and $\textit{c}$ signifies the number of spectral bands. The HSI data $\textbf{\textit{I}}$ are divided into patches using a patch window size of $s \times s$, with each patch represented as $\textbf{\textit{P}} \in \mathbb{R} ^ {s \times s\times c}$. The label assigned to the center point of a patch determines its true label.
The proposed vision Transformer model for HSI classification is designed to categorize the center point of each patch cube. The hierarchical vision Transformer architecture for HSI classification is depicted in Figure~\ref{fig: overall_framework_hyper}, referred to as the \emph{unified architecture}.
The network comprises four stages: $Stage~1$, $Stage~2$, $Stage ~3$, and $Stage~4$. In each stage, feature information is extracted through the iterative stacking of token embedding and the mixer module. The number of layers in each stage is represented by $s1$, $s2$, $s3$, and $s4$ respectively. Given the unique characteristics of the input patch window, discriminative features are accumulated across various layers, emphasizing the information in the spectral and spatial dimensions. The respective channel numbers for each stage are denoted as $c1$, $c2$, $c3$, and $c4$. Similar with the Swin Transformer's linear embedding technique \cite{liu2021swin}, this paper utilizes $nn.Conv2d(\cdot)$ for processing raw-valued features. To guarantee compatibility in data shape with the mixer blocks modules, $Seq2Img(\cdot)$ and $Img2Seq(\cdot)$ are judiciously placed before and after the $nn.Conv2d(\cdot)$ operation, respectively. Importantly, $Seq2Img(\cdot)$ serves as the inverse operation to $Img2Seq(\cdot)$. The token embedding strategy employed here is designed to project the spectral dimension to an arbitrary dimension, without impacting its spatial dimension.
Employing the token embedding module, channel/pixel feature information is consolidated to produce a hierarchical representation that prioritizes the spectral/spatial dimension. Once processed by the token embedding module, the feature signals are relayed to the mixer blocks module for further discriminative feature extraction. After the feature extraction through four stages, the latent representation will further undergo processing by the adaptive average pooling, flattening, and a fully connected layer to output the predicted values for the center position of each patch. Notably, the Swin Transformer achieves hierarchical representation by reducing resolution and simultaneously expanding the number of channels. In contrast, the HSI datasets from Houston 2013, Botswana, and Pavia University consist of high-dimensional input channels, with 144, 145, and 103 channels respectively, which always include redundant information. To effectively extract latent representations without substantially enlarging the parameter size of the vision Transformer architecture, the feature dimensions for the initial three layers are reduced, while the spatial dimension size remains constant. Therefore, the hierarchical paradigm for the proposed unified architecture is achieved by leveraging the spectral dimension.

To promote the model's capacity to generalize across classification tasks, the label smoothing cross-entropy is selected as the loss. It is computed as follows:
\begin{equation}
    \begin{split}
    \label{loss}
    \mathcal{L}(y, \hat{y}) = -\sum_{c=1}^C \left( (1 - \alpha) \cdot y_c + \frac{\alpha}{C} \right) \log(\hat{y}_c)
    \end{split}
\end{equation}
where the $y$ is the ground truth label for the one-hot vector. $\hat{y}$ is the predicted probability distribution. $C$ is the class number. $\alpha$ is set at $0.1$ to control the extent of smoothing. $y_c$ is the value of the $c$-th element in the true label vector, while 
$\hat{y}_c$ is the value of the $c$-th element in the probability distribution vector predicted by the model.

\subsection{Mixer block options}
\label{ssec:overall_architecture}

HSI has tens to hundreds of spectral bands. In HSI, each pixel is characterized by a spectrum comprising reflectance values across these bands. This provides a rich representation of the scene or object, allowing for in-depth analysis and identification of materials or features through their unique spectral signatures. As a result, beyond the patch flattening, each pixel in HSI can also be interpreted as a sequence of data. This characteristic makes it possible to devise a variety of mixer blocks tailored to their specific attributes, including the CNN-mixer, SSA-mixer, CSA-mixer, SSA+CNN-mixer, and CSA+CNN-mixer.

\textbf{CNN-mixer}: Similar to the vision Transformer block from the Swin Transformer, the CNN-mixer module embeds an MLP, but opts for a CNN in place of the MSA mechanism. Notably, it sets itself apart by integrating an inductive bias, which fosters local feature connections.As highlighted in \cite{zhang2023vitaev2}, the CNN-mixer module possesses the capability to model locality, which is governed by the kernel size, as well as scale-invariance. To avoid the influence of the attention mechanism on model performance, this paper employs a simple two-layer convolutional module. The module's representation is as follows:

\begin{equation}
    \begin{split}
    \label{deqn_MBConv}
    CN(\mathbf{X}) = Conv_{3 \times 3}(SiLU(BN(Conv_{3 \times 3}(\mathbf{X}))))
    \end{split}
\end{equation}
where the $Conv_{3 \times 3}$ operation amplifies the channel count fourfold using $3 \times 3$ filters. This is followed by the application of the $BN$ batch normalization. The module further integrates the $SiLU$ activation function, which precedes the $Conv_{3 \times 3}$ operation to refine the features. In this paper, unless stated otherwise, $\mathbf{X}$ represents each patch input of $\textbf{\textit{P}} \in \mathbb{R}^{s \times s \times c}$.

The CNN-mixer module, incorporating the CNN block, is computed as follows:
\begin{equation}
    \begin{split}
    \label{deqn_CNN_mixer}
    \mathbf{\hat{X}}& = Seq2Img(\mathbf{X})\\
    \mathbf{Y}& = \mathbf{\hat{X}} + CN(\mathbf{\hat{X}})\\
    \mathbf{\hat{Y}}& = Img2Seq(\mathbf{Y})\\    
    \mathbf{Z}& = \mathbf{\hat{Y}} + MLP(LN(\mathbf{\hat{Y}}))
    \end{split}
\end{equation}
where $CN(\cdot)$ is the CNN block. $MLP(\cdot)$ is the multilayer perceptron operation. The function $Seq2Img(\cdot)$ denotes a basic reshaping operation that transforms a one-dimensional sequence into a feature map. $Img2Seq(\cdot)$ signifies the inverse operation of $Seq2Img(\cdot)$. Utilizing these reshaping techniques ensures the smooth integration of the CNN-mixer module within the vision Transformer framework.

\textbf{SSA-mixer}: To maximize the benefits of the numerous spectral bands in HSI, the SSA-mixer regards each pixel within a patch window as a sequence. Consequently, the length of the input sequence corresponds to the spectral bands' feature dimension, while the number of sequences is defined by the window size, $s \times s$. This sequential feature information is then input to the MSA module to further distill discriminative features.\\
\begin{equation}
    \begin{split}
    \label{deqn_SSA_mixer}
    \mathbf{Y}& = \mathbf{X} + MSA(LN(\mathbf{X}))\\
    \mathbf{Z}& = \mathbf{Y} + MLP(LN(\mathbf{Y}))
    \end{split}
\end{equation}
where $LN(\cdot)$ is linear normalization. $MSA(\cdot)$ is the computation of MSA. It can be described as follows:
\begin{equation}
    \begin{split}
    \label{deqn_MSA}
    Attention(Q, K, V)& = SoftMax(QK^T/\sqrt{d})V
    \end{split}
\end{equation}
where $Q$, $K$, $V$ are the vectors of $query$, $key$ and $value$. These vectors are produced by projecting the input token embeddings through three distinct linear projection layers. $d$ is the token embedding dimension.

\textbf{CSA-mixer}: In a similar manner, the sequential information derived from token embedding is converted into three-dimensional feature data using the $Seq2Img(\cdot)$ function. These data are subsequently transposed and processed via the $Img2Seq(\cdot)$ function, facilitating its transformation into sequences for each channel. This sequential feature data is then channeled through the MSA module to further refine and extract key features. The process can be detailed as follows:

\begin{equation}
    \begin{split}
    \label{deqn_CSA_mixer}
    \mathbf{\hat{X}}& = Img2Seq(Transpose(Seq2Img(\mathbf{X})))\\
    \mathbf{Y}& = \mathbf{\hat{X}} + MSA(LN(\mathbf{\hat{X}}))\\
    \mathbf{Z}& = \mathbf{Y} + MLP(LN(\mathbf{Y}))\\
    \mathbf{\hat{Y}}& = Img2Seq(Transpose(Seq2Img(\mathbf{\hat{Y}}))   
    \end{split}
\end{equation}
where $Transpose(\cdot)$ operation involves swapping the order of the three axes in the image latent features, facilitating their conversion into sequence data along different directions.

\textbf{SSA+CNN-mixer}: This architecture is designed by integrating a CNN module alongside the SSA-mixer. The goal is to explore potential improvements in the vision Transformer model's HSI classification performance by introducing the CNN module. The structure can be outlined as follows:

\begin{equation}
    \begin{split}
    \label{deqn_SSA+CNN_mixer}
    \mathbf{Y}& = \mathbf{X} + MSA(LN(\mathbf{X})) + Img2Seq(CN(Seq2Img(\mathbf{X})))\\
    \mathbf{Z}& = \mathbf{Y} + MLP(LN(\mathbf{Y}))
    \end{split}
\end{equation}

\textbf{CSA+CNN-mixer}: This architecture is formulated by integrating a CNN module alongside the CSA-mixer. The intention is to explore potential improvements in the vision Transformer model's HSI classification efficacy with the inclusion of the CNN module. The configuration can be detailed as follows:
\begin{small}
\begin{equation}
    \begin{split}
    \label{deqn_CSA+CNN_mixer}
    \mathbf{\hat{X}}& = Img2Seq(Transpose(Seq2Img(\mathbf{X})))\\
    \mathbf{Y}& = \mathbf{\hat{X}} + MSA(LN(\mathbf{\hat{X}})) \\ &\quad + Img2Seq(Transpose(CN(Seq2Img(\mathbf{X}))))\\
    \mathbf{\hat{Y}}& = \mathbf{Y} + MLP(LN(\mathbf{Y}))\\
    \mathbf{Z}& = Img2Seq(Transpose(Seq2Img(\mathbf{\hat{Y}}))   
    \end{split}
\end{equation}
\end{small}

\subsection{Representation of the training process}
\label{ssec:training_process_analyis}
In the evaluation of HSI classification models, especially when contrasting vision Transformer and CNN models, it is common for researchers to focus on performance metrics. They often employ reverse engineering and empirical analysis to emphasize the strength of specific methods. However, to our knowledge, few studies have seriously investigated the unique attributes of vision Transformer models from a model training perspective. This paper delves into the distinctions between vision Transformer and CNN models during the HSI classification training phase, analyzing them through the 'best' pretrained weight disturbance robustness and the largest eigenvalue of the Hessian. Specifically, the 'best' pretrained weight refers to the training weight achieved after completing 300 epochs on the training dataset, while the maximum eigenvalue of the Hessian is calculated using the Hessian matrix. This matrix is constructed from the second-order partial derivatives of the neural network's loss function. It effectively describes the local curvature of a multi-variable function. In the realm of deep learning training, the 'loss landscape' refers to the visualization or portrayal of the loss function across the parameter space of a network \cite{li2018visualizing, park2022vision}. This landscape offers critical insights into the evolution of the loss function as network parameters change during training. It reveals useful insights about the model's behavior relative to the loss during its training phase. Notably, a smoother loss surface in proximity to the closest point tends to improve the model's generalization capabilities. However, given the huge number of parameters in deep learning models, capturing the intricacies of the loss landscape with a simple three-dimensional representation during the training process is a challenging task.

Building on the technique to produce three-dimensional loss landscapes, we can develop a new understanding of the model's training process. By introducing random disturbances along two unique vector directions with different magnitudes, based on the 'best' pretrained weights, the response of the loss value to these shifts can be assessed. This offers an avenue to analyze the robustness of various models to disturbances in post-training. To depict the three-dimensional loss surface subsequent to the disturbance, the model's loss value can be illustrated as follows:

\begin{equation}
    \begin{split}
    \label{loss_landscape}
    V(w_x, w_y) = Loss({\Theta}^\ast + w_x{\nu }_x + w_y{\nu }_y)
    \end{split}
\end{equation}
where ${\Theta}^\ast$ is the 'best' pretrained weight after training, which is stored in the format of the dictionary. $w_x$ and $w_y$ are scale parameters ranging from -1 to 1 \cite{park2022vision}. The vectors ${\nu}_x$ and ${\nu}_y$ are the basis vectors associated with the $x$-axis and $y$-axis, respectively. The procedures outlined in \cite{li2018visualizing} are established through the following two steps. Initially, two new dictionaries are created based on the function $randn(\cdot)$, and these dictionaries are initialized with the same attributes as ${\Theta}^\ast$. Next, the weights and biases of each item in the these dictionaries are normalized separately.

In a given deep learning model, the 'best' pretrained weights act as a baseline, with $w_x$ and $w_y$ serving as the horizontal axes. By varying the values of $w_x$ and $w_y$, introducing different levels of perturbations to the weight, the corresponding loss values of the model on the training dataset can be determined. From the data derived from this set of three-dimensional points, the associated three-dimensional loss surface can be constructed. The framework of calculating the loss value with varying magnitude of disturbance on the 'best' pretrained weight is shown as Algorithm~\ref{alg:disturbance}.

% \RestyleAlgo{ruled}

% %% This is needed if you want to add comments in
% %% your algorithm with \Comment
% \SetKwComment{Comment}{/* }{ */}

% \begin{algorithm}[hbt!]
% \caption{Framework of plotting three-dimensional disturbance robustness surface contour}\label{alg:disturbance}
% \KwData{$Best pretrained weight {\Theta}^\ast$}
% \KwResult{$y = x^n$}
% $y \gets 1$\;
% $X \gets x$\;
% $N \gets n$\;
% \While{$N \neq 0$}{
%   \eIf{$N$ is even}{
%     $X \gets X \times X$\;
%     $N \gets \frac{N}{2} $ \Comment*[r]{This is a comment}
%   }{\If{$N$ is odd}{
%       $y \gets y \times X$\;
%       $N \gets N - 1$\;
%     }
%   }
% }
% \end{algorithm}

\newcommand\mycommfont[1]{\small\ttfamily\textcolor{blue}{#1}}
\SetCommentSty{mycommfont}

% \begin{algorithm}
%     \SetKwFunction{isOddNumber}{isOddNumber}
%     % \SetKwInput{Input}{Input}
%     % \SetKwInput{Output}{Output}
%     \SetKwInOut{KwIn}{Input}
%     \SetKwInOut{KwOut}{Output}

%     \KwIn{A list $[a_i]$, $i=1, 2, \cdots, n$, where each element is an
%     integer.}
%     \KwOut{Processed list.}

%     $newList = [\ ]$

%     \tcc{For odd elements in the list, we add 1, and for even elements, we add 2.
%     After the loop, all elements are even.}
%     \For{$i \leftarrow 0$ \KwTo $n-1$}{
%         \eIf{$\isOddNumber(a_i)$}{

%             $newList.append(a_i + 1)$ \tcp*[f]{Some thought-provoking comment.}
%          }{
%             \tcp{Another comment}
%             $newList.append(a_i + 2)$
%          }
%     }

%     \KwRet{$V(w_x, w_y)$}
%     \caption{Framework of calculating loss value with varying magnitude of best pretrained weight disturbance}\label{alg:disturbance}
% \end{algorithm}

\begin{algorithm}
    \SetKwFunction{isOddNumber}{isOddNumber}
    % \SetKwInput{Input}{Input}
    % \SetKwInput{Output}{Output}
    \SetKwInOut{KwIn}{Input}
    \SetKwInOut{KwOut}{Output}

    \KwIn{'Best' pretrained weight ${\Theta}^\ast$.}
    \KwOut{Loss value array $V_{array}$.}

    \tcc{The loss value is calculated on the training dataset.}

    $V = [\ ]$
    
    % $n \leftarrow 21$

    $w_x \leftarrow np.linspace(-1, 1, n)$
    \tcp*[f]{$n$ represents the number of sampling points.}
    $w_y \leftarrow np.linspace(-1, 1, n)$
    
    Initialize two random normal vectors ${\nu }_x^{ini}$ and ${\nu }_y^{ini}$ 
    \tcp*[f]{${\nu }_x^{ini}$ and ${\nu }_y^{ini}$ are same shape with ${\Theta}^\ast$.}

    Normalize ${\nu }_x^{ini}$ and ${\nu }_y^{ini}$: $\{{\nu }_x[m,n], {\nu }_y[m,n]\} \leftarrow \{\frac{{\nu }_x^{ini}[m,n]}{\lVert{{\nu }_x^{ini}[m,n]}\rVert}\lVert{{\Theta}^\ast[m,n]}\rVert, \{\frac{{\nu }_y^{ini}[m,n]}{\lVert{{\nu }_y^{ini}[m,n]}\rVert}\lVert{{\Theta}^\ast[m,n]}\rVert\}$ \tcp*[f]{${\Theta}^\ast[m,n]$ denotes the $m$-th filter corresponding to the $n$-th layer.}   
    $V_0 = LabelSmoothingCrossEntropy({\Theta}^\ast) + weight\_decay*L_2$ \tcp*[f]{$L_2$ represents regularization.}

    \For{$i \leftarrow 0$ \KwTo $n-1$}{
        \For{$j \leftarrow 0$ \KwTo $n-1$}{

            ${\Theta}_{dis}^\ast = {\Theta}^\ast + w_{x}[i]*{\nu }_x + w_{x}[j]*{\nu }_y $
            
            $V_{align} = LabelSmoothingCrossEntropy({\Theta}_{dis}^\ast) + weight\_decay*L_2 - V_0$
            
            $V.append(V_{align})$
         }
         
         % \eIf{($i == n//2$ and $j == n//2$)}
         % {
         %    \tcp{Another comment}
         %    $newList.append(a_i + 2)$
         % }
         {}
    }

   $V_{array} = np.array(V).reshape(n,n)$\\
    \KwRet{$V_{array}$}
    \caption{Framework of calculating loss value with varying magnitude disturbance of the best pretrained weight}\label{alg:disturbance}
\end{algorithm}

To further investigate local flatness and convergence properties, a qualitative analysis using the maximum eigenvalue of the Hessian is necessary \cite{ghorbani2019investigation}. This explores the local characteristics of the loss surface, highlighting both flat and steep regions. These insights are pivotal in identifying areas that might either impede or aid convergence. The eigenvalue of the Hessian at a given point play a pivotal role in revealing the inherent characteristics of the model's loss function at that specific location. They are instrumental in discerning whether the point under consideration is a local minimum, a local maximum, or a saddle point. Furthermore, they offer valuable insights into the function's curvature in various directions. A negative eigenvalue in the Hessian is indicative of the curvature being concave along at least one direction. In practical terms, this means that a slight movement in the direction of the corresponding eigenvector would lead to an increase in the function's value, signifying that the point in question is not situated in a convex region of the function. Conversely, a scenario in which all the Hessian's eigenvalues at a specific point are positive denotes that the function exhibits local convexity at that juncture, categorizing the point as a local minimum \cite{ghorbani2019investigation}. Based on the 'best' pretrained weight after the training process, this paper conducts a thorough analysis of the distribution of the maximum eigenvalue of the Hessian. In the distribution curve representing the maximum eigenvalue, the ideal situation is for the horizontal coordinate of the curve's peak to not only exceed zero but also remain in close proximity to it. This scenario is indicative of an augmented level of local smoothness in the vicinity of the 'best' pretrained point, a state achieved in post-training. This is indicative of the model’s generalization ability, showcasing its superior performance capabilities. In this paper, the $PyHessian$ tool \cite{yao2020pyhessian} is employed to compute the maximum eigenvalue of the Hessian. Notably, if model parameter gradients are absent, they are excluded from consideration. The derived maximum eigenvalue of the Hessian then becomes the foundation for applying the $KernelDensity$ function from the $sklearn$ library, paired with a Gaussian kernel, to shape a distribution curve.

To this end, an in-depth representation of the distinctions between CNN and vision Transformer models, as well as the impact of different mixer modules on the vision Transformer model in HSI classification, can be illustrated by combining performance metrics and training process analysis.

\section{Experimental Setup and Results}\label{sec:result}
\subsection{Dataset description and implementation detail}
\label{ssec:dataset}
The performance of the proposed vision Transformer models for HSI classification is evaluated using three commonly analyzed HSI datasets: Houston 2013, Botswana, and Pavia University\cite{hyperspectral2023, uh_hyperspectral}.

\begin{figure}
    \centering
    \subfloat[]{\includegraphics[scale=0.13]{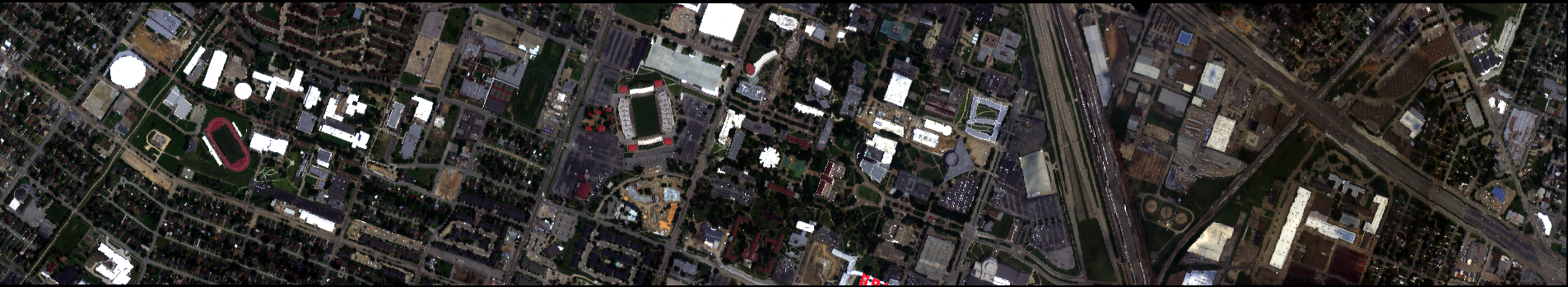}}\par
    \vspace{-0.3cm} 
    \subfloat[]{\includegraphics[scale=0.13]{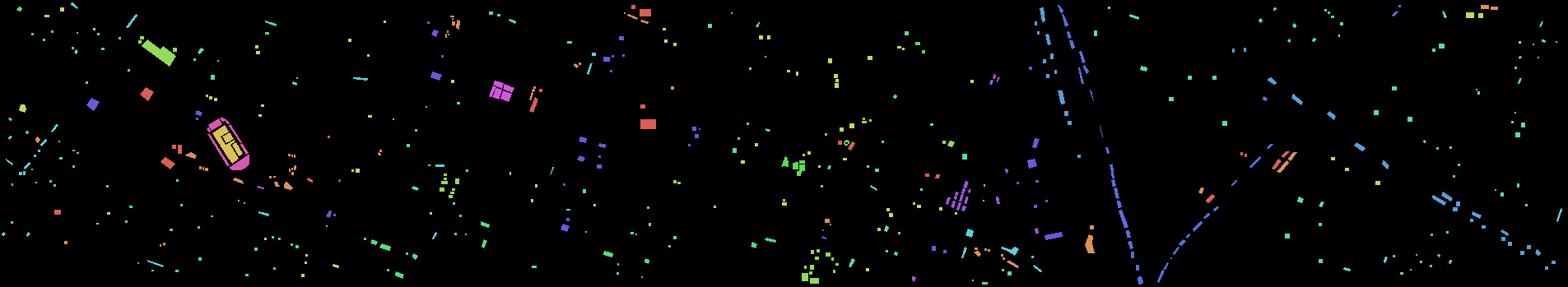}}\par
    \vspace{-0.3cm}    \subfloat{\includegraphics[scale=0.5]{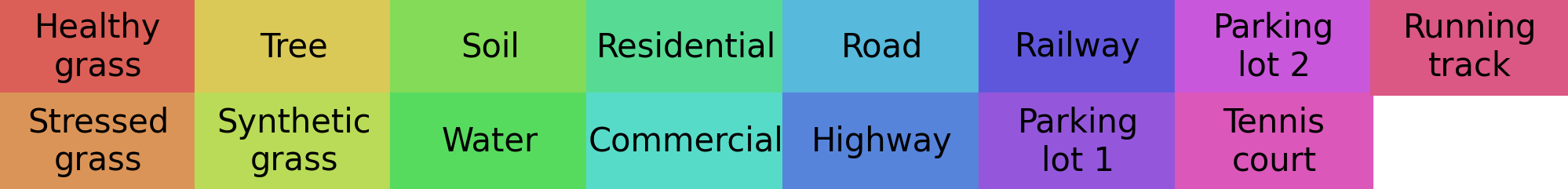}}

\caption{Houston 2013 dataset. (a) False color image (band R: 60, G: 45, B: 20). (b) Ground truth map.}
\label{fig:hu_description}
\end{figure}

1) \textit{\textbf{Houston 2013}}: Houston 2013 airborne hyperspectral data consist of 144 spectral bands. The dataset was collected over the University of Houston campus and the surrounding urban area. It comprises a total of 349 $\times$ 1905 pixels, with each pixel of the orthorectified dataset having a spatial resolution of 2.5m. The dataset has 15 thematic classes. It was partitioned into three subsets for the analysis: a training set (5\%), a validation set (5\%), and a test set (90\%). The class information and the number of training, validation, and testing samples for each class are presented in Table~\ref{Houston2013_dataset}. The false color image and ground reference map of the Houston 2013 dataset are shown in Fig. \ref{fig:hu_description}.

\begin{table}[]
\begin{center}
\caption{Number of samples for each class of Houston 2013 dataset.}
\label{Houston2013_dataset}
\begin{tabular}{l|cccc}
\hline
\multicolumn{1}{c|}{Class} & Training & Validation & Testing & Total \\ \hline
1: Healthy grass              & 63       & 62         & 1126    & 1251  \\
2: Stressed grass             & 62       & 63         & 1129    & 1254  \\
3: Synthetic grass            & 35       & 35         & 627     & 697   \\
4: Trees                      & 62       & 62         & 1120    & 1244  \\
5: Soil                       & 62       & 62         & 1118    & 1242  \\
6: Water                      & 17       & 16         & 292     & 325   \\
7: Residential                & 63       & 64         & 1141    & 1268  \\
8: Commercial                 & 62       & 62         & 1120    & 1244  \\
9: Road                       & 63       & 62         & 1127    & 1252  \\
10: Highway                   & 61       & 62         & 1104    & 1227  \\
11: Railway                   & 62       & 61         & 1112    & 1235  \\
12: Parking lot 1             & 61       & 62         & 1110    & 1233  \\
13: Parking lot 2             & 23       & 24         & 422     & 469   \\
14: Tennis court              & 22       & 21         & 385     & 428   \\
15: Running track             & 33       & 33         & 594     & 660   \\ \hline
\end{tabular}
\end{center}
\end{table}

\begin{figure}
    \centering
    \subfloat[]{\includegraphics[scale=0.12]{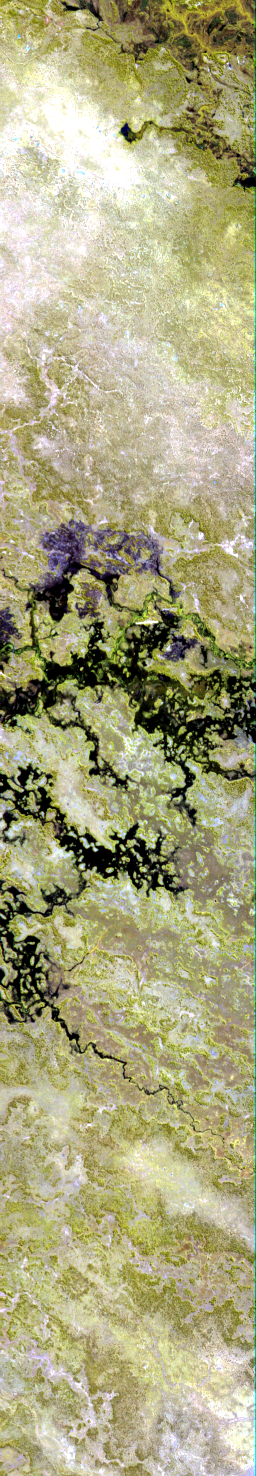}}\vspace{1pt}
    \subfloat[]{\includegraphics[scale=0.12]{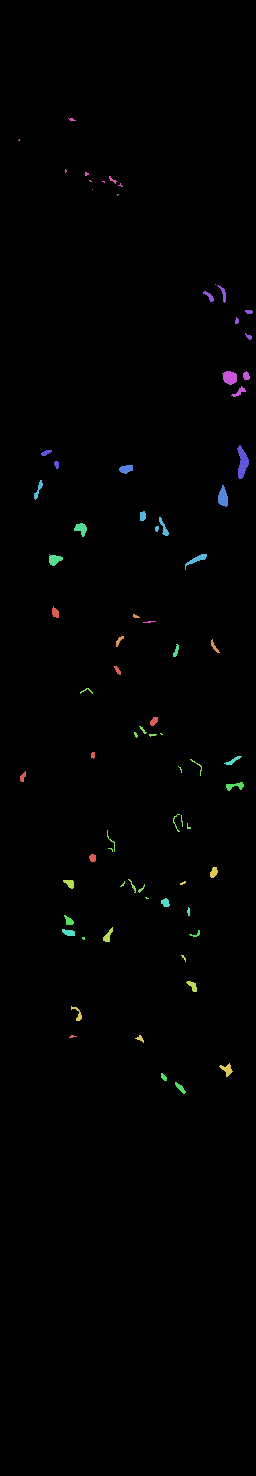}}\vspace{1pt} \subfloat{\includegraphics[trim=0.07cm 0.1cm 0 0, clip, scale=0.7]{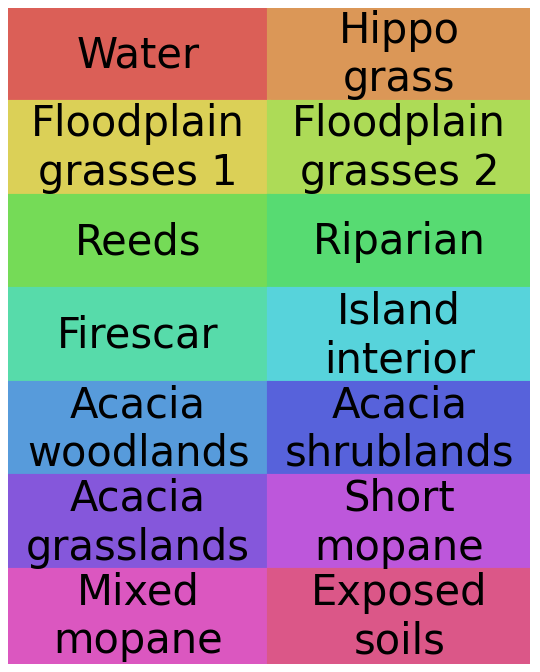}}

\caption{Botswana dataset. (a) False color image (band R: 60, G:45, B: 15). (b) Ground truth map.}
\label{fig:bot_description}
\end{figure}

2) \textit{\textbf{Botswana}}: The Botswana dataset, acquired by the Hyperion sensor on the EO-1 satellite over the Okavano Delta, consists of 242 spectral bands. After eliminating the noisy and water absorption features bands, the dataset has 145 bands. Each pixel in the imagery has a spatial resolution of 30m. 14 classes were identified in the scene. The dataset was partitioned into three subsets for the analysis: a training set (10\%), a validation set (10\%), and a test set (80\%). The class information and the number of training, validation, and testing samples for each class are detailed in Table~\ref{Botswana_dataset}. The false color image and ground reference map of the Botswana dataset are shown in Fig. \ref{fig:bot_description}.

\begin{table}[]
\begin{center}
\caption{Number of samples for each class of Botswana dataset.}
\label{Botswana_dataset}
\begin{tabular}{l|cccc}
\hline
\multicolumn{1}{c|}{Class} & Training & Validation & Testing & Total \\ \hline
1: Water                      & 27       & 27         & 216     & 270   \\
2: Hippo grass                & 10       & 10         & 81      & 101   \\
3: Floodplain grasses 1       & 25       & 25         & 201     & 251   \\
4: Floodplain grasses 2       & 22       & 21         & 172     & 215   \\
5: Reeds                      & 27       & 27         & 215     & 269   \\
6: Riparian                   & 27       & 27         & 215     & 269   \\
7: Firescar                   & 26       & 26         & 207     & 259   \\
8: Island interior            & 21       & 20         & 162     & 203   \\
9: Acacia woodlands           & 31       & 32         & 251     & 314   \\
10: Acacia shrublands         & 24       & 25         & 199     & 248   \\
11: Acacia grasslands         & 30       & 31         & 244     & 305   \\
12: Short mopane              & 18       & 18         & 145     & 181   \\
13: Mixed mopane              & 26       & 27         & 215     & 268   \\
14: Exposed soils             & 10       & 9          & 76      & 95    \\ \hline
\end{tabular}
\end{center}
\end{table}

\begin{figure}
    \centering
    \subfloat[]{\includegraphics[scale=0.25]{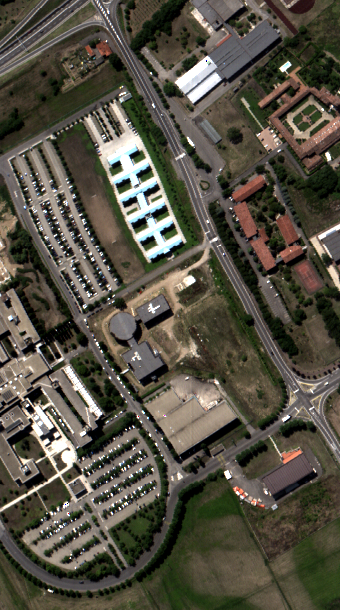}}\vspace{1pt}
    % \vspace{-0.3cm} 
    \subfloat[]{\includegraphics[scale=0.25]{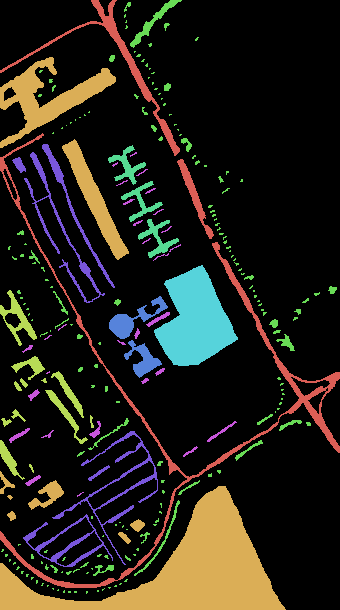}}\vspace{1pt}
    \subfloat{\includegraphics[scale=0.6]{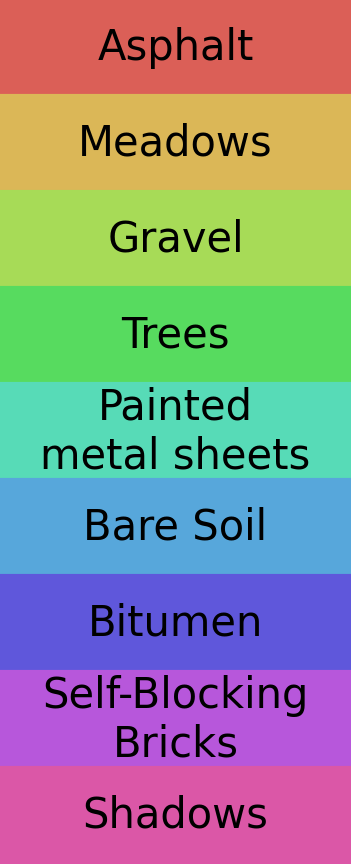}}

\caption{Pavia University dataset. (a) False color image (band R: 40, G: 30, B: 20). (b) Ground truth map.}
\label{fig:pu_description}
\end{figure}

3) \textit{\textbf{Pavia University}}: 
This scene was collected by the ROSIS sensor during a flight campaign over Pavia, northern Italy. There are 103 bands with 1.3m spatial resolution in this 610 $\times$ 340 image, for which 9 classes have been identified. The dataset was partitioned into three subsets for the analysis: a training set (2\%), a validation set (2\%), and a test set (96\%). The class information and the number of training, validation, and testing samples for each class are presented in Table~\ref{pu_dataset}. The false color image and ground reference map of the Pavia University dataset are shown in Fig. \ref{fig:pu_description}.

Hyperspectral data are targeted for specific projects.  Airborne data are expensive to acquire, and high dimensional. The data are standard common testbed data sets for algorithms, and we did not undertake any additional processing on the data. The benchmark data sets we analyze are widely used to compare classification methods.  The Houston data covers the University of Houston and some of the city of Houston.  The focus was to acquire information over a range of targets in an urban area with different spatial and spectral characteristics. Pavia University dataset is high resolution and covers a small area with less diversity in the classes and where the spatial representation of structures such as buildings was uniform and are easy to indicate in the ground reference. Botswana dataset was totally different both in terms of the sensor (30m data from space) and as a natural environment. The ground reference information was obtained using small homogeneous patches obtained on the ground and by interpretation of high resolution remotely sensed imagery. Thus, our analysis covers three totally different scenarios.

The proposed method, along with other established common methods, was implemented in Pytorch. The network was implemented on an NVIDIA Quadro RTX 6000 GPU with 22 GB RAM. The corresponding versions of Pytorch and CUDA were 1.10.1 and 10.2, respectively. The training process consisted of 300 epochs, with a batch size of 64. In this paper, the proposed algorithms utilized the Stochastic Gradient Descent (SGD) optimizer, configured with a learning rate of 0.001, momentum at 0.9, and a weight decay parameter of 0.0001. Parameters for the seven popular algorithms evaluated for comparison are consistent with those in the original papers. For the loss function, all algorithms employed label smoothing cross-entropy, ensuring a consistent methodological approach across the comparative analysis. To provide a quantitative comparison of the proposed method's performance with other classical methods, the evaluation metrics employed were overall accuracy (OA), average accuracy (AA), and kappa coefficient ($\kappa$). Each reported accuracy value represents an average obtained from training with five different random seeds.

\begin{table}[]
\begin{center}
\caption{Number of samples for each class of Pavia University dataset.}
\label{pu_dataset}
\begin{tabular}{l|cccc}
\hline
\multicolumn{1}{c|}{Class} & Training & Validation & Testing & Total \\ \hline
1: Asphalt                      & 132       & 133         & 6366     & 6631   \\
2: Meadows                & 373       & 373         & 17903      & 18649   \\
3: Gravel       & 42       & 42         & 2015     & 2099   \\
4: Trees       & 62       & 61         & 2941     & 3064   \\
5: Painted metal sheets                      & 27       & 27         & 1291     & 1345   \\
6: Bare soil                   & 100       & 101         & 4828     & 5029   \\
7: Bitumen                   & 27       & 26         & 1277     & 1330   \\
8: Self-blocking bricks            & 73       & 74         & 3535     & 3682   \\
9: Shadows           & 19       & 19         & 909     & 947   \\ \hline
\end{tabular}
\end{center}
\end{table}

\subsection{Comparison (baseline) methods}
\label{ssec: comparison method}
  In the comparison study, several representative baseline methods are evaluated, including DFFN \cite{song2018hyperspectral}, CNN3D \cite{hamida20183}, M3D-DCNN \cite{he2017multi}, RSSAN \cite{zhu2020residual}, SpectralFormer \cite{hong2021spectralformer}, SSFTT \cite{sun2022spectral}, GroupTransformer \cite{mei2022hyperspectral}. The DFFN utilizes residual learning to construct a deep 2D-CNN network. The CNN3D integrates traditional CNN architecture with 3D convolution operations. Similarly, the M3D-DCNN jointly learns both 2D multi-scale spatial features and 1D spectral features through a multiscale 3D deep convolutional neural network. The RSSAN combines a spectral-spatial residual attention network with long-short term memory (LSTM) to extract more discriminative spectral and spatial features. In SpectraFormer, which extends the vanilla vision Transformer architecture, a cross-layer skip connection is introduced to merge features across different layers. The SSFTT integrates a 3D convolution layer, a 2D convolution layer, and a vision Transformer module to construct a hybrid CNN-Transformer model for HSI classification. The GroupTransformer introduces a hierarchical Transformer alongside a 2D group convolution network for HSI classification.
 Thus, the aforementioned comparison of methods includes the common 2D and 3D CNNs, as well as the vision Transformer network. To ensure that each class of interest is adequately represented, stratified random sampling was employed for the dataset split. This technique consists of forcing the distribution of the target variables among the different splits to be the same. The strategy results in training on the same population in which it is being evaluated, achieving better predictions. Is is implemented by the function of $sklearn.model\_selection.train\_test\_split()$.

\begin{small}
\begin{table*}[]
\begin{center}
\caption{Parameter size and FLOPs of different models.}
\label{FLOPs}
\scalebox{0.68}{
\begin{tabular}{cc|cccccccccccc}
\hline
\multirow{2}{*}{Datasets}                                                    & \multirow{2}{*}{Complexity} & \multicolumn{4}{c|}{CNN-based method} & \multicolumn{8}{c}{Transformer-based method}                                                                    \\ \cline{3-14} 
                                                                             &                             & CNN3D    & DFFN    & M3D-DCNN  & RSSAN  & SpectralFormer & SSFTT   & GroupTransformer & CNN-mixer & SSA-mixer & CSA-mixer & SSA+CNN-mixer & CSA+CNN-mixer \\ \hline
\multirow{2}{*}{Houston 2013}                                                & Parameters (M)              & 0.52     & 0.51    & 0.68    & 0.09   & 0.24           & 0.67    & 0.97             & 1.05      & 0.47      & 1.02      & 1.23          & 2.54          \\
                                                                             & FLOPs (M)                   & 6054.43  & 3979.24 & 3341.64 & 615.34 & 2429.33        & 2261.94 & 8628.74          & 8172.74   & 4760.42   & 5801.27   & 10636.20      & 16192.77      \\ \hline
\multirow{2}{*}{Botswana}                                                    & Parameters (M)              & 0.11     & 0.51    & 0.17    & 0.09   & 0.23           & 0.68    & 0.98             & 1.10      & 0.48      & 0.36      & 1.28          & 0.91          \\
                                                                             & FLOPs (M)                   & 1230.20   & 1611.91  & 996.46  & 250.53 & 2327.45        & 447.00   & 3241.43          & 3340.80   & 1694.47    & 1544.98    & 4189.52       & 3268.84       \\ \hline
\multirow{2}{*}{\begin{tabular}[c]{@{}c@{}}Pavia \\ University\end{tabular}} & Parameters (M)              & 0.25     & 0.51    & 0.28    & 0.07   & 0.18           & 0.48    & 0.93             & 0.84      & 0.37      & 1.36      & 0.97          & 2.62          \\
                                                                             & FLOPs (M)                   & 4343.47  & 3933.49 & 2282.17 & 523.19 & 1462.65        & 1615.03 & 8271.87          & 6508.44   & 3919.84   & 5324.28   & 8579.87       & 15089.90      \\ \hline
\end{tabular}
}
\end{center}
\end{table*}
\end{small}

\subsection{Model structure and complexity analysis}
\label{ssec: model complexity}
\begin{figure*}
    \centering
    \subfloat[]{\includegraphics[scale=0.6]{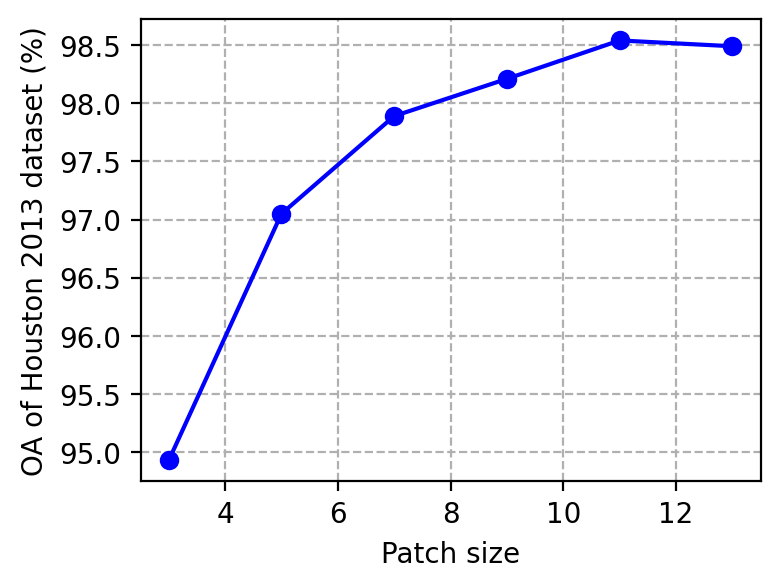}}
    \subfloat[]{\includegraphics[scale=0.6]{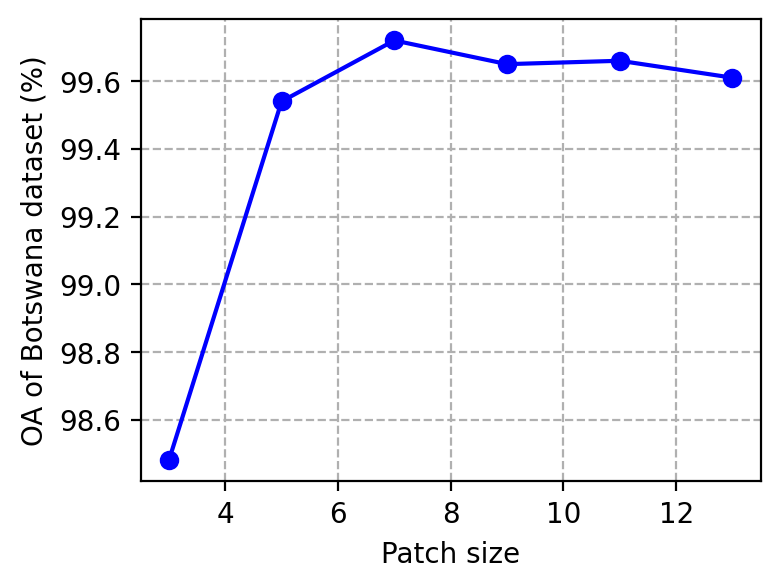}}
    \subfloat[]{\includegraphics[scale=0.6]{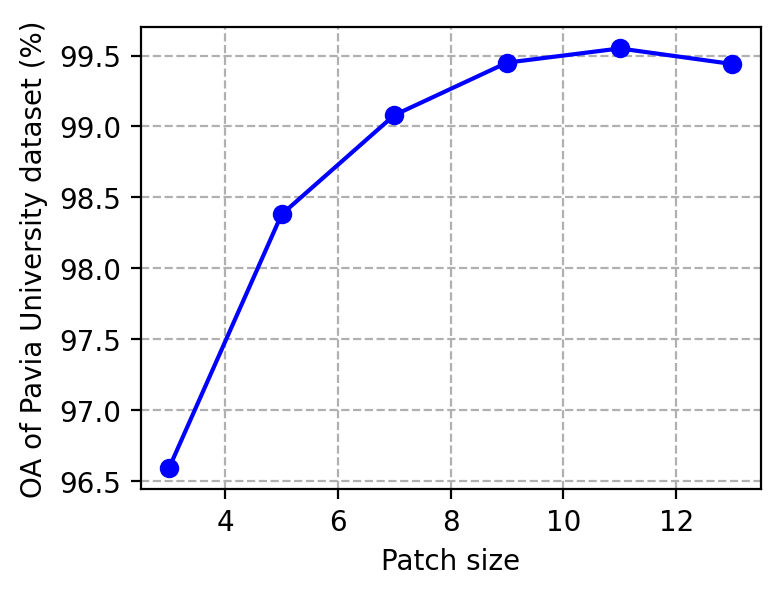}}

\caption{Training patch size effect on the overall accuracy. (a) Houston 2013 dataset. (b) Botswana dataset. (c) Pavia University dataset.}
\label{fig:patch_effect_result}
\end{figure*}

Given the dataset variations outlined in Section \ref{ssec:dataset}, tailoring model structural parameters considering the data characteristics is crucial in HSI classification. Based on the proposed unified architecture, the number of blocks per layer $[s1, s2, s3, s4]$ for the Houston 2013, Botswana, and Pavia University datasets were set to $[3, 2, 4, 2]$, $[3, 3, 2, 2]$, and $[2, 2, 6, 2]$, respectively. The dimension of the features per layer $[c1, c2, c3, c4]$ were set to $[96, 64, 32, 16]$, $[96, 64, 32, 32]$, and $[96, 64, 32, 16]$, respectively. In the joint tuning process of the number of blocks per layer and the dimension of features per layer, the initial setting for the number of blocks per layer was established as $[2, 2, 6, 2]$, with the Swin Transformer serving as a reference. Simultaneously, the dimensions of the features per layer were set to $[96, 64, 32, 16]$, ensuring a gradual decrease in feature dimension as layer depth increased. With these initial settings, the models were constructed to ensure the parameter size comparable to the baseline methods. The models were further optimized by adjusting the number of blocks per layer and the dimension of the last layer's features, with careful consideration to avoid significant changes in the model's overall parameter size. It should be noted that changes to the dimensions of the features in the first three layers were avoided, as they have a greater impact on the size of the parameter size. Furthermore, the selection of patch sizes for each dataset was determined by an analytical comparison study and evaluation of the spatial resolution of the data relative to that of the scale of spatial information in the image. As shown in Fig. \ref{fig:patch_effect_result}, the optimal patch sizes were determined to be 11, 7, and 11 for the Houston 2013, Botswana, and Pavia University datasets, respectively.

Two metrics were introduced to represent the complexity of the model, the size of the parameter set and FLOPs. The results are shown in Table \ref{FLOPs}. All measurement results use a patch cube as input, with a batch size that matches the training batch, which was set at 64. In the Houston 2013 and Pavia University datasets, the models built on SSA-mixer have the smallest number of parameters and FLOPs among the five proposed models. The number of parameters is even less than that of the SSFTT and GroupTransformer algorithms. In the Botswana dataset, the model built on the CSA-mixer has the smallest number of parameters and FLOPs. This is because the patch size in the Houston 2013 and Pavia University datasets was set to 11, which is significantly larger than the patch size of 7 set for the Botswana dataset. As the patch size increases, the number of features in the models built on CSA-mixer increases significantly, leading to a considerable increase in the number of model parameters and FLOPs. Furthermore, it is also observed that the models based on the CNN-mixer, despite their simple construction, do not have the smallest number of parameters and FLOPs among the five proposed models. In addition, after adding a CNN branch to the MSA, the two hybrid vision Transformer models (SSA+CNN-mixer and CSA+CNN-mixer) have more parameters and FLOPs than the classical Transformer models, resulting in greater computational costs.

\subsection{Experimental results}
\label{ssec: Classification result}

\begin{small}
\begin{table*}[]
\begin{center}
\caption{Classification results on the Houston 2013 dataset.}
\label{HU_classification}
\resizebox{\textwidth}{!}{
\begin{tabular}{lcccccccccccc}
\hline
\multicolumn{1}{c|}{\multirow{2}{*}{Class}} & \multicolumn{4}{c|}{CNN-based method}                                                              & \multicolumn{8}{c}{Transformer-based method}                                                                                                                                           \\ \cline{2-13} 
\multicolumn{1}{c|}{}                       & CNN3D                 & DFFN                  & M3D-DCNN                & \multicolumn{1}{c|}{RSSAN} & SpectralFormer & SSFTT                 & GroupTransformer      & CNN-mixer             & SSA-mixer           & CSA-mixer         & SSA+CNN-mixer             & CSA+CNN-mixer           \\ \hline
\multicolumn{1}{l|}{1}                      & 92.43$\pm$3.09           & 96.16$\pm$1.84           & 94.01$\pm$2.73           & 97.50$\pm$1.31                & 92.22$\pm$2.88    & \textbf{98.72$\pm$2.04}           & 97.96$\pm$2.38           & 97.50$\pm$2.24  & 97.99$\pm$1.25           & 98.35$\pm$1.85           & 98.38$\pm$2.64           & 98.15$\pm$1.60           \\
\multicolumn{1}{l|}{2}                      & 99.68$\pm$0.16           & 99.42$\pm$0.20           & 99.27$\pm$0.84           & 98.42$\pm$0.70                & 96.35$\pm$2.09    & \textbf{99.61$\pm$0.20}           & 98.87$\pm$0.85           & 99.03$\pm$0.62           & 99.19$\pm$0.45           & 98.57$\pm$0.62           & 99.47$\pm$0.21           & 99.33$\pm$0.52  \\
\multicolumn{1}{l|}{3}                      & \textbf{100.00$\pm$0.00} & 99.81$\pm$0.19           & \textbf{100.00$\pm$0.00} & 99.87$\pm$0.16                & 99.04$\pm$0.62    & 99.71$\pm$0.43           & 99.84$\pm$0.17           & 99.94$\pm$0.13           & 99.78$\pm$0.13           & 99.65$\pm$0.42           & 99.94$\pm$0.08           & 99.90$\pm$0.13           \\
\multicolumn{1}{l|}{4}                      & 99.57$\pm$0.19  & 99.45$\pm$0.49           & 99.18$\pm$0.54           & 98.79$\pm$1.09                & 96.12$\pm$2.02    & 99.27$\pm$0.78           & 99.11$\pm$0.67           & 99.34$\pm$0.58           & 99.34$\pm$0.35           & \textbf{99.66$\pm$0.47}           & 99.21$\pm$0.56           & 99.14$\pm$0.77           \\
\multicolumn{1}{l|}{5}                      & 99.18$\pm$0.44           & 99.86$\pm$0.20           & 99.19$\pm$0.37           & 98.30$\pm$1.27                & 98.07$\pm$0.66    & \textbf{100.00$\pm$0.00} & \textbf{100.00$\pm$0.00} & 99.98$\pm$0.04 & \textbf{100.00$\pm$0.00} & \textbf{100.00$\pm$0.00} & \textbf{100.00$\pm$0.00}           & 99.95$\pm$0.11           \\
\multicolumn{1}{l|}{6}                      & 87.88$\pm$4.52           & 93.49$\pm$6.08           & 86.78$\pm$5.35           & 82.05$\pm$2.57                & 87.40$\pm$5.19    & 95.89$\pm$5.31           & 96.03$\pm$5.10           & 95.48$\pm$5.00           & 94.25$\pm$5.28           & 94.45$\pm$5.07           & 95.41$\pm$4.12           & \textbf{96.44$\pm$5.11}  \\
\multicolumn{1}{l|}{7}                      & 95.34$\pm$0.78           & 98.23$\pm$0.82           & 96.69$\pm$1.14           & 96.39$\pm$0.57                & 91.15$\pm$3.14    & 98.07$\pm$1.11           & \textbf{99.54$\pm$0.33}           & 99.28$\pm$0.53  & 99.12$\pm$0.66           & 99.16$\pm$0.71           & 99.04$\pm$0.76           & 99.51$\pm$0.54           \\
\multicolumn{1}{l|}{8}                      & 84.38$\pm$3.36           & 92.04$\pm$2.68           & 88.88$\pm$3.05           & 89.95$\pm$1.98                & 87.68$\pm$3.47    & 95.09$\pm$0.91           & 93.91$\pm$1.44           & 94.39$\pm$0.64           & 94.07$\pm$1.28           & 94.91$\pm$1.24           & \textbf{95.43$\pm$0.35}  & 94.52$\pm$0.75           \\
\multicolumn{1}{l|}{9}                      & 91.54$\pm$1.95           & 95.21$\pm$0.68           & 92.83$\pm$1.34           & 93.58$\pm$0.83                & 86.14$\pm$0.98    & 97.04$\pm$0.88           & 97.52$\pm$0.91           & \textbf{98.01$\pm$1.03}           & 97.59$\pm$0.87           & 96.95$\pm$1.06           & 97.37$\pm$1.16           & 97.64$\pm$1.06  \\
\multicolumn{1}{l|}{10}                     & 93.04$\pm$1.11           & 98.79$\pm$0.77           & 96.12$\pm$1.22           & 96.78$\pm$2.11                & 91.52$\pm$2.54    & 99.26$\pm$0.41           & 99.33$\pm$0.82           & 99.78$\pm$0.19           & 99.80$\pm$0.31           & 99.80$\pm$0.20           & \textbf{99.87$\pm$0.25}  & 99.49$\pm$0.85           \\
\multicolumn{1}{l|}{11}                     & 91.76$\pm$1.74           & 97.45$\pm$1.02           & 93.47$\pm$2.23           & 96.46$\pm$0.85                & 89.46$\pm$2.44    & 99.30$\pm$0.96           & 98.87$\pm$1.17           & 99.59$\pm$0.54  & 99.51$\pm$0.97           & 99.78$\pm$0.29           & \textbf{99.86$\pm$0.25}  & 99.51$\pm$0.75           \\
\multicolumn{1}{l|}{12}                     & 92.29$\pm$3.47           & 97.26$\pm$1.53           & 95.17$\pm$0.73           & 95.24$\pm$1.31                & 93.21$\pm$2.31    & 97.21$\pm$1.78           & 97.15$\pm$1.39           & 97.37$\pm$2.11  & 97.75$\pm$0.97           & 97.68$\pm$1.37           & 97.03$\pm$2.21           & \textbf{98.25$\pm$1.02}           \\
\multicolumn{1}{l|}{13}                     & 91.04$\pm$1.88           & 97.39$\pm$1.42           & 90.81$\pm$2.36           & 89.05$\pm$3.08                & 69.19$\pm$5.99    & 97.87$\pm$1.49           & \textbf{98.77$\pm$1.60}           & 98.34$\pm$1.68           & 97.30$\pm$2.56           & 96.64$\pm$2.22           & 97.73$\pm$2.27           & 98.67$\pm$1.74  \\
\multicolumn{1}{l|}{14}                     & 99.32$\pm$0.45           & 99.74$\pm$0.52           & 99.53$\pm$0.42           & 98.81$\pm$0.71                & 95.69$\pm$1.44    & \textbf{100.00$\pm$0.00} & \textbf{100.00$\pm$0.00} & \textbf{100.00$\pm$0.00} & \textbf{100.00$\pm$0.00} & \textbf{100.00$\pm$0.00} & \textbf{100.00$\pm$0.00} & \textbf{100.00$\pm$0.00} \\
\multicolumn{1}{l|}{15}                     & 99.93$\pm$0.13           & \textbf{100.00$\pm$0.00} & 99.87$\pm$0.27           & 98.48$\pm$1.13                & 96.73$\pm$1.07    & 99.97$\pm$0.07           & 99.56$\pm$0.88           & \textbf{100.00$\pm$0.00}           & \textbf{100.00$\pm$0.00}           & \textbf{100.00$\pm$0.00} & \textbf{100.00$\pm$0.00}           & \textbf{100.00$\pm$0.00}           \\ \hline
\multicolumn{1}{l|}{OA}                     & 94.41$\pm$0.65           & 97.59$\pm$0.39           & 95.67$\pm$0.40           & 95.97$\pm$0.23                & 91.99$\pm$1.07    & 98.47$\pm$0.41           & 98.38$\pm$0.29           & 98.54$\pm$0.38           & 98.48$\pm$0.26           & 98.50$\pm$0.25           & 98.64$\pm$0.39  & \textbf{98.68$\pm$0.21}           \\
\multicolumn{1}{l|}{AA}                     & 94.49$\pm$0.70           & 97.62$\pm$0.32           & 95.45$\pm$0.66           & 95.31$\pm$0.23                & 91.33$\pm$1.28    & 98.47$\pm$0.48           & 98.43$\pm$0.41           & 98.53$\pm$0.48           & 98.38$\pm$0.28           & 98.37$\pm$0.26           & 98.58$\pm$0.39  & \textbf{98.70$\pm$0.30}           \\
\multicolumn{1}{l|}{Kappa}                  & 93.95$\pm$0.70           & 97.40$\pm$0.42           & 95.32$\pm$0.44           & 95.64$\pm$0.25                & 91.34$\pm$1.16    & 98.34$\pm$0.44           & 98.25$\pm$0.31           & 98.43$\pm$0.41           & 98.36$\pm$0.28           & 98.38$\pm$0.27           & 98.53$\pm$0.42  & \textbf{98.57$\pm$0.23}           \\ \hline
\multicolumn{1}{c}{}                        &                       &                       &                       &                            &                &                       &                       &                       &                       &                       &                       &                       \\
\multicolumn{1}{c}{}                        &                       &                       &                       &                            &                &                       &                       &                       &                       &                       &                       &                       \\
\multicolumn{1}{c}{}                        &                       &                       &                       &                            &                &                       &                       &                       &                       &                       &                       &                      
\end{tabular}
}
\end{center}
\end{table*}
\end{small}

\begin{small}
\begin{table*}[]
\begin{center}
\caption{Classification results on the Botswana dataset.}
\label{bot_classification}
\resizebox{\textwidth}{!}{
\begin{tabular}{lcccccccccccc}
\hline
\multicolumn{1}{c|}{\multirow{2}{*}{Class}} & \multicolumn{4}{c|}{CNN-based method}                                                              & \multicolumn{8}{c}{Transformer-based method}                                                                                                                                                  \\ \cline{2-13} 
\multicolumn{1}{c|}{}                       & CNN3D                 & DFFN                  & M3D-DCNN                & \multicolumn{1}{c|}{RSSAN} & SpectralFormer        & SSFTT                 & GroupTransformer      & CNN-mixer             & SSA-mixer           & CSA-mixer         & SSA+CNN-mixer             & CSA+CNN-mixer           \\ \hline
\multicolumn{1}{l|}{1}                      & \textbf{100.00$\pm$0.00} & \textbf{100.00$\pm$0.00}           & \textbf{100.00$\pm$0.00} & \textbf{100.00$\pm$0.00}    & \textbf{100.00$\pm$0.00} & 99.91$\pm$0.19 & 99.91$\pm$0.19 & \textbf{100.00$\pm$0.00} & \textbf{100.00$\pm$0.00} & \textbf{100.00$\pm$0.00} & \textbf{100.00$\pm$0.00} & \textbf{100.00$\pm$0.00} \\
\multicolumn{1}{l|}{2}                      & 96.54$\pm$2.39          & \textbf{100.00$\pm$0.00}           & \textbf{100.00$\pm$0.00}           & 99.75$\pm$0.49              & 98.52$\pm$1.44           & \textbf{100.00$\pm$0.00} & \textbf{100.00$\pm$0.00} & \textbf{100.00$\pm$0.00} & \textbf{100.00$\pm$0.00} & \textbf{100.00$\pm$0.00} & \textbf{100.00$\pm$0.00} & \textbf{100.00$\pm$0.00} \\
\multicolumn{1}{l|}{3}                      & 98.11$\pm$1.49           & 99.50$\pm$0.77           & 99.50$\pm$0.63           & \textbf{100.00$\pm$0.00}    & 96.12$\pm$2.56           & \textbf{100.00$\pm$0.00} & \textbf{100.00$\pm$0.00} & \textbf{100.00$\pm$0.00} & \textbf{100.00$\pm$0.00} & \textbf{100.00$\pm$0.00} & \textbf{100.00$\pm$0.00} & \textbf{100.00$\pm$0.00} \\
\multicolumn{1}{l|}{4}                      & 98.60$\pm$1.31          & 99.88$\pm$0.23           & 99.30$\pm$0.93           & 97.79$\pm$2.13              & 97.91$\pm$1.86           & 99.88$\pm$0.23 & \textbf{100.00$\pm$0.00}           & \textbf{100.00$\pm$0.00}           & \textbf{100.00$\pm$0.00}           & 99.88$\pm$0.23           & \textbf{100.00$\pm$0.00} & 99.88$\pm$0.23 \\
\multicolumn{1}{l|}{5}                      & 89.40$\pm$3.85          & 98.05$\pm$2.27           & 96.28$\pm$2.56           & 95.35$\pm$3.59              & 78.79$\pm$3.12          & 95.53$\pm$4.27           & 99.35$\pm$0.47           & 98.79$\pm$1.43           & 98.70$\pm$1.62  & 99.53$\pm$0.93           & 97.86$\pm$2.62           & \textbf{99.63$\pm$0.35}           \\
\multicolumn{1}{l|}{6}                      & 78.23$\pm$7.61           & 98.33$\pm$1.37           & 92.84$\pm$3.79           & 98.23$\pm$0.74              & 93.95$\pm$1.72          & 98.23$\pm$1.12           & 97.40$\pm$2.57           & 98.42$\pm$1.60  & 98.42$\pm$1.37           & 98.42$\pm$1.46           & 97.12$\pm$3.16           & \textbf{98.70$\pm$1.15}           \\
\multicolumn{1}{l|}{7}                      & \textbf{100.00$\pm$0.00}           & \textbf{100.00$\pm$0.00}           & \textbf{100.00$\pm$0.00} & \textbf{100.00$\pm$0.00}    & 99.32$\pm$0.66           & \textbf{100.00$\pm$0.00} & \textbf{100.00$\pm$0.00} & \textbf{100.00$\pm$0.00} & 99.71$\pm$0.58 & \textbf{100.00$\pm$0.00} & \textbf{100.00$\pm$0.00} & \textbf{100.00$\pm$0.00} \\
\multicolumn{1}{l|}{8}                      & 95.06$\pm$5.05           & \textbf{100.00$\pm$0.00}           & \textbf{100.00$\pm$0.00}           & \textbf{100.00$\pm$0.00}              & 96.30$\pm$4.40           & \textbf{100.00$\pm$0.00}           & \textbf{100.00$\pm$0.00} & \textbf{100.00$\pm$0.00} & \textbf{100.00$\pm$0.00} & \textbf{100.00$\pm$0.00} & \textbf{100.00$\pm$0.00} & \textbf{100.00$\pm$0.00} \\
\multicolumn{1}{l|}{9}                      & 93.94$\pm$4.57           & 99.76$\pm$0.20           & 97.61$\pm$3.02           & 98.73$\pm$1.81              & 92.59$\pm$4.54           & 98.49$\pm$2.05           & 99.84$\pm$0.32           & \textbf{100.00$\pm$0.00}           & \textbf{100.00$\pm$0.00} & 99.92$\pm$0.16           & \textbf{100.00$\pm$0.00} & 99.92$\pm$0.16           \\
\multicolumn{1}{l|}{10}                     & 97.09$\pm$0.80           & \textbf{100.00$\pm$0.00}           & 99.60$\pm$0.59           & \textbf{100.00$\pm$0.00}              & 97.99$\pm$0.95           & \textbf{100.00$\pm$0.00}           & 99.70$\pm$0.40           & \textbf{100.00$\pm$0.00} & \textbf{100.00$\pm$0.00} & \textbf{100.00$\pm$0.00}           & \textbf{100.00$\pm$0.00}           & 99.90$\pm$0.20           \\
\multicolumn{1}{l|}{11}                     & 96.39$\pm$0.88           & 97.95$\pm$2.07           & 99.84$\pm$0.20           & 99.75$\pm$0.33              & 98.69$\pm$0.40           & 99.51$\pm$0.98           & 99.43$\pm$0.71           & 99.92$\pm$0.16           & 99.67$\pm$0.66           & \textbf{100.00$\pm$0.00} & 99.43$\pm$0.80           & 99.75$\pm$0.49           \\
\multicolumn{1}{l|}{12}                     & 99.31$\pm$0.44           & 99.03$\pm$0.83 & \textbf{100.00$\pm$0.00} & 98.07$\pm$1.60              & 95.45$\pm$2.33           & \textbf{100.00$\pm$0.00} & 99.86$\pm$0.28 & \textbf{100.00$\pm$0.00} & 99.59$\pm$0.83 & \textbf{100.00$\pm$0.00} & 99.86$\pm$0.28 & 99.72$\pm$0.55 \\
\multicolumn{1}{l|}{13}                     & 99.81$\pm$0.23           & \textbf{100.00$\pm$0.00}           & \textbf{100.00$\pm$0.00}           & \textbf{100.00$\pm$0.00}    & 99.53$\pm$0.42           & \textbf{100.00$\pm$0.00} & \textbf{100.00$\pm$0.00} & \textbf{100.00$\pm$0.00} & \textbf{100.00$\pm$0.00} & \textbf{100.00$\pm$0.00} & \textbf{100.00$\pm$0.00} & \textbf{100.00$\pm$0.00}           \\
\multicolumn{1}{l|}{14}                     & 91.05$\pm$2.55          & 96.32$\pm$7.37           & 97.63$\pm$4.74           & 97.11$\pm$3.94              & 84.74$\pm$5.17           & 98.95$\pm$2.11           & 96.84$\pm$5.03           & 98.42$\pm$2.11           & 96.84$\pm$3.87           & 97.37$\pm$3.63           & \textbf{99.47$\pm$0.64}  & 97.63$\pm$3.16           \\ \hline
\multicolumn{1}{l|}{OA}                     & 95.21$\pm$0.78           & 99.28$\pm$0.40           & 98.67$\pm$0.30           & 98.98$\pm$0.38                & 95.24$\pm$0.92           & 99.25$\pm$0.38           & 99.53$\pm$0.24           & 99.72$\pm$0.19  & 99.59$\pm$0.24           & \textbf{99.74$\pm$0.20}           & 99.51$\pm$0.45           & 99.73$\pm$0.08           \\
\multicolumn{1}{l|}{AA}                     & 95.25$\pm$0.81           & 99.20$\pm$0.68           & 98.76$\pm$0.32           & 98.91$\pm$0.52                & 94.99$\pm$1.05           & 99.32$\pm$0.40           & 99.45$\pm$0.40           & \textbf{99.68$\pm$0.17}  & 99.49$\pm$0.33           & 99.65$\pm$0.29           & 99.55$\pm$0.39           & 99.65$\pm$0.19           \\
\multicolumn{1}{l|}{Kappa}                  & 94.81$\pm$0.85           & 99.22$\pm$0.44           & 98.56$\pm$0.33           & 98.89$\pm$0.41                & 94.85$\pm$0.99           & 99.18$\pm$0.42           & 99.49$\pm$0.26           & 99.69$\pm$0.17  & 99.56$\pm$0.26           & \textbf{99.72$\pm$0.22}          & 99.47$\pm$0.48           & 99.71$\pm$0.08           \\ \hline
\multicolumn{1}{c}{}                        &                       &                       &                       &                            &                       &                       &                       &                       &                       &                       &                       &                       \\
\multicolumn{1}{c}{}                        &                       &                       &                       &                            &                       &                       &                       &                       &                       &                       &                       &                       \\
\multicolumn{1}{c}{}                        &                       &                       &                       &                            &                       &                       &                       &                       &                       &                       &                       &                      
\end{tabular}
}
\end{center}
\end{table*}
\end{small}

\begin{small}
\begin{table*}[!t]
\begin{center}
\caption{Classification results on the Pavia University dataset.}
\label{PU_classification}
\resizebox{\textwidth}{!}{
\begin{tabular}{lcccccccccccc}
\hline
\multicolumn{1}{l|}{\multirow{2}{*}{Class}} & \multicolumn{4}{c|}{CNN-based method}                                                           & \multicolumn{8}{c}{Transformer-based method}                                                                                                                                            \\ \cline{2-13} 
\multicolumn{1}{l|}{}                       & CNN3D                & DFFN                 & M3D-DCNN               & \multicolumn{1}{c|}{RSSAN} & SpectralFormer       & SSFTT               & GroupTransformer     & CNN-mixer             & SSA-mixer           & CSA-mixer        & SSA+CNN-mixer            & CSA+CNN-mixer          \\ \hline
\multicolumn{1}{l|}{1}                      & 96.79$\pm$0.50          & 96.44$\pm$2.98          & 97.89$\pm$0.48          & 99.46$\pm$0.32                & 93.57$\pm$1.77          & 99.38$\pm$0.27          & 99.85$\pm$0.10          & 99.73$\pm$0.25  & 99.80$\pm$0.18           & 99.85$\pm$0.20          & 99.59$\pm$0.39          & \textbf{99.88$\pm$0.15}          \\
\multicolumn{1}{l|}{2}                      & 99.46$\pm$0.30          & 95.66$\pm$4.63          & 99.89$\pm$0.07          & 99.82$\pm$0.17                & 99.67$\pm$0.20          & 99.89$\pm$0.07          & 99.89$\pm$0.08          & \textbf{99.97$\pm$0.02}  & 99.95$\pm$0.02           & 99.96$\pm$0.01          & 99.91$\pm$0.08 & 99.92$\pm$0.06          \\
\multicolumn{1}{l|}{3}                      & 76.36$\pm$3.86          & 99.25$\pm$0.62          & 89.71$\pm$3.64          & 96.38$\pm$1.40                & 85.86$\pm$2.63          & 98.00$\pm$1.22          & 97.70$\pm$1.70          & 98.26$\pm$0.86           & 99.31$\pm$0.37           & 98.02$\pm$1.60          & \textbf{99.47$\pm$0.81}          & 98.97$\pm$0.54          \\
\multicolumn{1}{l|}{4}                      & 97.86$\pm$0.34          & 97.32$\pm$3.75          & \textbf{98.48$\pm$0.66}          & 96.38$\pm$0.99                & 93.67$\pm$1.68          & 98.40$\pm$0.76          & 96.89$\pm$0.42          & 98.00$\pm$0.16           & 97.27$\pm$0.49           & 97.38$\pm$0.49          & 97.97$\pm$0.39          & 97.78$\pm$0.26          \\
\multicolumn{1}{l|}{5}                      & 99.95$\pm$0.09          & 99.23$\pm$0.58          & 99.78$\pm$0.20          & 99.89$\pm$0.14                & 99.91$\pm$0.19          & 99.85$\pm$0.18          & 99.88$\pm$0.18          & 99.98$\pm$0.03           & 99.97$\pm$0.04 & \textbf{100.00$\pm$0.00}          & 99.80$\pm$0.40          & 99.98$\pm$0.03          \\
\multicolumn{1}{l|}{6}                      & 94.75$\pm$0.96          & 81.09$\pm$8.77          & 99.18$\pm$0.39          & 99.67$\pm$0.28                & 96.51$\pm$3.66          & 99.95$\pm$0.05          & 99.96$\pm$0.03          & 99.81$\pm$0.31 & 99.98$\pm$0.03           & \textbf{100.00$\pm$0.00}          & 99.90$\pm$0.11          & 99.76$\pm$0.36          \\
\multicolumn{1}{l|}{7}                      & 83.65$\pm$4.21          & 93.39$\pm$3.17          & 93.91$\pm$2.71          & 96.13$\pm$1.92                & 78.20$\pm$4.94          & 99.15$\pm$0.73          & 99.58$\pm$0.21          & 99.51$\pm$0.74           & 99.87$\pm$0.18  & 99.73$\pm$0.33          & 99.81$\pm$0.34          & \textbf{99.94$\pm$0.06}          \\
\multicolumn{1}{l|}{8}                      & 95.09$\pm$1.26          & 81.41$\pm$2.70          & 96.17$\pm$1.20          & 96.08$\pm$1.03                & 88.48$\pm$3.13          & 98.55$\pm$0.46          & 97.13$\pm$1.81          & 99.10$\pm$0.70           & \textbf{98.28$\pm$1.51}  & 97.92$\pm$1.46          & 98.14$\pm$1.96          & 98.34$\pm$1.26          \\
\multicolumn{1}{l|}{9}                      & 99.71$\pm$0.24          & 87.82$\pm$2.71          & 99.85$\pm$0.09          & 99.12$\pm$0.89                & 95.93$\pm$1.96          & 97.40$\pm$1.34          & 98.37$\pm$1.31          & 97.62$\pm$2.50           & 97.27$\pm$2.67           & 97.73$\pm$1.25          & \textbf{98.13$\pm$1.62} & 99.08$\pm$0.49          \\ \hline
\multicolumn{1}{l|}{OA}                     & 96.40$\pm$0.30          & 93.53$\pm$0.35          & 98.38$\pm$0.23          & 98.88$\pm$0.19                & 95.54$\pm$0.53          & 99.43$\pm$0.17          & 99.29$\pm$0.13          & \textbf{99.55$\pm$0.15}           & 99.50$\pm$0.15           & 99.44$\pm$0.11          & 99.50$\pm$0.14 & 99.54$\pm$0.10          \\
\multicolumn{1}{l|}{AA}                     & 93.74$\pm$0.78          & 93.58$\pm$0.62          & 97.21$\pm$0.56          & 98.10$\pm$0.29                & 92.42$\pm$0.62          & 98.95$\pm$0.30          & 98.80$\pm$0.21          & 99.11$\pm$0.47           & 99.08$\pm$0.35           & 98.96$\pm$0.19          & 99.19$\pm$0.32 & \textbf{99.29$\pm$0.11}          \\
\multicolumn{1}{l|}{Kappa}                  & 95.21$\pm$0.41          & 93.01$\pm$0.38          & 97.86$\pm$0.31          & 98.52$\pm$0.25                & 94.07$\pm$0.72          & 99.24$\pm$0.22          & 99.06$\pm$0.17          & \textbf{99.40$\pm$0.20}           & 99.34$\pm$0.19           & 99.25$\pm$0.14          & 99.34$\pm$0.18 & \textbf{99.40$\pm$0.14}          \\ \hline
                                            & \multicolumn{1}{l}{} & \multicolumn{1}{l}{} & \multicolumn{1}{l}{} & \multicolumn{1}{l}{}       & \multicolumn{1}{l}{} & \multicolumn{1}{l}{} & \multicolumn{1}{l}{} & \multicolumn{1}{l}{}  & \multicolumn{1}{l}{}  & \multicolumn{1}{l}{} & \multicolumn{1}{l}{} & \multicolumn{1}{l}{} \\
                                            & \multicolumn{1}{l}{} & \multicolumn{1}{l}{} & \multicolumn{1}{l}{} & \multicolumn{1}{l}{}       & \multicolumn{1}{l}{} & \multicolumn{1}{l}{} & \multicolumn{1}{l}{} & \multicolumn{1}{l}{}  & \multicolumn{1}{l}{}  & \multicolumn{1}{l}{} & \multicolumn{1}{l}{} & \multicolumn{1}{l}{} \\
                                            & \multicolumn{1}{l}{} & \multicolumn{1}{l}{} & \multicolumn{1}{l}{} & \multicolumn{1}{l}{}       & \multicolumn{1}{l}{} & \multicolumn{1}{l}{} & \multicolumn{1}{l}{} & \multicolumn{1}{l}{}  & \multicolumn{1}{l}{}  & \multicolumn{1}{l}{} & \multicolumn{1}{l}{} & \multicolumn{1}{l}{}
\end{tabular}
}
\end{center}
\end{table*}
\end{small}

\begin{figure*}
    \centering
    \subfloat[]{\includegraphics[scale=0.11]{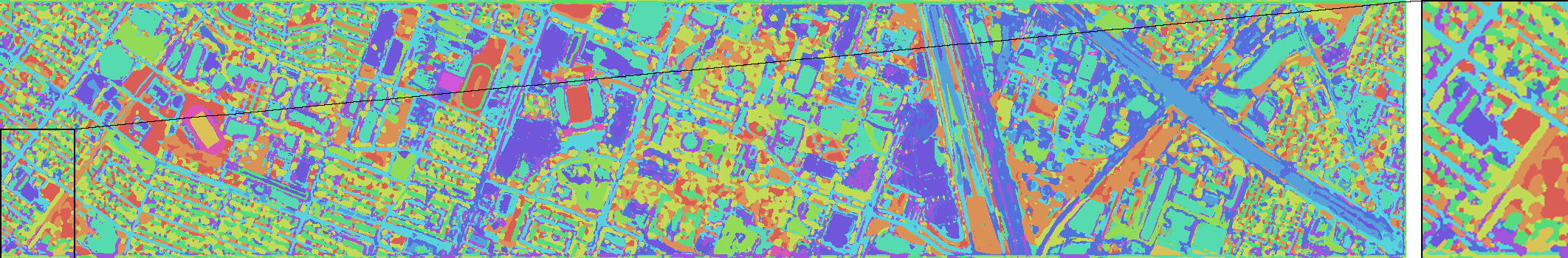}}\hspace{0.15cm}
    \subfloat[]{\includegraphics[scale=0.11]{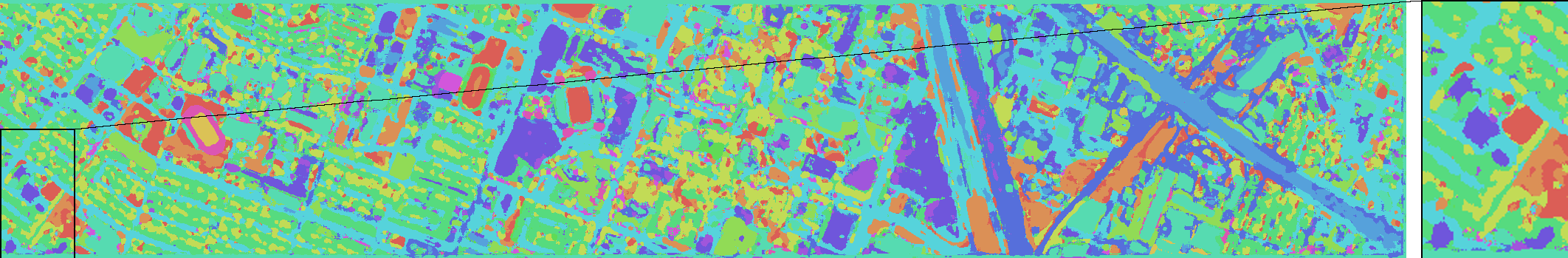}}
    \vspace{-0.3cm} 
    \centering
    \subfloat[]{\includegraphics[scale=0.11]{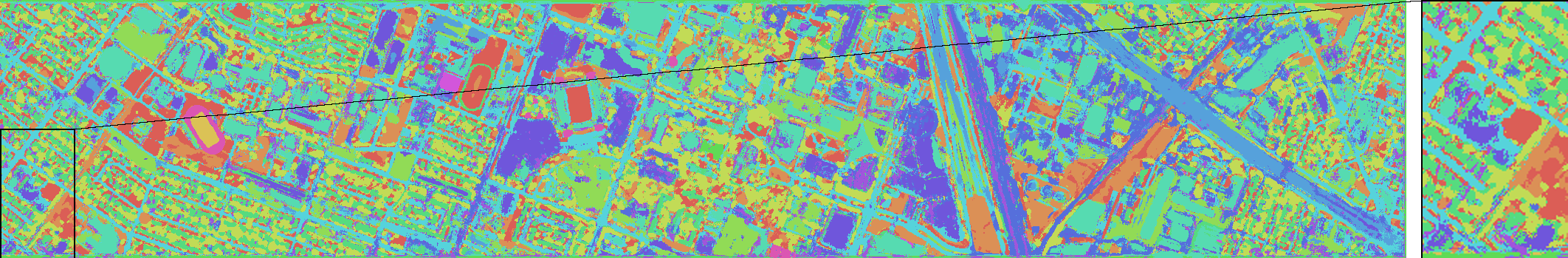}}\hspace{0.15cm}
    \subfloat[]{\includegraphics[scale=0.11]{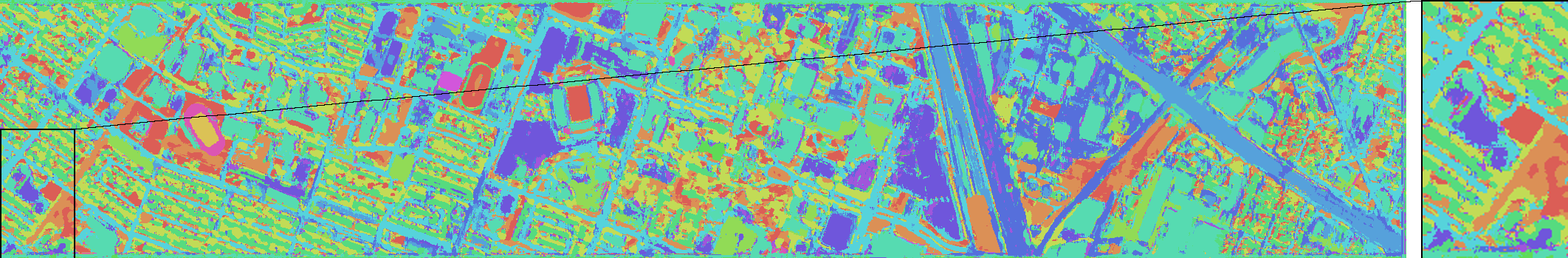}}
    \vspace{-0.3cm} 
    \centering
    \subfloat[]{\includegraphics[scale=0.11]{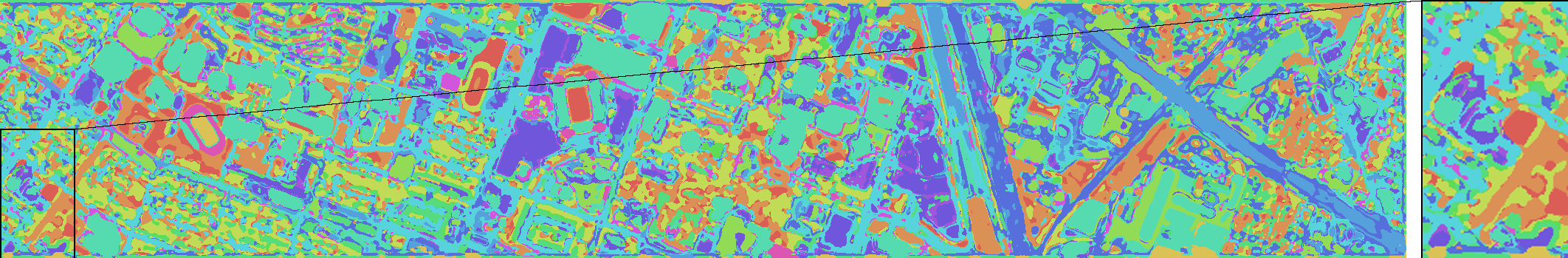}}\hspace{0.15cm}
    \subfloat[]{\includegraphics[scale=0.11]{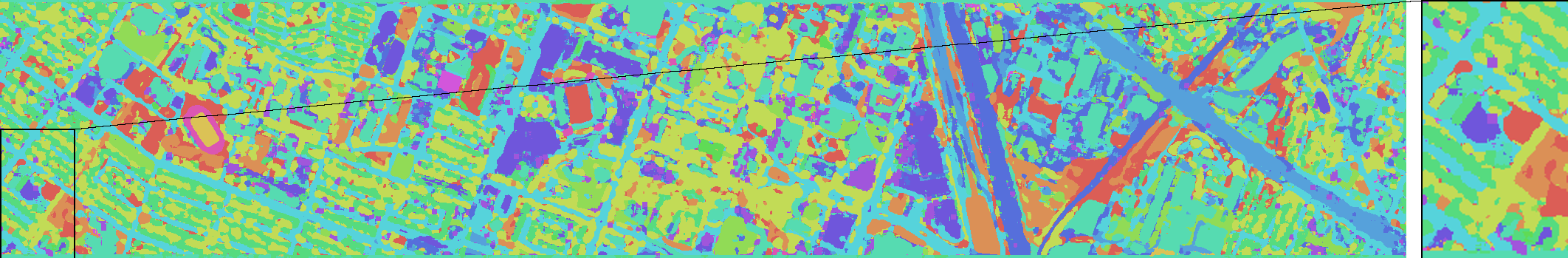}}
    \vspace{-0.3cm} 
    \centering
    \subfloat[]{\includegraphics[scale=0.11]{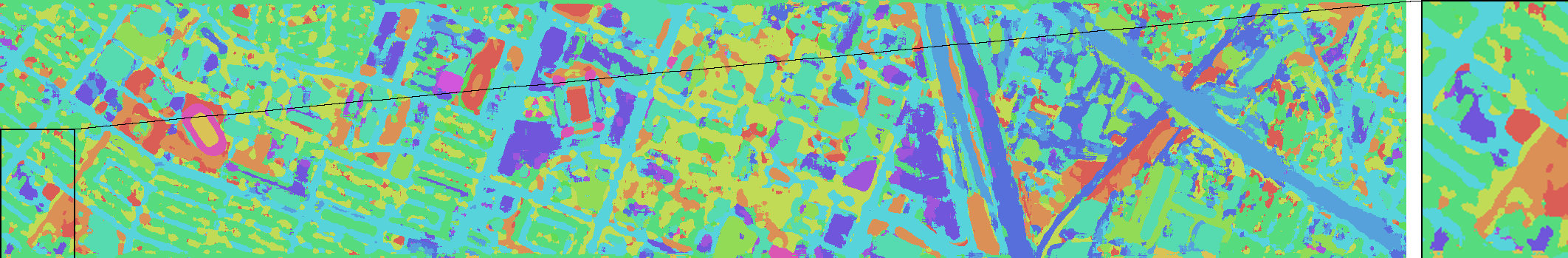}}\hspace{0.15cm}
    \subfloat[]{\includegraphics[scale=0.11]{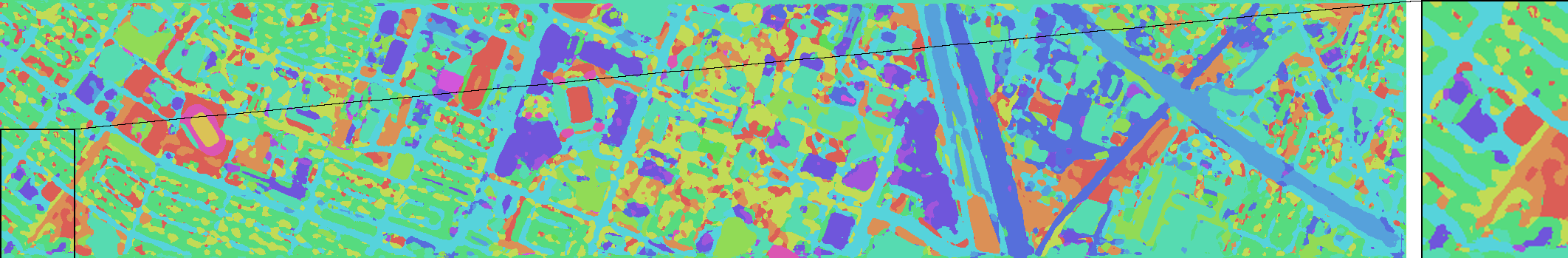}}
    \vspace{-0.3cm} 
    \centering
    \subfloat[]{\includegraphics[scale=0.11]{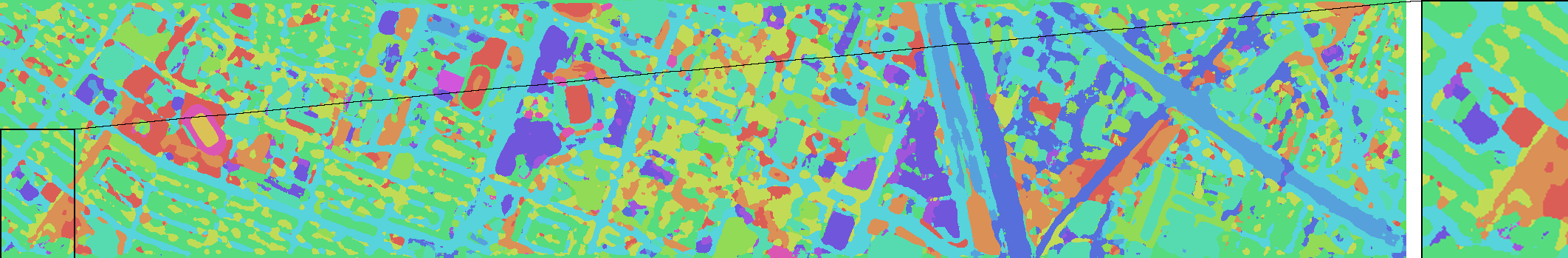}}\hspace{0.15cm}
    \subfloat[]{\includegraphics[scale=0.11]{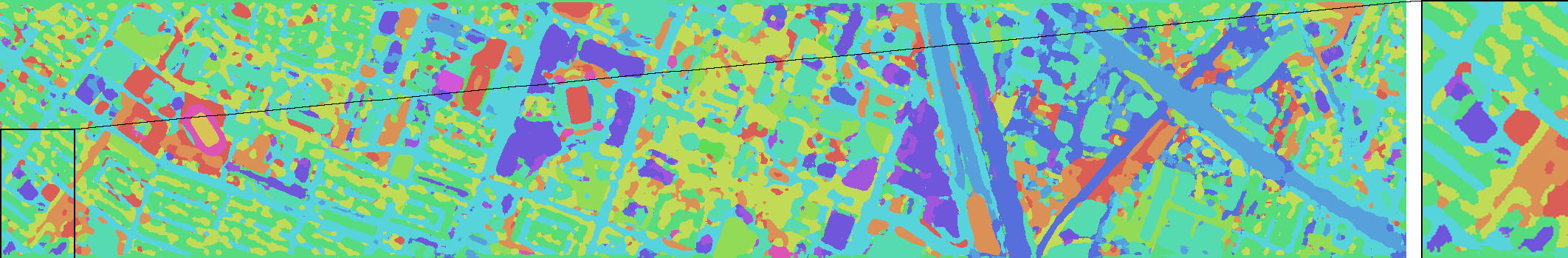}}
    \vspace{-0.3cm} 
    \centering
    \subfloat[]{\includegraphics[scale=0.11]{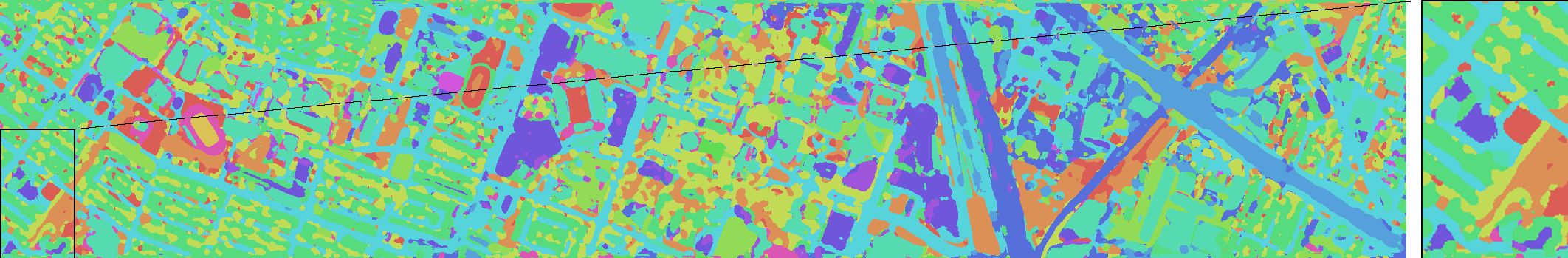}}\hspace{0.15cm}
    \subfloat[]{\includegraphics[scale=0.11]{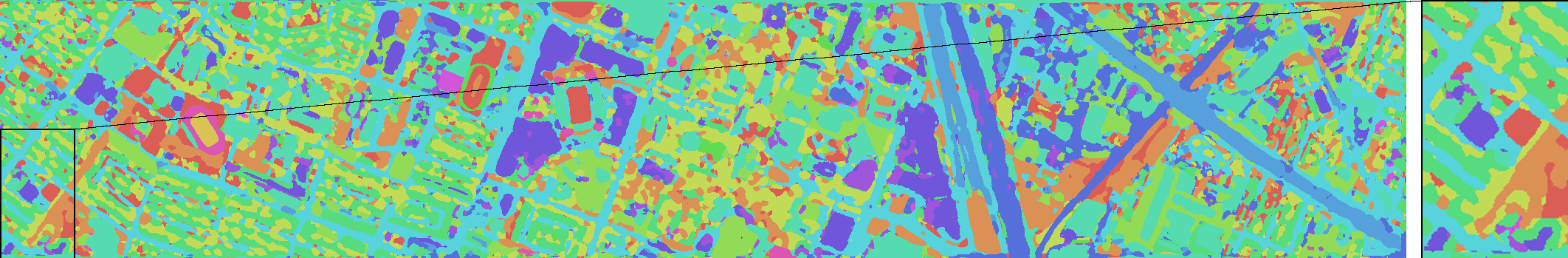}}
    \vspace{-0.3cm} 
    \centering
    % \vspace{0.05cm} 
    % \subfloat[]{\raisebox{\dimexpr \height - 0.76cm}
    \subfloat[]{\includegraphics[scale=0.11]
    {Figure_package/hu/pre_map/hu_gt.png}}\hspace{0.15cm}
    \par
    \vspace{-0.3cm}
    \centering
    {\includegraphics[scale=0.45]{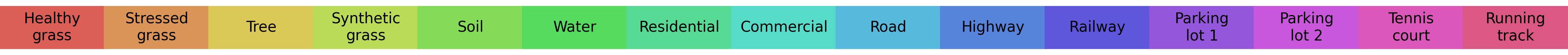}}
    \par
    % \par 

\caption{Prediction map on Houston 2013 dataset. (a) CNN3D, (b) DFFN. (c) M3D-DCNN. (d) RSSAN. (e) SpectralFormer. (f) SSFTT. (g) GroupTransformer. (h) Proposed CNN-mixer. (i) Proposed SSA-mixer. (j) Proposed CSA-mixer. (k) Proposed SSA+CNN-mixer. (l) Proposed CSA+CNN-mixer. (m) Ground truth.}
\label{fig:hu_prediction_map}
\end{figure*}

\begin{figure*}
    \centering
    \subfloat[]{\includegraphics[scale=0.13]{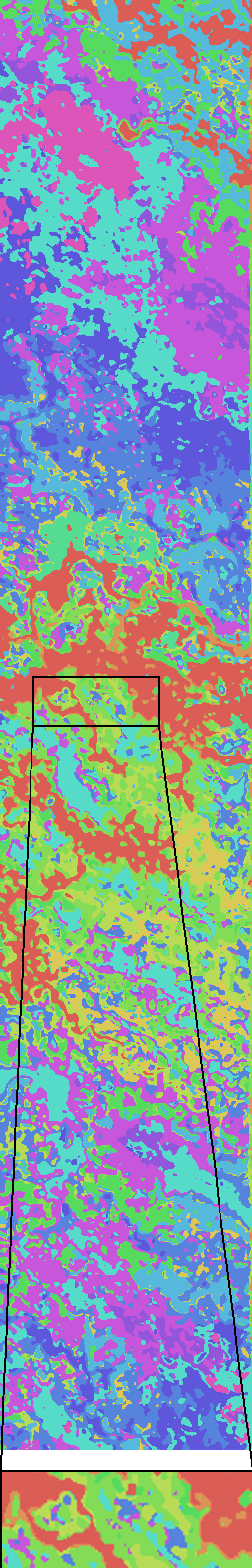}}\hfill
    \subfloat[]{\includegraphics[scale=0.13]{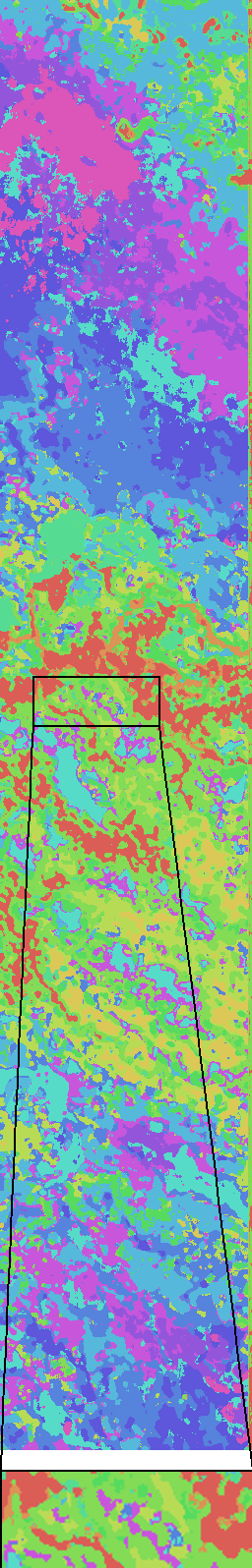}}\hfill
    \subfloat[]{\includegraphics[scale=0.13]{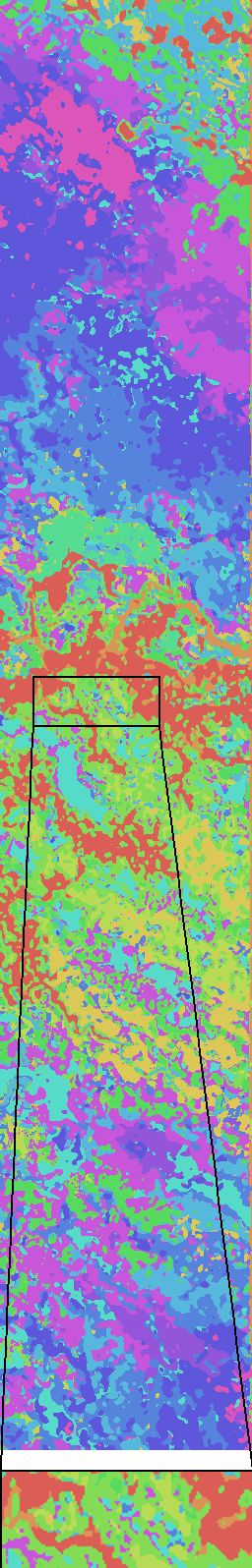}}\hfill
    \subfloat[]{\includegraphics[scale=0.13]{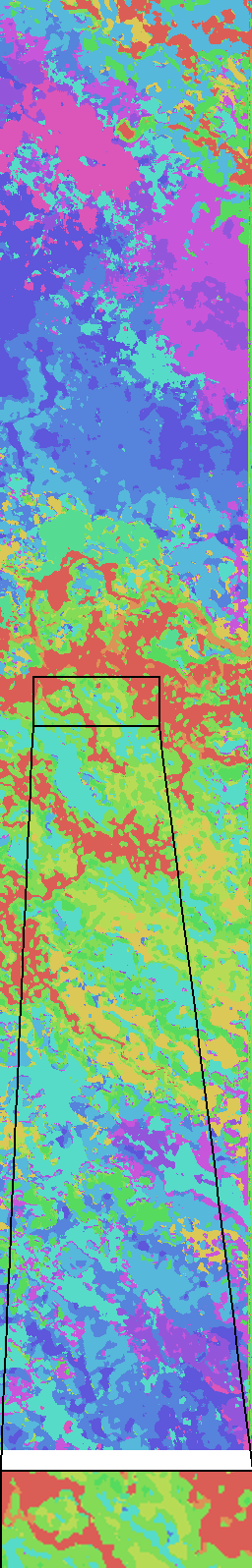}}\hfill
    \subfloat[]{\includegraphics[scale=0.13]{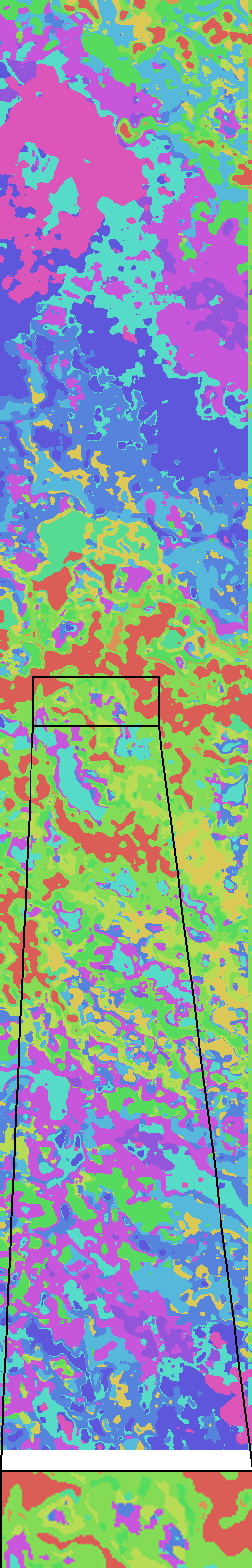}}\hfill
    \subfloat[]{\includegraphics[scale=0.13]{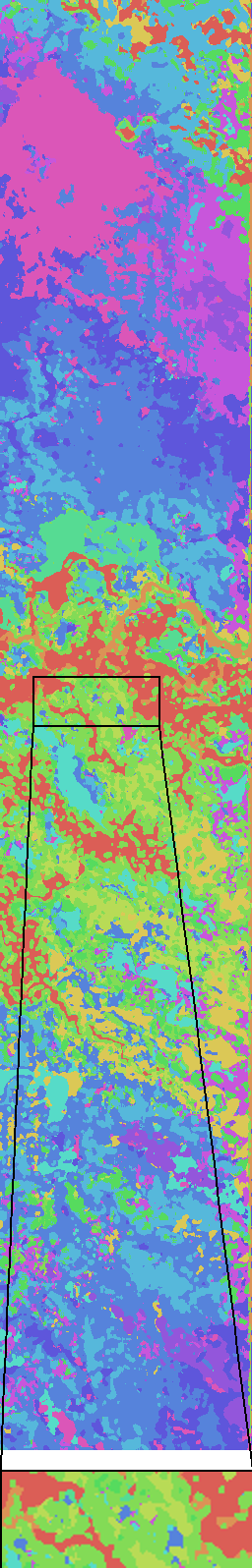}}\hfill
    \subfloat[]{\includegraphics[scale=0.13]{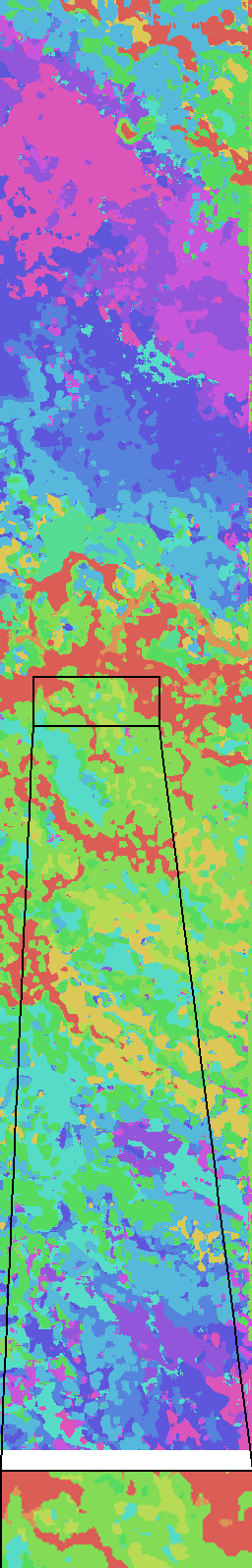}}\hfill
    \subfloat[]{\includegraphics[scale=0.13]{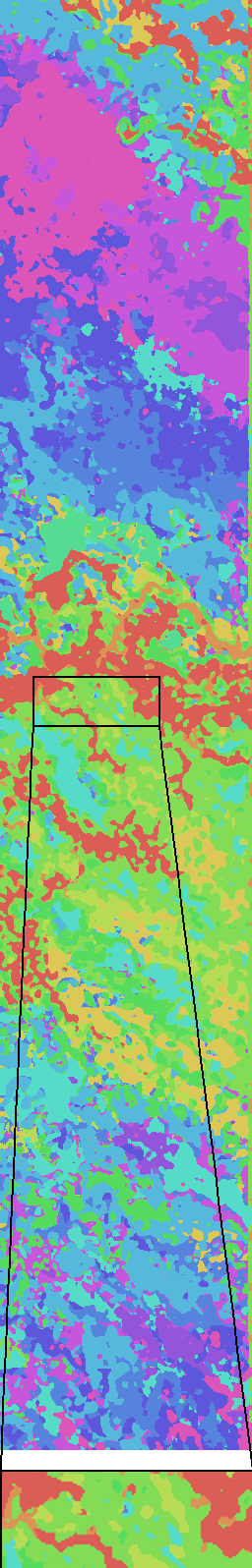}}\hfill
    \subfloat[]{\includegraphics[scale=0.13]{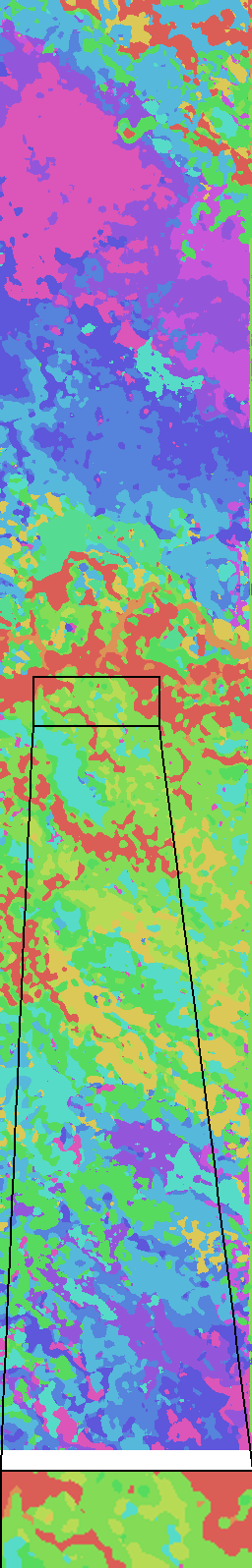}}\hfill
    \subfloat[]{\includegraphics[scale=0.13]{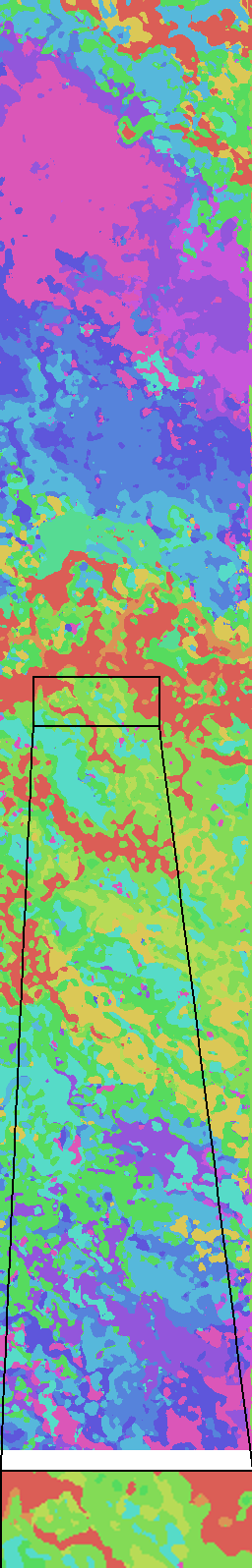}}\hfill
    \subfloat[]{\includegraphics[scale=0.13]{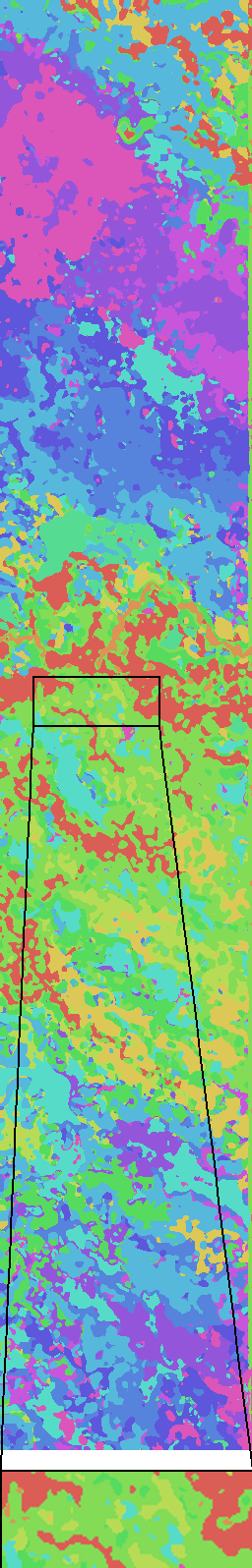}}\hfill
    \subfloat[]{\includegraphics[scale=0.13]{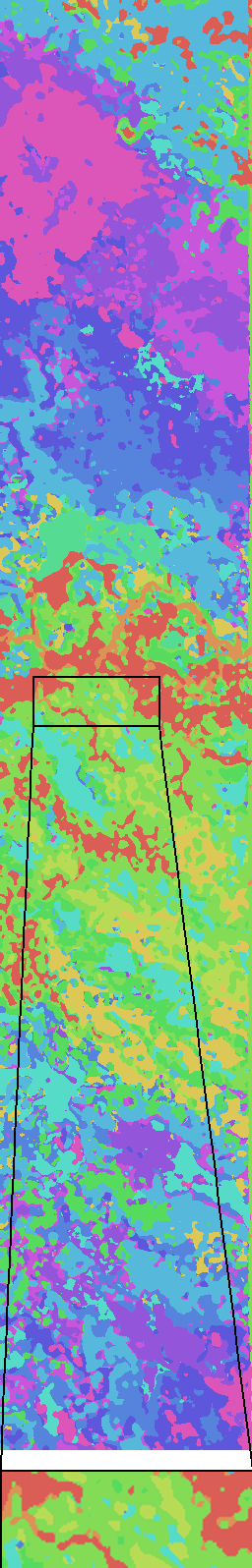}}\hfill
    \subfloat[]{\includegraphics[scale=0.13]{Figure_package/bot/pre_map/bot_gt.png}}\hfill
    \par
    \vspace{-0.3cm}
    \subfloat
    {\includegraphics[scale=0.5]{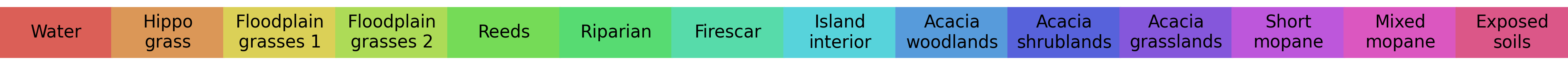}}\hfill

\caption{Prediction map on Botswana dataset. (a) CNN3D. (b) DFFN. (c) M3D-DCNN. (d) RSSAN. (e) SpectralFormer. (f) SSFTT. (g) GroupTransformer. (h) Proposed CNN-mixer. (i) Proposed SSA-mixer. (j) Proposed CSA-mixer. (k) Proposed SSA+CNN-mixer. (l) Proposed CSA+CNN-mixer. (m) Ground truth.}
\label{fig:bot_prediction_map}
\end{figure*}

\begin{figure*}
    \centering
    \subfloat[]{\includegraphics[scale=0.2]{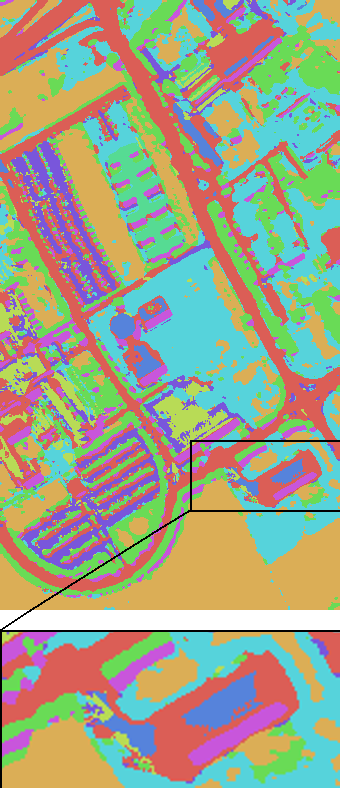}}\vspace{1pt}
    \subfloat[]{\includegraphics[scale=0.2]{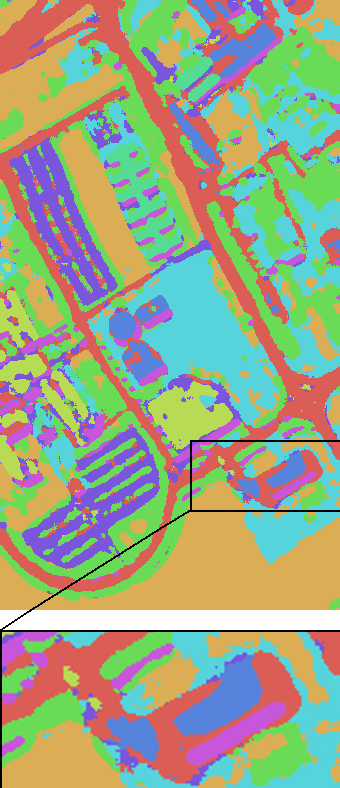}}\vspace{1pt}
    \subfloat[]{\includegraphics[scale=0.2]{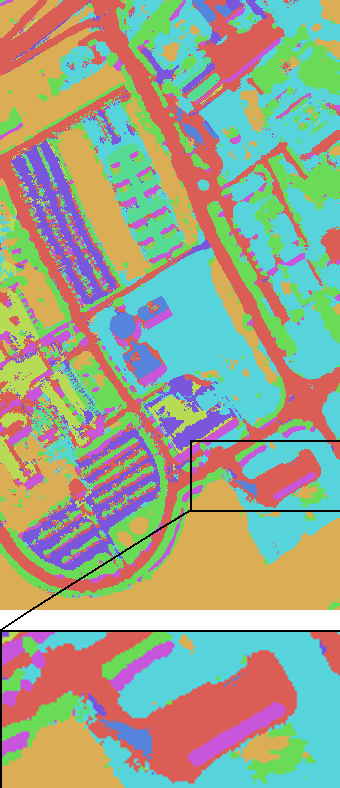}}\vspace{1pt}
    \subfloat[]{\includegraphics[scale=0.2]{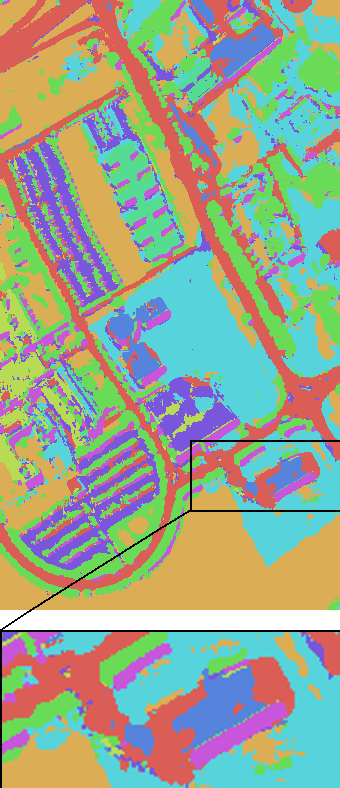}}\vspace{1pt}
    \subfloat[]{\includegraphics[scale=0.2]{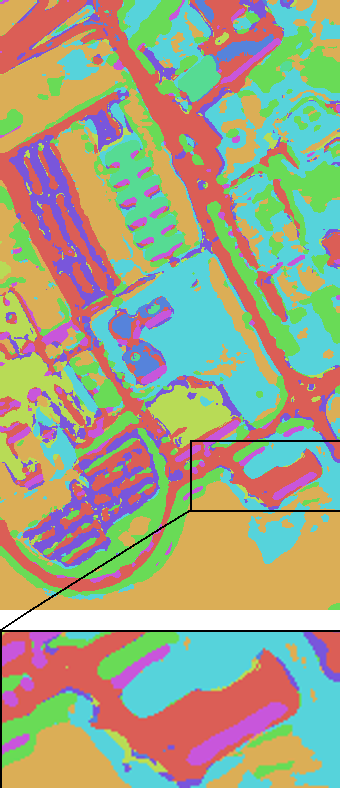}}\vspace{1pt}
    \subfloat[]{\includegraphics[scale=0.2]{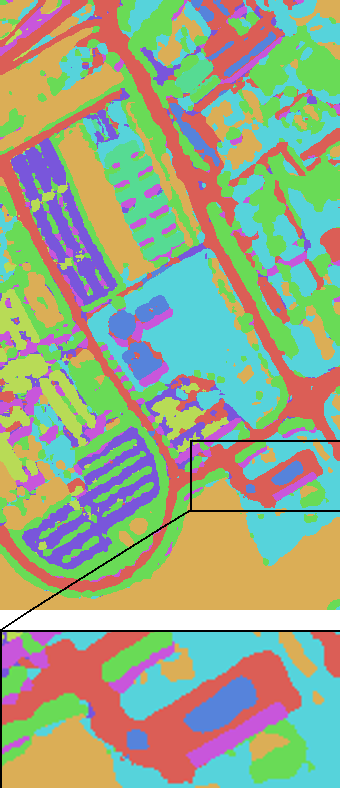}}\vspace{1pt}
    \subfloat[]{\includegraphics[scale=0.2]{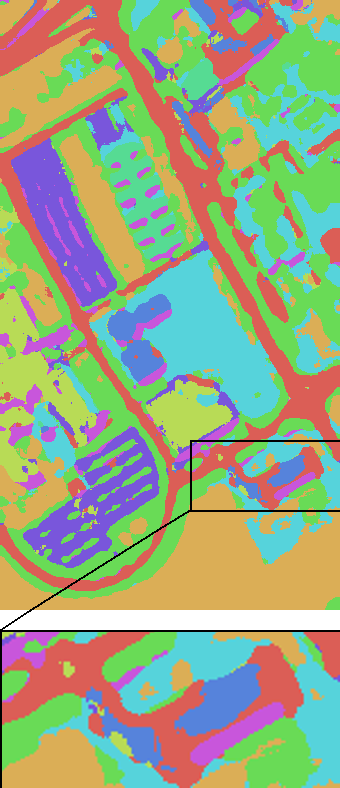}}\par
    \subfloat[]{\includegraphics[scale=0.2]{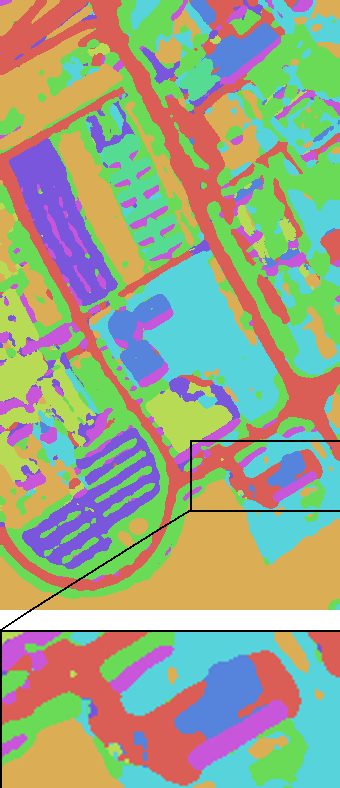}}\vspace{1pt}
    \subfloat[]{\includegraphics[scale=0.2]{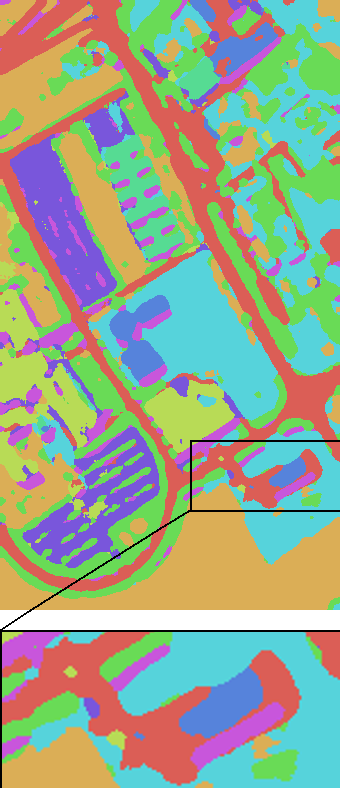}}\vspace{1pt}
    \subfloat[]{\includegraphics[scale=0.2]{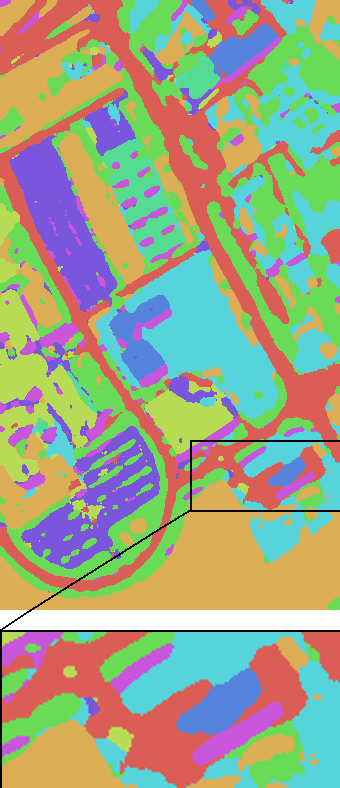}}\vspace{1pt}
    \subfloat[]{\includegraphics[scale=0.2]{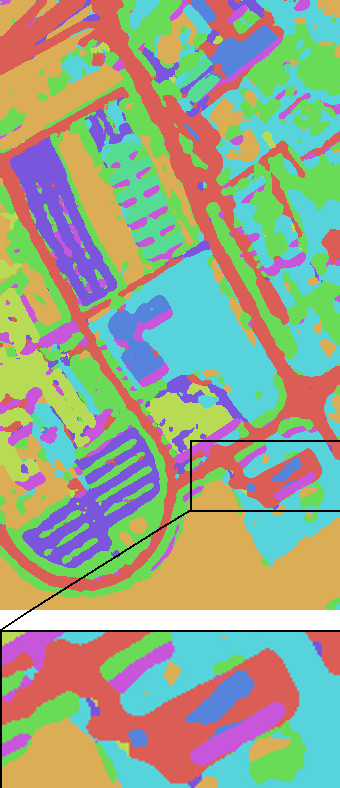}}\vspace{1pt}
    \subfloat[]{\includegraphics[scale=0.2]{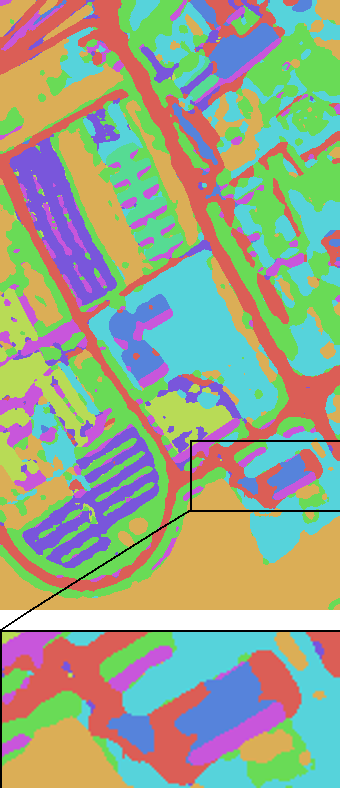}}\vspace{1pt}
    \subfloat[]{\includegraphics[scale=0.2]{Figure_package/pu/pre_map/pu_gt.png}}
    \par
    \vspace{-0.3cm}
    \subfloat
    {\includegraphics[scale=0.5]{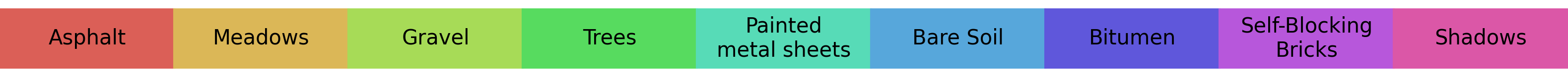}}

\caption{Prediction map on Pavia University dataset. (a) CNN3D. (b) DFFN. (c) M3D-DCNN. (d) RSSAN. (e) SpectralFormer. (f) SSFTT. (g) GroupTransformer. (h) Proposed CNN-mixer. (i) Proposed SSA-mixer. (j) Proposed CSA-mixer. (k) Proposed SSA+CNN-mixer. (l) Proposed CSA+CNN-mixer. (m) Ground truth.}
\label{fig:pu_prediction_map}
\end{figure*}

1) Analysis of classification performance: The mean and standard deviation of each criterion index across the three datasets are presented in Table~\ref{HU_classification}-\ref{PU_classification}. The highest value is highlighted in bold. 

The first comparison experiment was conducted on the Houston 2013 dataset. Table~\ref{HU_classification} reports the prediction results on the test dataset in terms of OA, AA, $\kappa$, and the accuracy of each class. 
Among the four CNN-based models, DFFN stands out with an OA of 97.59\%, marking a 3.18\% increase in performance compared to the CNN3D algorithm, which has the lowest OA in this group. As for the three classical vision Transformer algorithms, SpectralFormer, SSFTT, and GroupTransformer have OA values of 91.99\%, 98.47\%, and 98.38\%, respectively. In comparison to the SpectralFormer algorithm based on the vanilla vision Transformer, the latter two show significant improvements in classification accuracy. Utilizing the unified hierarchical Transformer architecture proposed in this paper, five mixer-based HSI classification models demonstrated exceptional OA, ranging from 98.48\% to 98.68\%. On this dataset, the model built upon the CSA+CNN-mixer outperforms other classical CNN and vision Transformer models, with accuracy improvements of 0.21\% and 6.69\%, respectively.
The corresponding prediction map is shown in Fig.~\ref{fig:hu_prediction_map}. In the prediction maps for (a), (c), (d), and (e), there is a comparatively higher occurrence of spurious anomalies. The prediction map outcomes for (h) through (l), which represent the five models developed using a unified architecture, show a remarkable similarity.

The second comparative experiment was carried out using the Botswana dataset. The prediction results are listed in Table~\ref{bot_classification}. On this dataset, the CNN3D algorithm has a significant decrease in accuracy, differing by at least 3.46\% compared to the other three common CNN algorithms. Among these, DFFN is the top-performing CNN algorithm, attaining a maximum accuracy of 99.28\%. Similar to the results on the Houston 2013 dataset, among the three typical vision Transformer algorithms, SpectralFormer achieved a lower accuracy at 95.24\%, in contrast to GroupTransformer achieved an OA value of 99.53\%. Among the five models employing different mixers, the one utilizing the CSA-mixer outperforms the rest, achieving an OA of 99.74\%. The SSA+CNN-mixer-based method performs the worst. However, its OA value is only 0.02\% lower than that of the GroupTransformer.
The prediction map is depicted in Fig.~\ref{fig:bot_prediction_map}. The maps for (a), (c), (d), (e), and (f) demonstrate suboptimal performance. Upon exploring a magnified view, it is evident that (b) shows a higher number of errors in predicting 'Riparian' compared to the proposed models based on the unified vision Transformer architecture. This phenomenon matches the metrics provided in Table~\ref{bot_classification}.

The third comparative experiment utilized the Pavia dataset, and Table~\ref{PU_classification} displays the prediction results obtained by different methods. 
Among the four prevalent CNN methods, DFFN obtains an OA of 93.53\%, which is comparatively lower, while RSSAN distinguishes itself with a higher OA of 98.88\%. Within the three popular vision Transformer-based methods, SpectralFormer lags slightly behind the other two algorithms. Notably, in contrast to the previous two datasets, the SSFTT algorithm outperforms the other two popular vision Transformer methods, reaching an accuracy of 99.43\% and surpassing the GroupTransformer by 0.14\%. Meanwhile, the five HSI classification algorithms proposed in this paper consistently demonstrate the highest accuracy, ranging from 99.44\% to 99.55\%, with the CNN-mixer-based algorithm arriving at 99.55\%. Overall, the differences in OA among these five algorithms remain relatively small.
The prediction map is depicted in Fig.~\ref{fig:pu_prediction_map}. It illustrates that the five models built on the unified Transformer architecture demonstrate superior classification accuracy for the categories 'Gravel' and 'Bitumen'.

In summary, the five models utilizing the unified hierarchical vision Transformer architecture demonstrate the highest OA across three testbed datasets. Differences in accuracies achieved by the five algorithms are generally insignificant. This also implies that the performance of the proposed HSI classification models predominantly hinges on the \emph{unified architecture}, rather than the specific mixer modules that attracted attention in previous research.

\begin{figure*}
    \centering
    % \subfigbottomskip=2pt
    \subfloat[]{\includegraphics[trim=0 0.2cm 0 1cm, clip, scale=0.5]{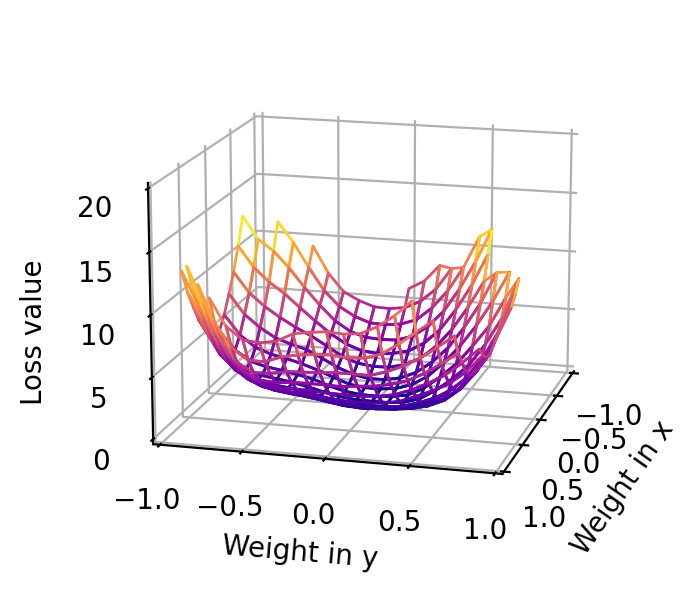}}
    \subfloat[]{\includegraphics[trim=0 0.2cm 0 1cm, clip, scale=0.5]{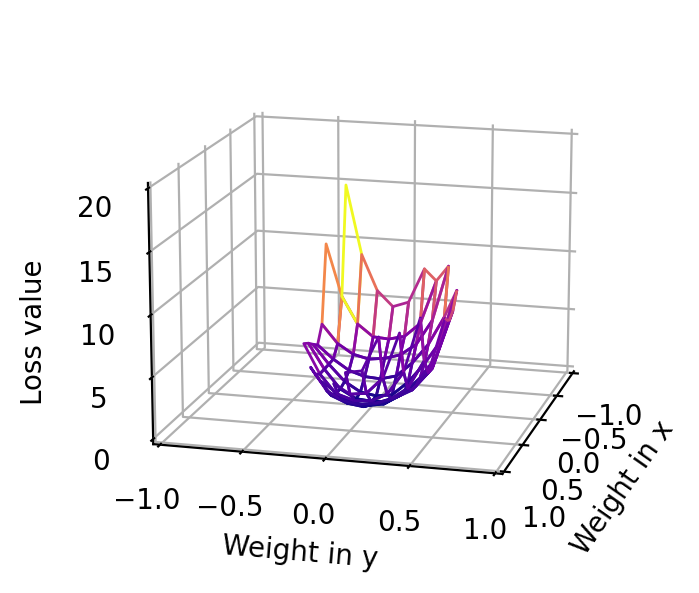}}
    \subfloat[]{\includegraphics[trim=0 0.2cm 0 1cm, clip, scale=0.5]{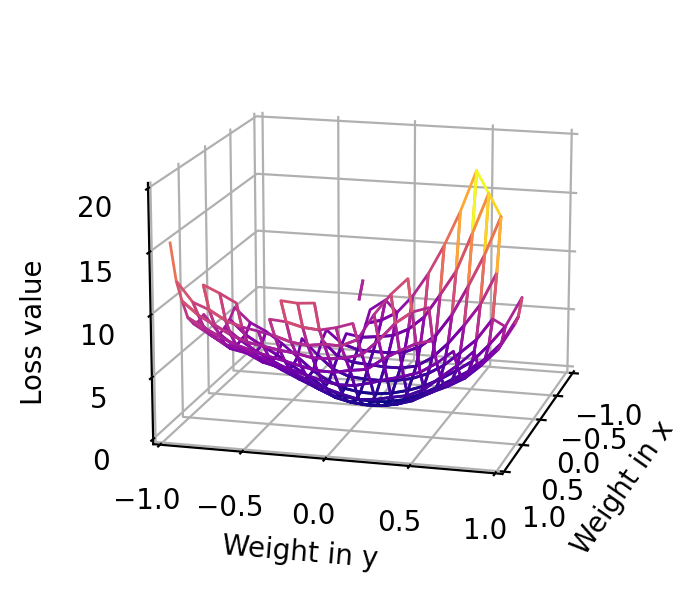}}
    \subfloat[]{\includegraphics[trim=0 0.2cm 0 1cm, clip, scale=0.5]{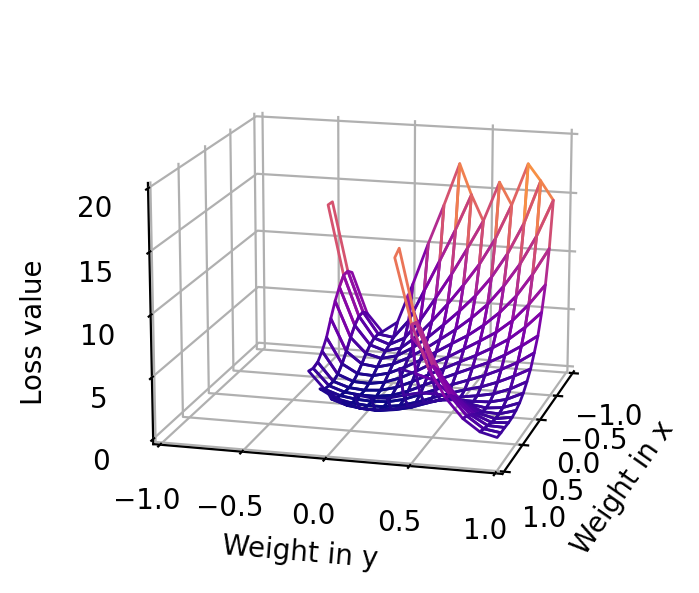}}\par
    \vspace{-0.38cm}
    % \quad
    \subfloat[]{\includegraphics[trim=0 0.2cm 0 1cm, clip, scale=0.5]{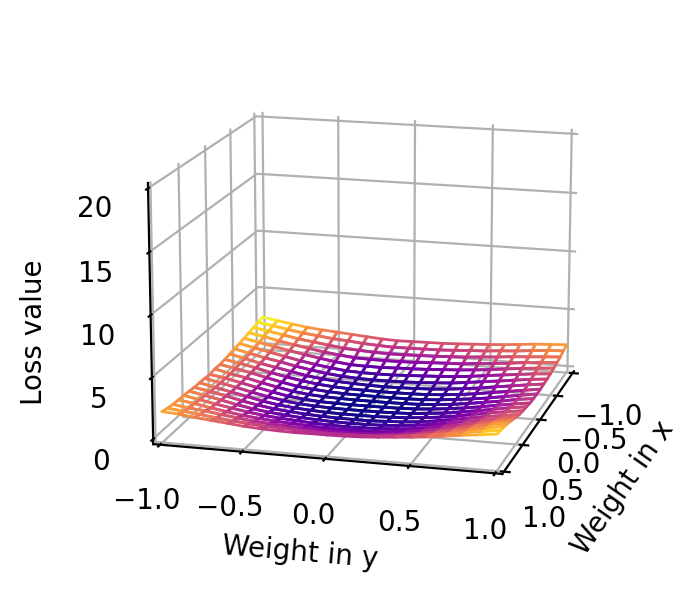}}
    \subfloat[]{\includegraphics[trim=0 0.2cm 0 1cm, clip, scale=0.5]{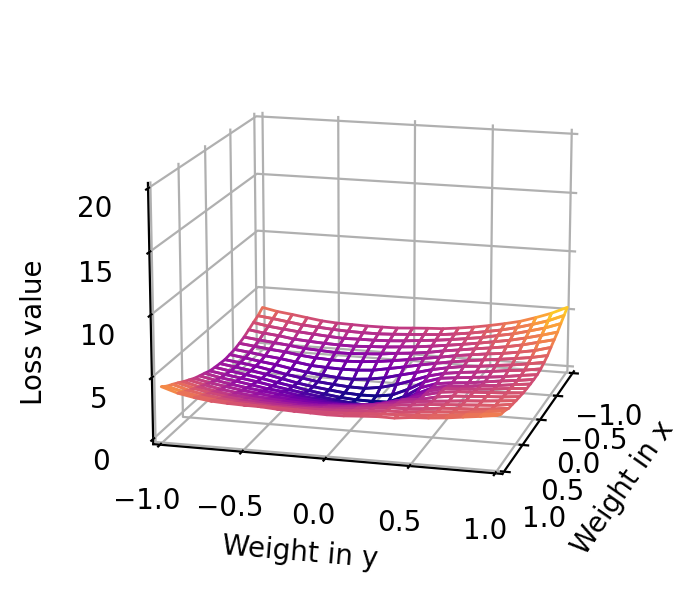}}
    \subfloat[]{\includegraphics[trim=0 0.2cm 0 1cm, clip, scale=0.5]{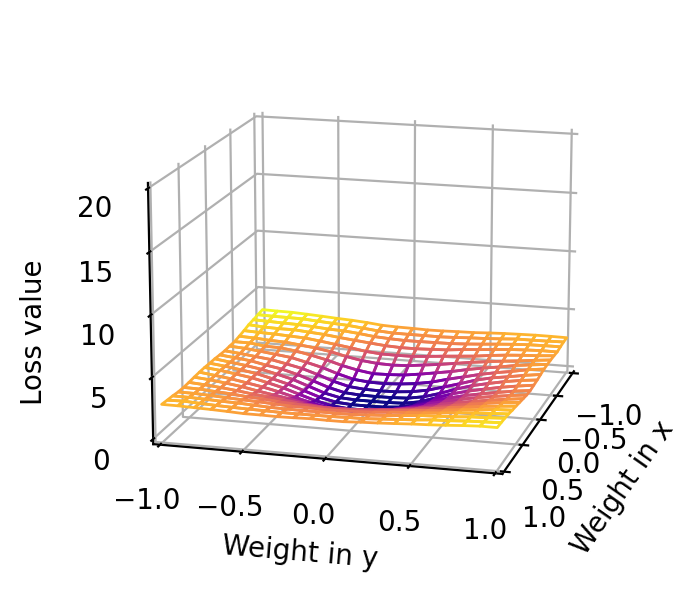}}
    \subfloat[]{\includegraphics[trim=0 0.2cm 0 1cm, clip, scale=0.5]{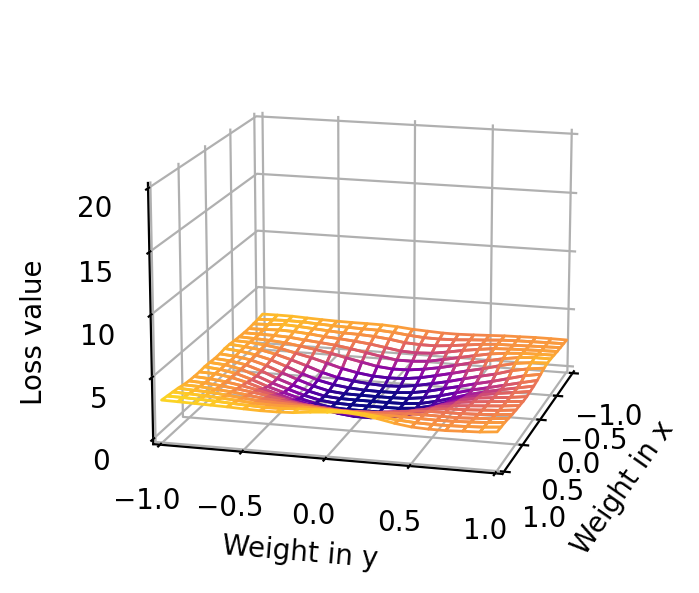}}\par
    \vspace{-0.38cm}
    \subfloat[]{\includegraphics[trim=0 0.2cm 0 1cm, clip, scale=0.5]{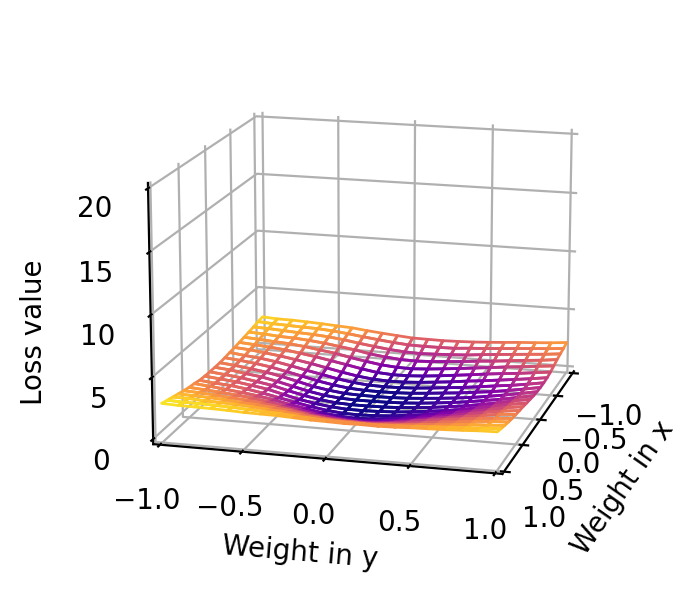}}
    \subfloat[]{\includegraphics[trim=0 0.2cm 0 1cm, clip, scale=0.5]{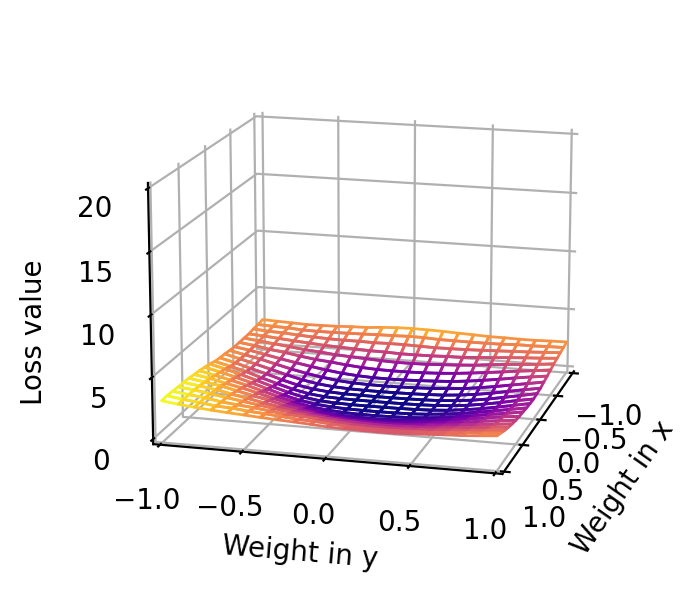}}
    \subfloat[]{\includegraphics[trim=0 0.2cm 0 1cm, clip, scale=0.5]{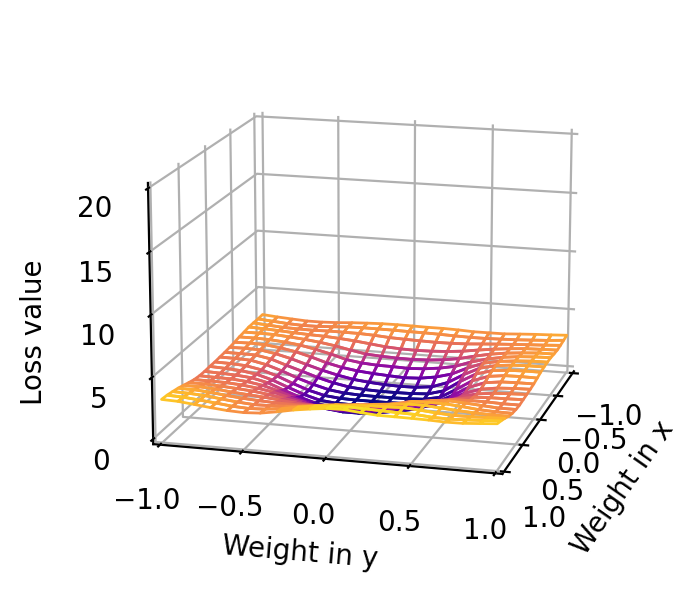}}
    \subfloat[]{\includegraphics[trim=0 0.2cm 0 1cm, clip, scale=0.5]{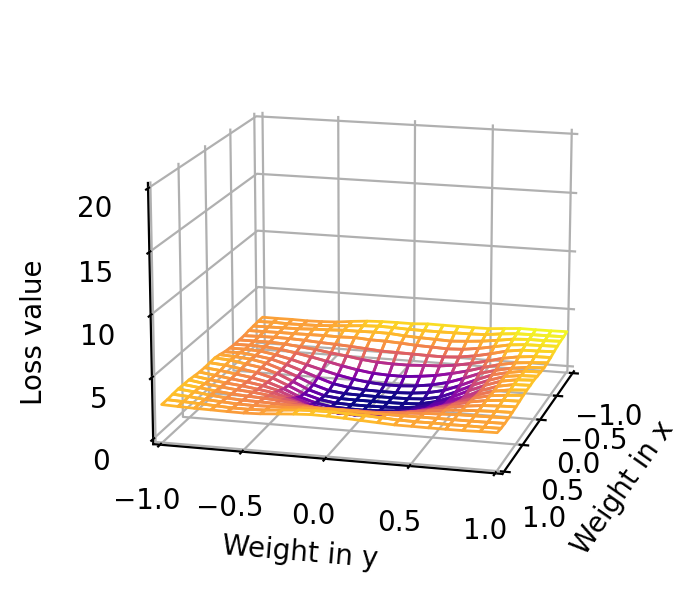}}
    \par 

\caption{Disturbance robustness visualization based on Houston 2013 dataset. (a) CNN3D. (b) DFFN. (c) M3D-DCNN. (d) RSSAN. (e) SpectralFormer. (f) SSFTT. (g) GroupTransformer. (h) Proposed CNN-mixer. (i) Proposed SSA-mixer. (j) Proposed CSA-mixer. (k) Proposed SSA+CNN-mixer. (l) Proposed CSA+CNN-mixer.}
\label{fig:hu_disturbance_robustness}
\end{figure*}

\begin{figure*}
    \centering
    \subfloat[]{\includegraphics[trim=0 0.2cm 0 1cm, clip, scale=0.5]{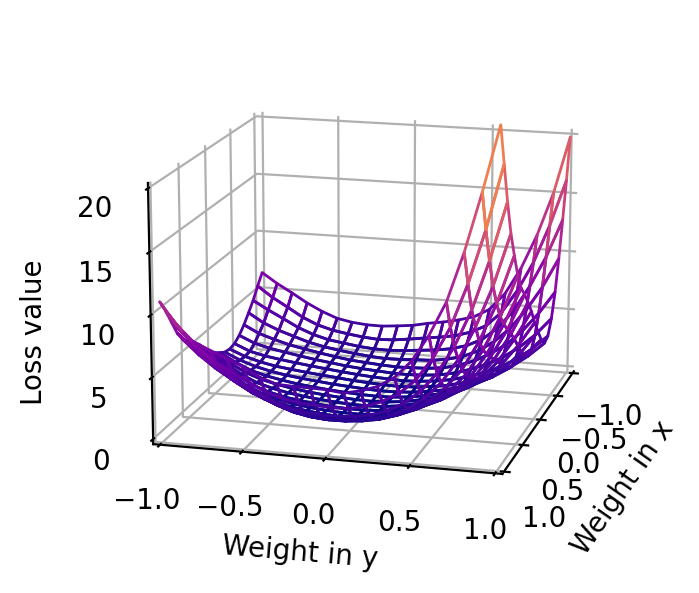}}
    \subfloat[]{\includegraphics[trim=0 0.2cm 0 1cm, clip, scale=0.5]{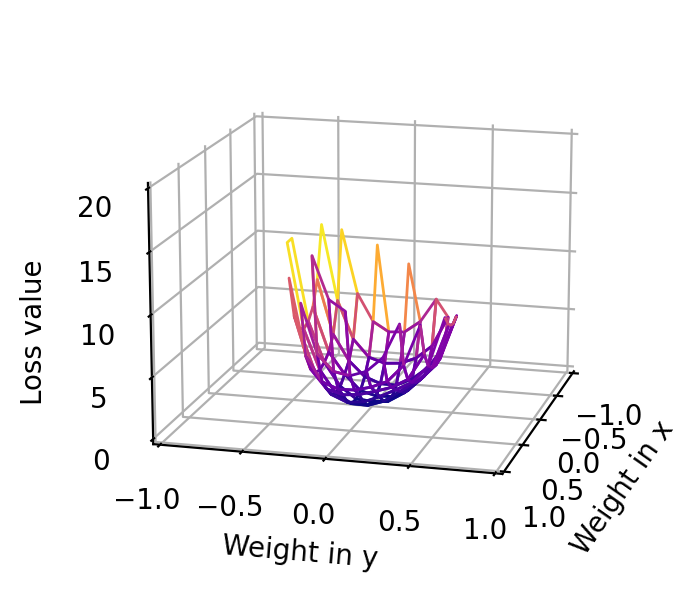}}
    \subfloat[]{\includegraphics[trim=0 0.2cm 0 1cm, clip, scale=0.5]{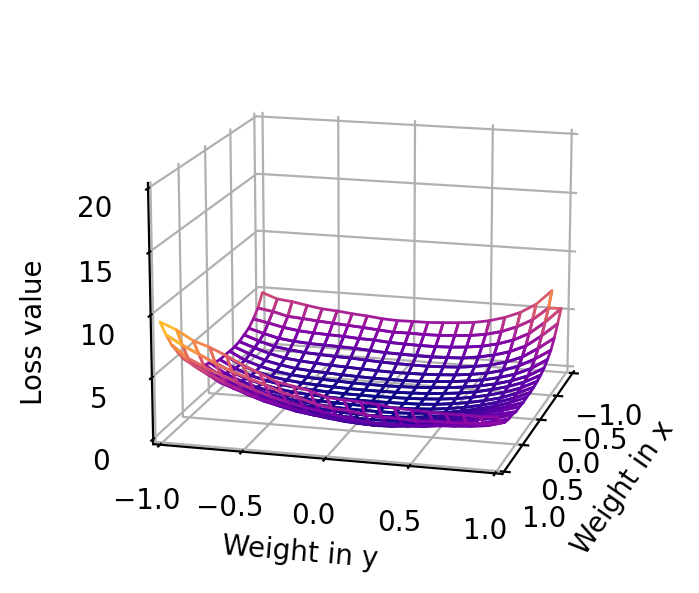}}
    \subfloat[]{\includegraphics[trim=0 0.2cm 0 1cm, clip, scale=0.5]{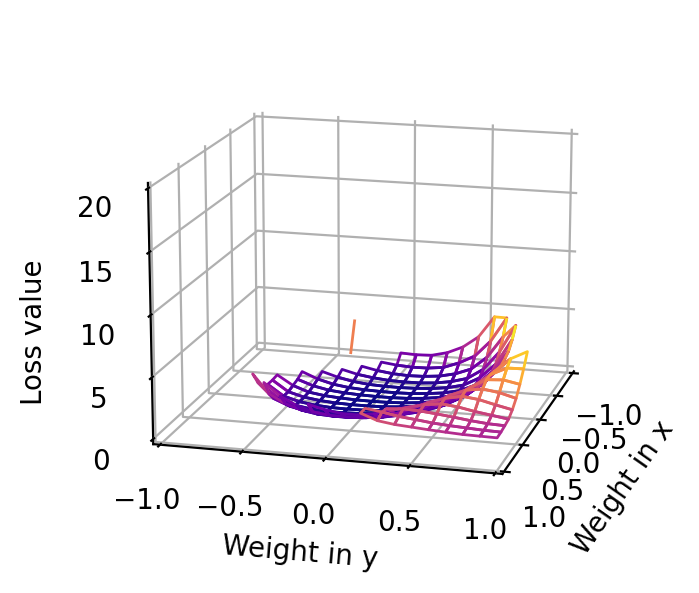}}\par
    \vspace{-0.38cm}
    \subfloat[]{\includegraphics[trim=0 0.2cm 0 1cm, clip, scale=0.5]{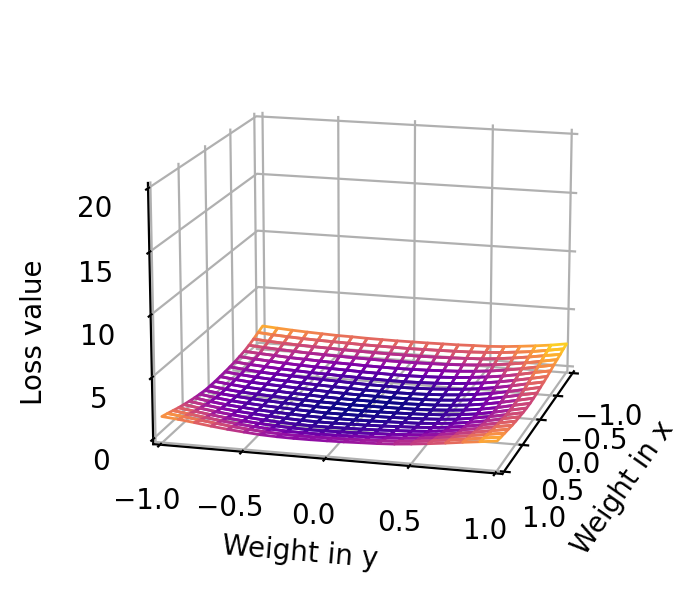}}
    \subfloat[]{\includegraphics[trim=0 0.2cm 0 1cm, clip, scale=0.5]{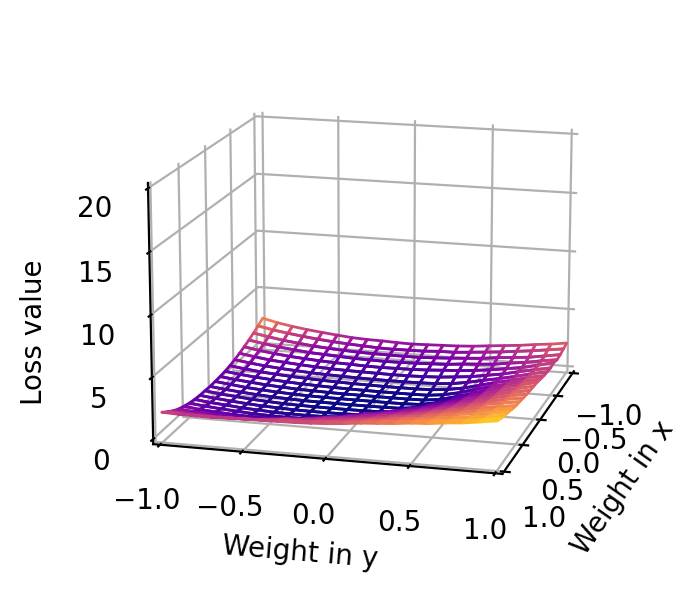}} 
    \subfloat[]{\includegraphics[trim=0 0.2cm 0 1cm, clip, scale=0.5]{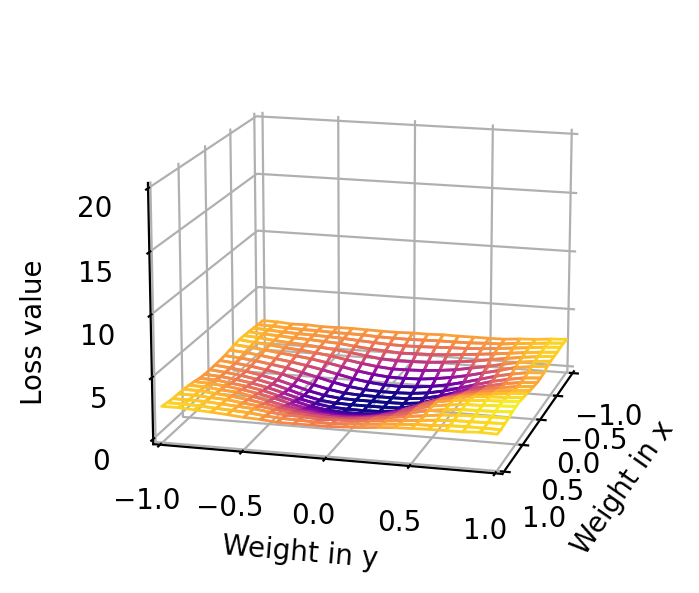}}
    \subfloat[]{\includegraphics[trim=0 0.2cm 0 1cm, clip, scale=0.5]{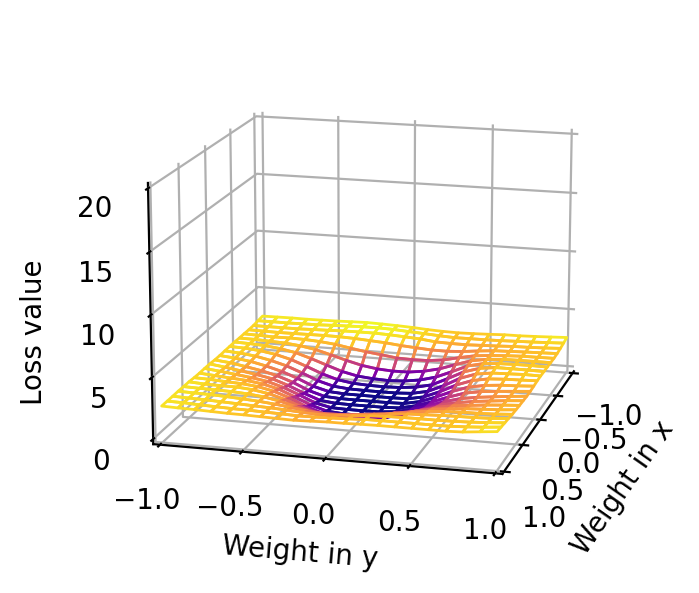}}\par
    \vspace{-0.38cm} 
    \subfloat[]{\includegraphics[trim=0 0.2cm 0 1cm, clip, scale=0.5]{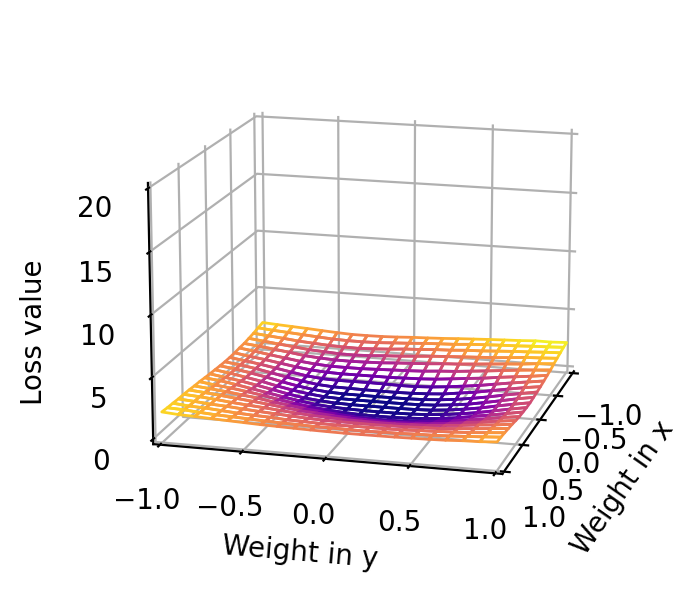}}
    \subfloat[]{\includegraphics[trim=0 0.2cm 0 1cm, clip, scale=0.5]{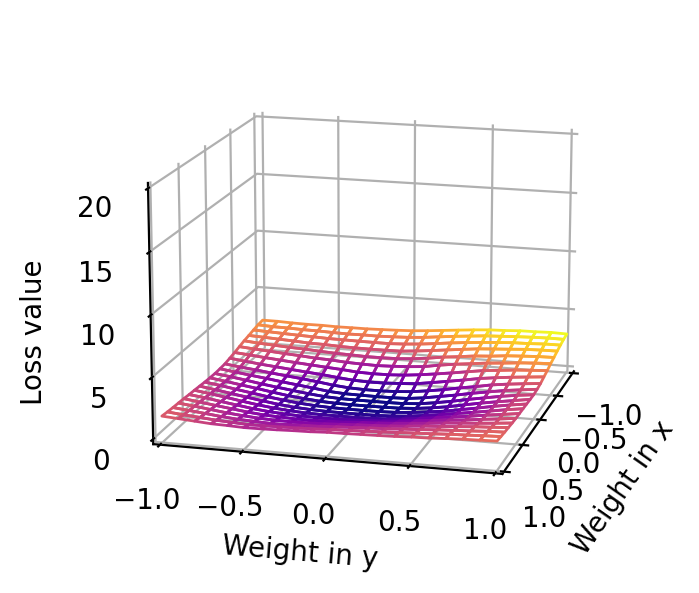}}
    \subfloat[]{\includegraphics[trim=0 0.2cm 0 1cm, clip, scale=0.5]{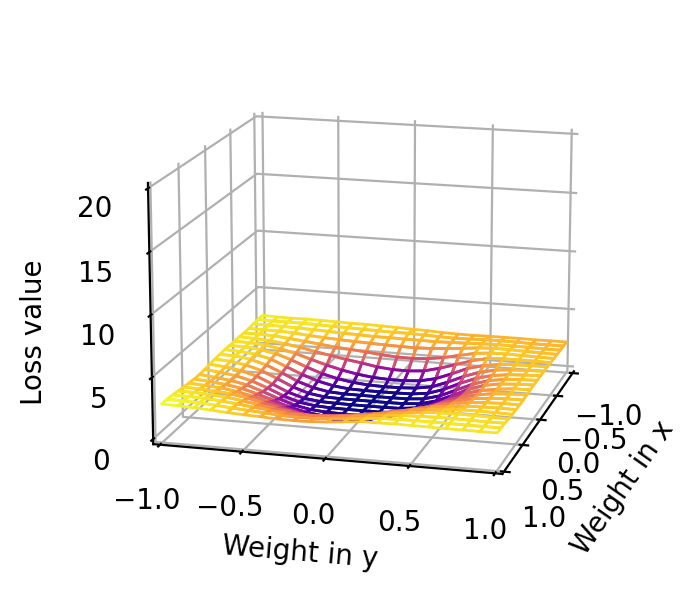}}
    \subfloat[]{\includegraphics[trim=0 0.2cm 0 1cm, clip, scale=0.5]{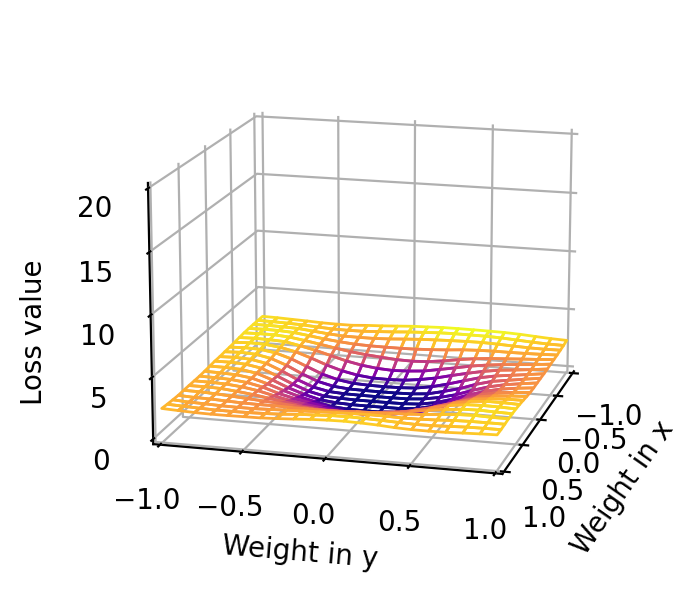}}
    \par 

\caption{Disturbance robustness visualization based on Botswana dataset. (a) CNN3D. (b) DFFN. (c) M3D-DCNN. (d) RSSAN. (e) SpectralFormer. (f) SSFTT. (g) GroupTransformer. (h) Proposed CNN-mixer. (i) Proposed SSA-mixer. (j) Proposed CSA-mixer. (k) Proposed SSA+CNN-mixer. (l) Proposed CSA+CNN-mixer.}
\label{fig:bot_disturbance_robustness}
\end{figure*}

2) Analysis of the training process: Based on traditional evaluation methods such as OA metric and prediction maps, the following two challenges remain difficult to address comprehensively:
(1) Is MSA indeed the pivotal module in vision Transformers for enhancing HSI classification?
(2) What are the critical differences in model characteristics between vision Transformer-based models and CNN models regarding training on hyperspectral datasets? To address these questions, as noted previously, the macro and micro-level characteristics of the models during the training process were investigated.

\begin{figure*}
    \centering
    \subfloat[]{\includegraphics[trim=0 0.2cm 0 1cm, clip, scale=0.5]{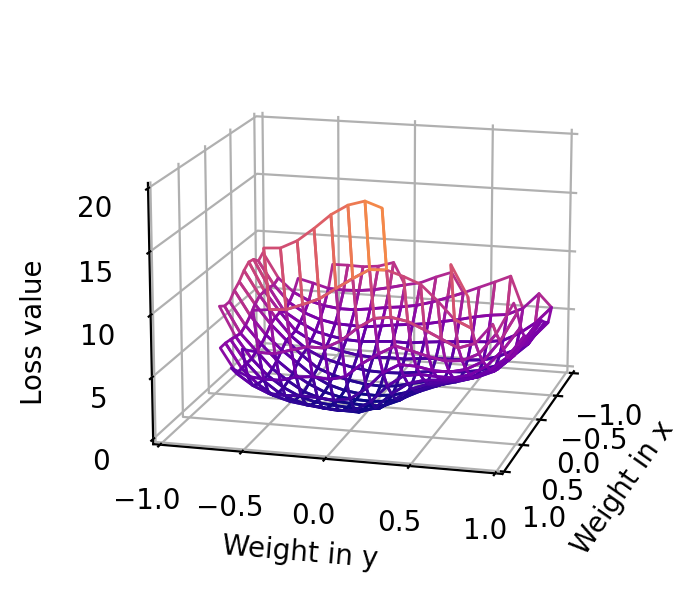}}
    \subfloat[]{\includegraphics[trim=0 0.2cm 0 1cm, clip, scale=0.5]{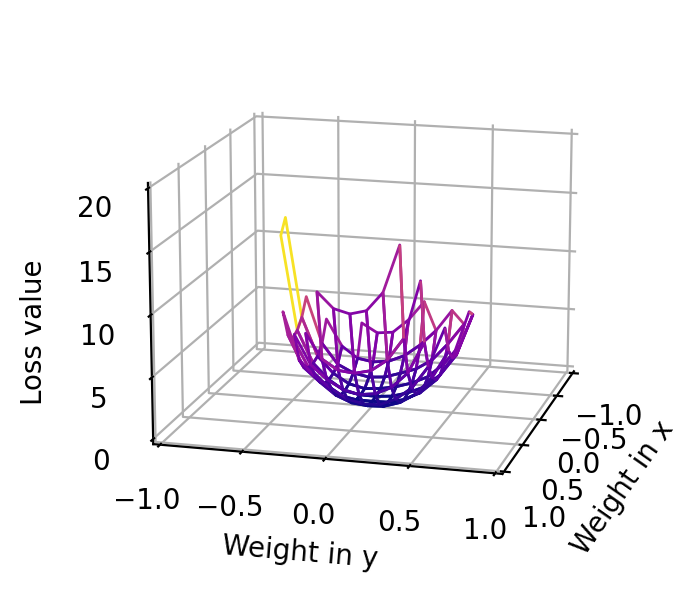}}
    \subfloat[]{\includegraphics[trim=0 0.2cm 0 1cm, clip, scale=0.5]{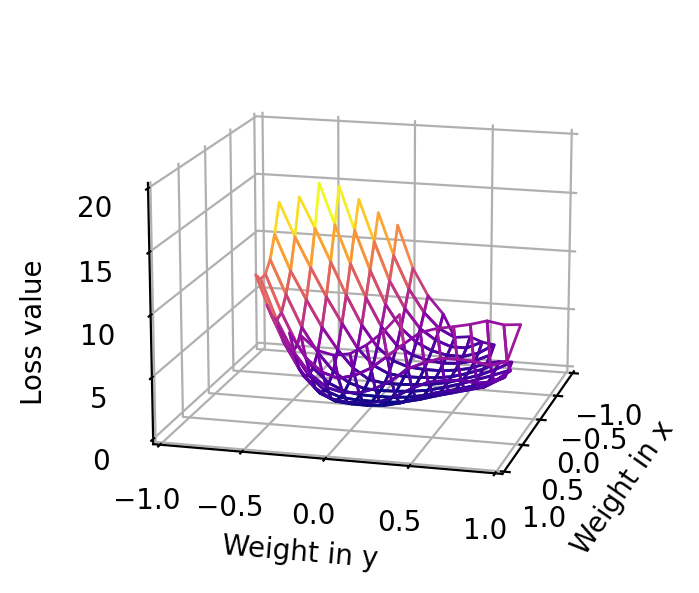}}
    \subfloat[]{\includegraphics[trim=0 0.2cm 0 1cm, clip, scale=0.5]{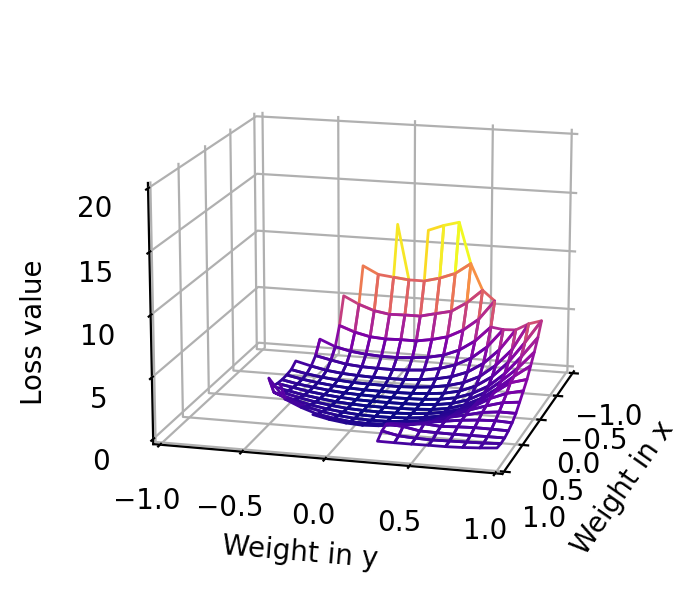}}\par
    \vspace{-0.38cm}
    \subfloat[]{\includegraphics[trim=0 0.2cm 0 1cm, clip, scale=0.5]{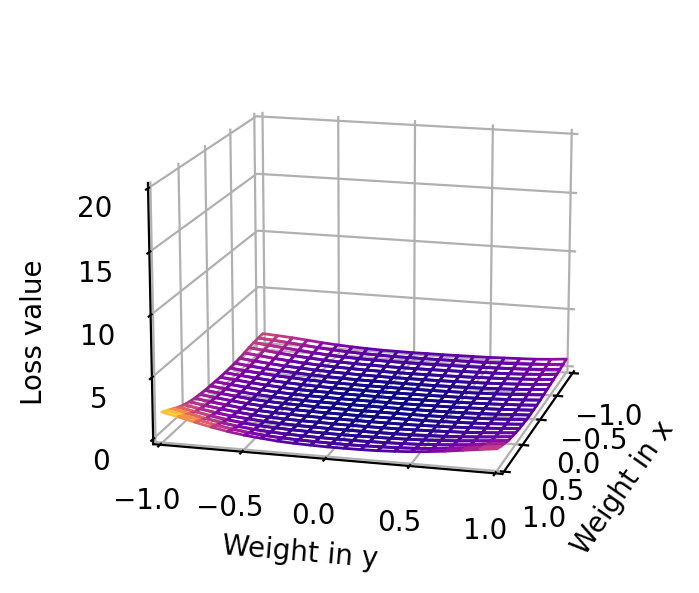}}
    \subfloat[]{\includegraphics[trim=0 0.2cm 0 1cm, clip, scale=0.5]{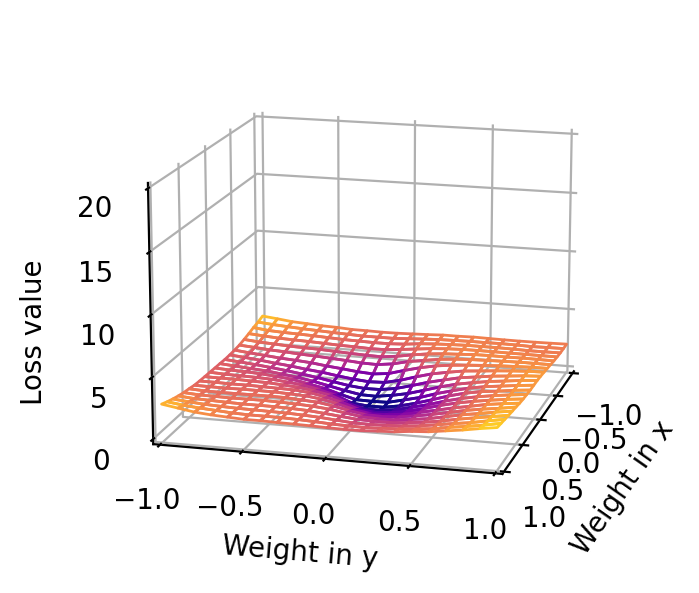}}
    \subfloat[]{\includegraphics[trim=0 0.2cm 0 1cm, clip, scale=0.5]{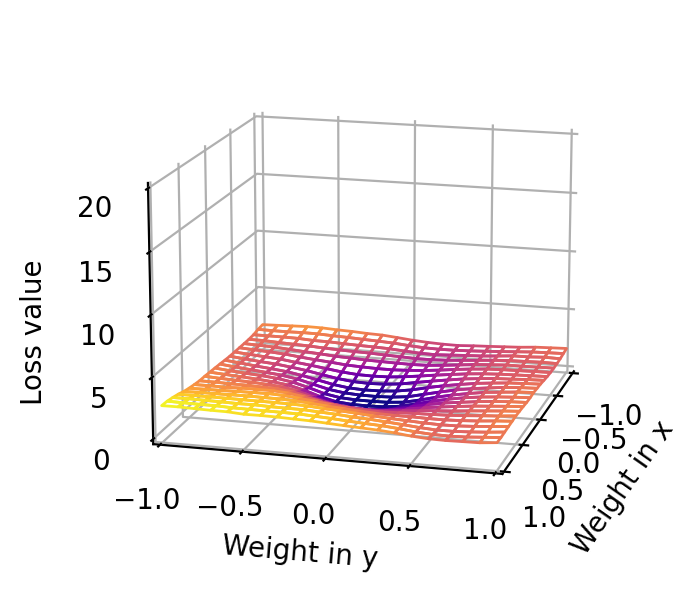}}
    \subfloat[]{\includegraphics[trim=0 0.2cm 0 1cm, clip, scale=0.5]{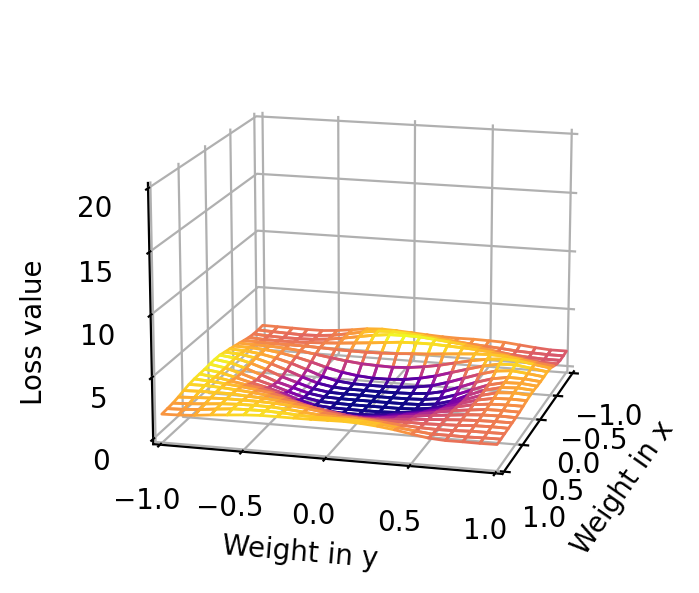}}\par
    \vspace{-0.38cm}
    \subfloat[]{\includegraphics[trim=0 0.2cm 0 1cm, clip, scale=0.5]{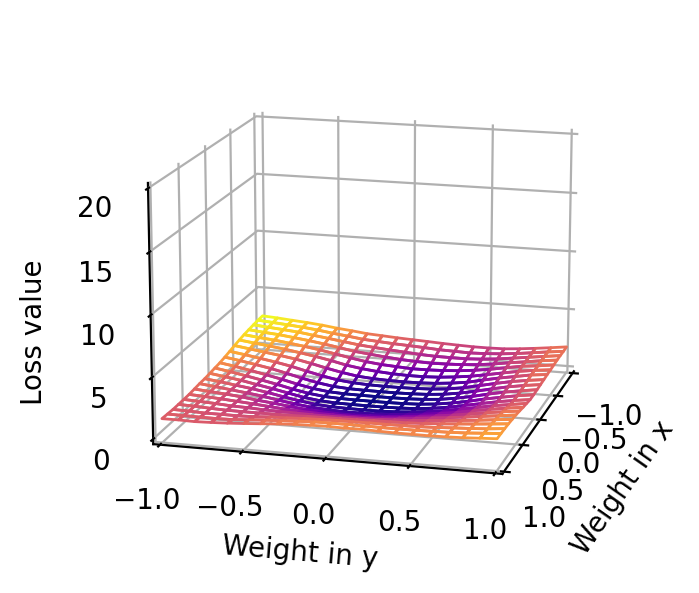}}
    \subfloat[]{\includegraphics[trim=0 0.2cm 0 1cm, clip, scale=0.5]{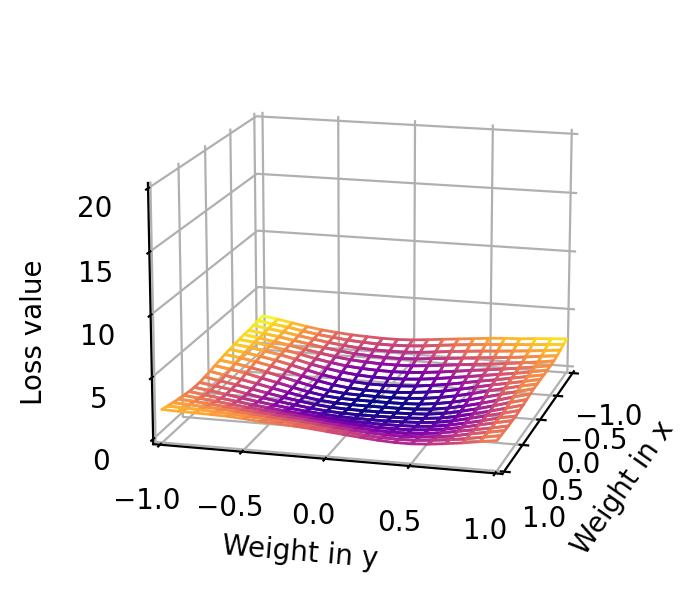}}
    \subfloat[]{\includegraphics[trim=0 0.2cm 0 1cm, clip, scale=0.5]{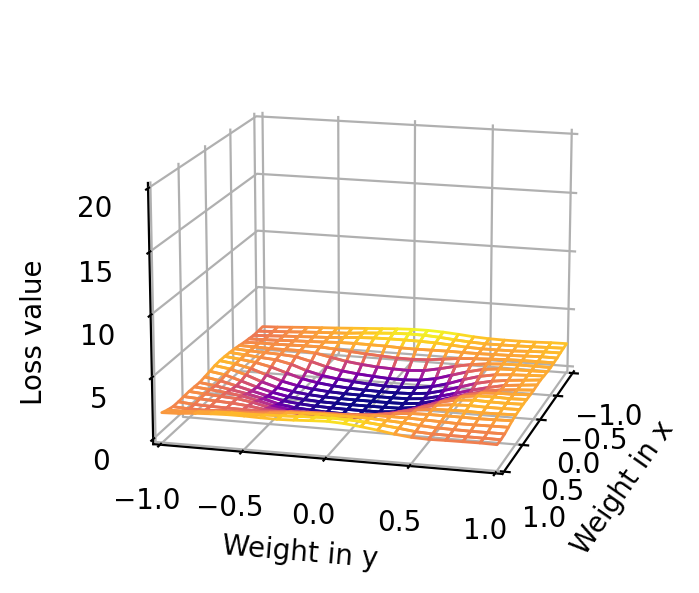}}
    \subfloat[]{\includegraphics[trim=0 0.2cm 0 1cm, clip, scale=0.5]{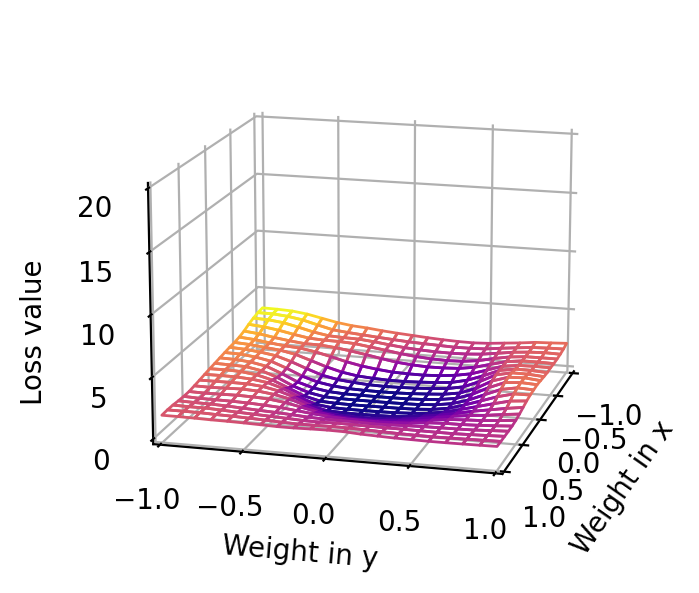}}
    \par 

\caption{Disturbance robustness visualization based on Pavia University dataset. (a) CNN3D. (b) DFFN. (c) M3D-DCNN. (d) RSSAN. (e) SpectralFormer. (f) SSFTT. (g) GroupTransformer. (h) Proposed CNN-mixer. (i) Proposed SSA-mixer. (j) Proposed CSA-mixer. (k) Proposed SSA+CNN-mixer. (l) Proposed CSA+CNN-mixer.}
\label{fig:pu_disturbance_robustness}
\end{figure*}

Fig. \ref{fig:hu_disturbance_robustness}, \ref{fig:bot_disturbance_robustness}, and \ref{fig:pu_disturbance_robustness} contain plots of three-dimensional surfaces representing the loss values for different models, following the introduction of disturbance with varying amplitudes into the 'best' pretrained weights on the three datasets. (a) - (d) depict loss surface contours based on four typical CNN algorithms. From a macroscopic perspective, it is evident that introducing two directional disturbances to the 'best' pretrained weights across the three datasets results in a noticeable spike in loss values calculated with these perturbed weights, thereby accentuating the contour of the surface. It is important to note that the HSI classification model based on DFFN, when exposed to substantial disturbances, produces excessively high loss values that surpass our predefined threshold of 100. This leads to scenarios, as illustrated in (b), where loss values become undefined as the values of the $x$ and $y$ axes approach 1. This suggests that the stability of the DFFN model is the least resilient to disturbances in the 'best' pretrained weight. (e) - (g) depict three-dimensional loss surface contours corresponding to three representative vision Transformer models, while (i) - (l) illustrate three-dimensional loss surface contours for the five algorithms proposed in this paper. The figures clearly show that the three-dimensional surface contours based on vision Transformer algorithms are notably smoother when compared to those based on CNN algorithms. However, in the case of (e), it can be observed that after introducing disturbances of varying magnitudes, the loss value along the $z$-axis for SpectralFormer changes at a notably lower rate. This suggests that the model exhibits a weak response to disturbances when starting from the 'best' pretrained weight. Interestingly, this phenomenon may not favor the model's ability to converge towards an optimal solution during training. This consequence may be attributed to the newly introduced structures such as group-wise spectral embedding and cross-layer adaptive fusion. Furthermore, by investigating (g) - (f), it is apparent that as the magnitude of disturbances increases, the model's loss value initially experiences a slight increase before eventually reaching saturation. An \emph{ideal model} should demonstrate the characteristic of a moderate increase in loss value when disturbances are applied to the optimal training weights. It's noteworthy that (h) corresponds to the proposed algorithm based on the CNN-mixer. This model lacks the MSA module, yet the shape of its loss surface contours across the three datasets differs significantly from those corresponding to CNN models. In contrast, the loss surface contours for vision Transformer models constructed with the five different mixers exhibit remarkably similar shapes. This further indicates that the distinctive characteristics of vision Transformer models in HSI classification primarily are derived from their unified hierarchical Transformer architecture rather than the specific mixer modules. It also suggests that the MSA module is not the fundamental reason for the differences between CNN and vision Transformer models.

\begin{figure*}
    \centering
    \subfloat[]{\includegraphics[trim=0 0 0 0, clip, scale=0.4]{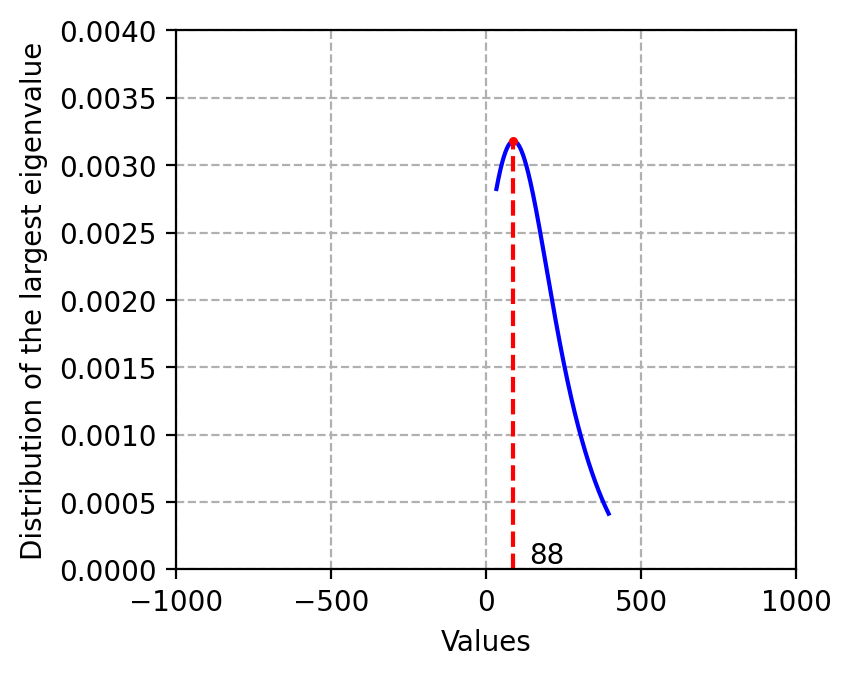}}
    \subfloat[]{\includegraphics[trim=0 0 0 0, clip, scale=0.4]{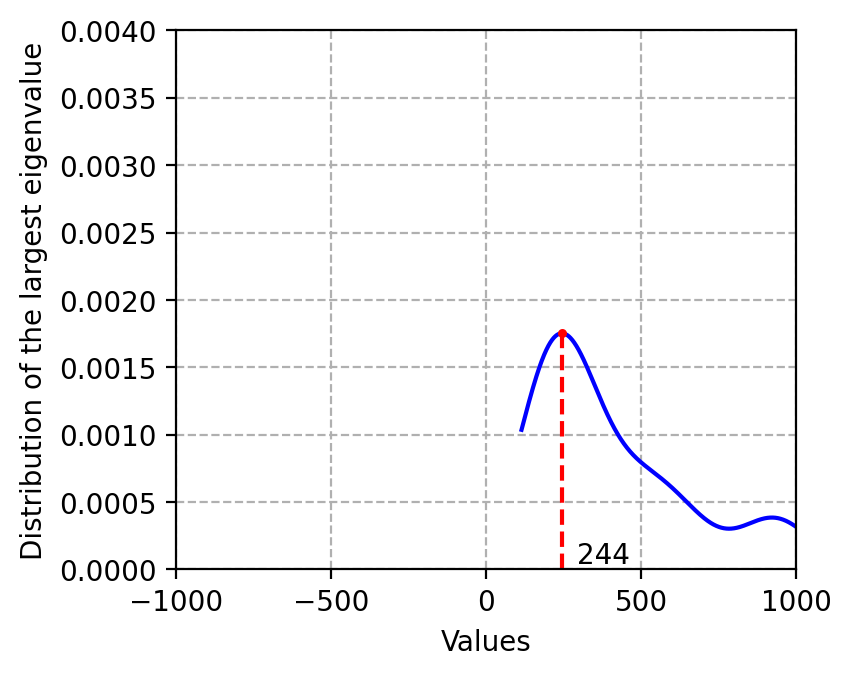}}
    \subfloat[]{\includegraphics[trim=0 0 0 0, clip, scale=0.4]
    {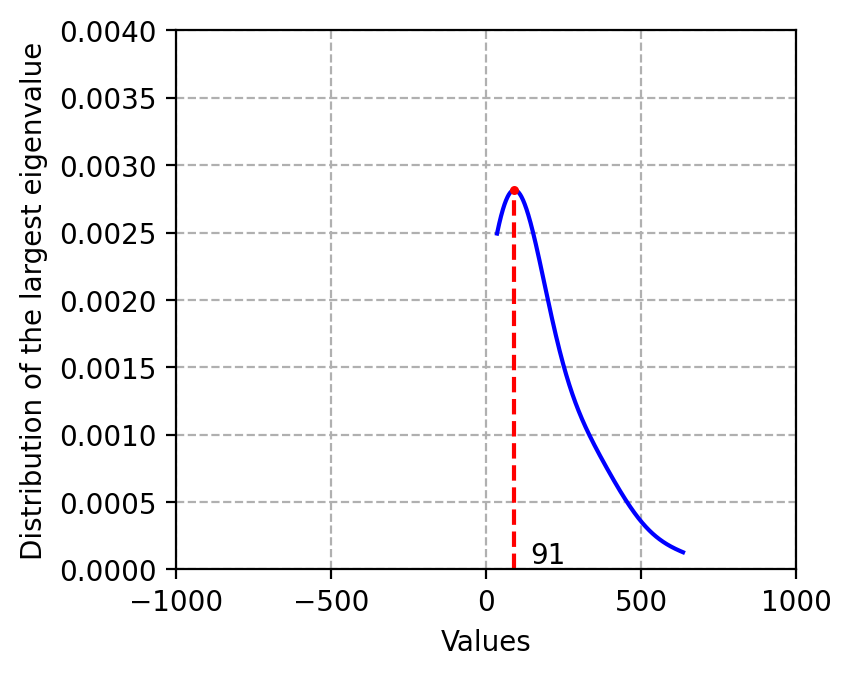}}
    \subfloat[]{\includegraphics[trim=0 0 0 0, clip, scale=0.4]{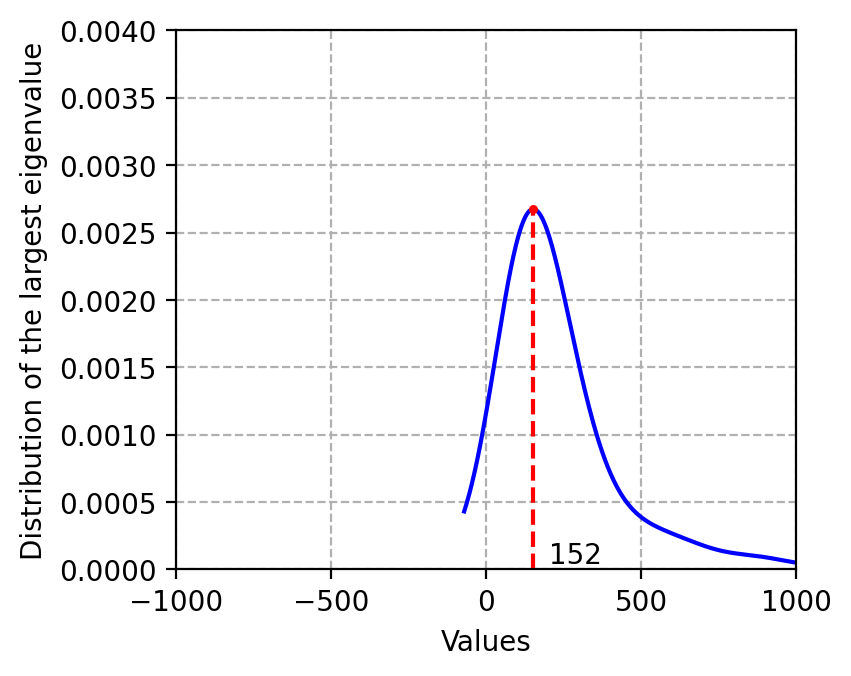}}\par
    \vspace{-0.38cm} 
    \subfloat[]{\includegraphics[trim=0 0 0 0, clip, scale=0.4]{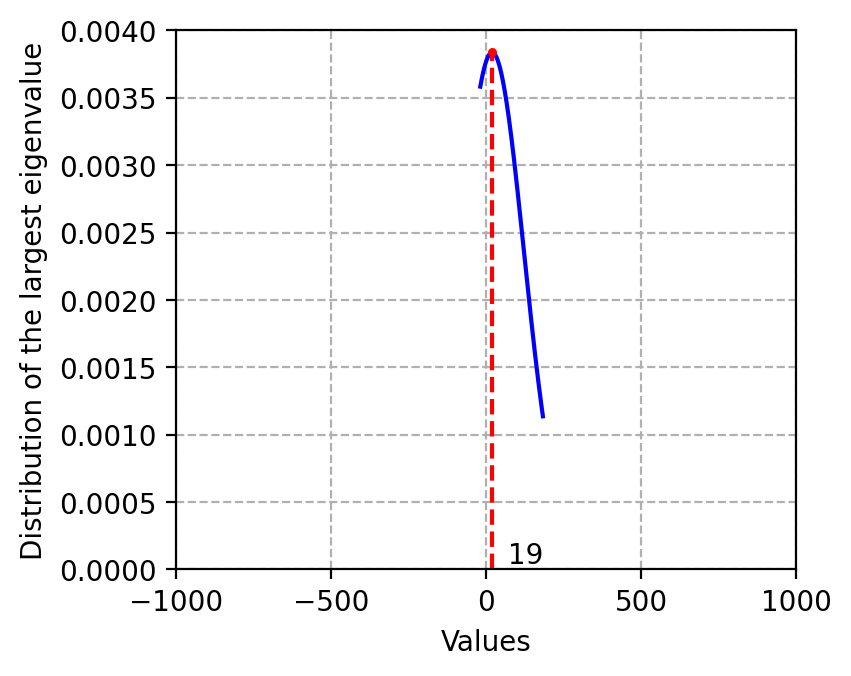}}
    \subfloat[]{\includegraphics[trim=0 0 0 0, clip, scale=0.4]{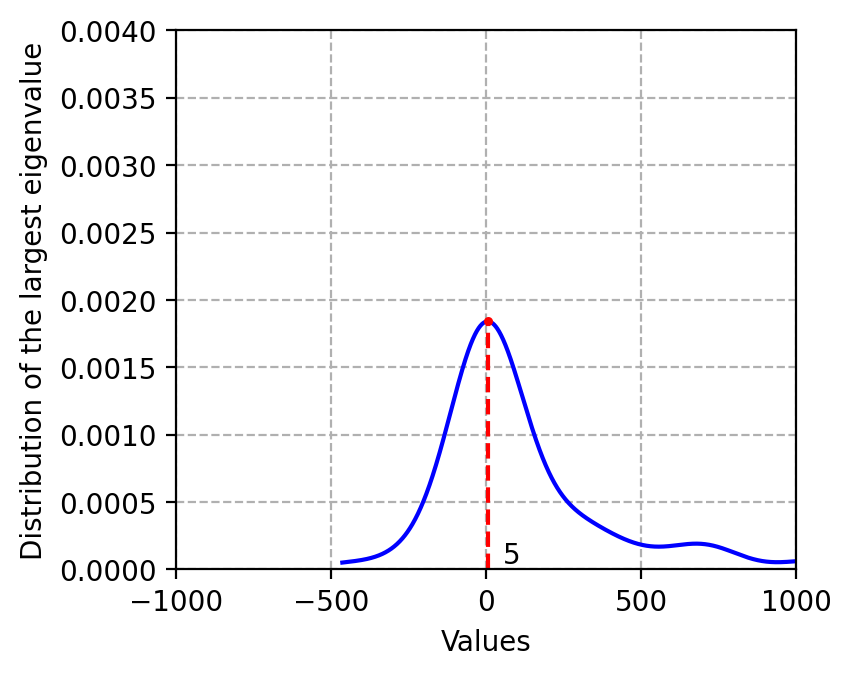}}
    \subfloat[]{\includegraphics[trim=0 0 0 0, clip, scale=0.4]{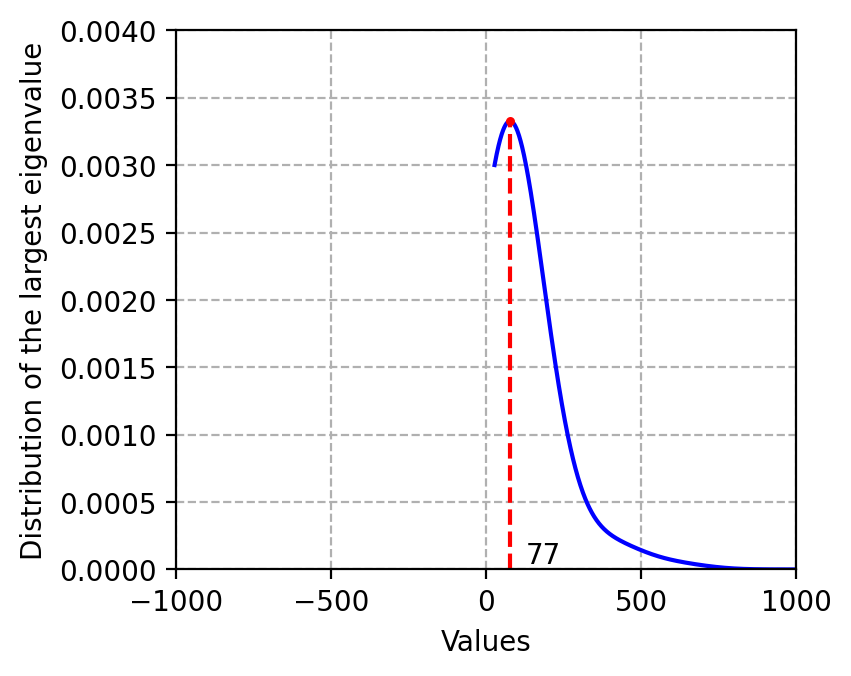}}
    \subfloat[]{\includegraphics[trim=0 0 0 0, clip, scale=0.4]
    {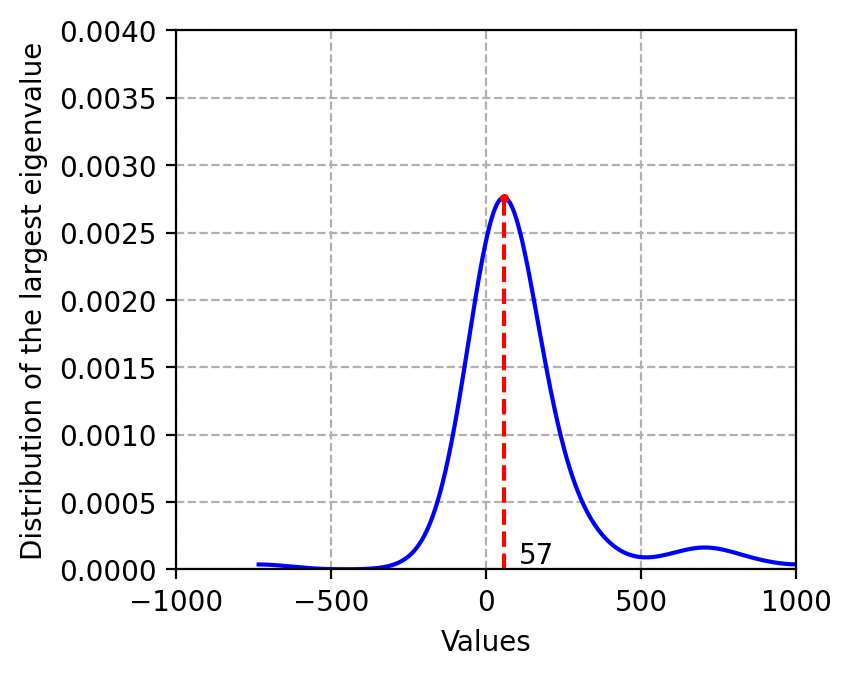}}\par
    \vspace{-0.38cm}
    \subfloat[]{\includegraphics[trim=0 0 0 0, clip, scale=0.4]
    {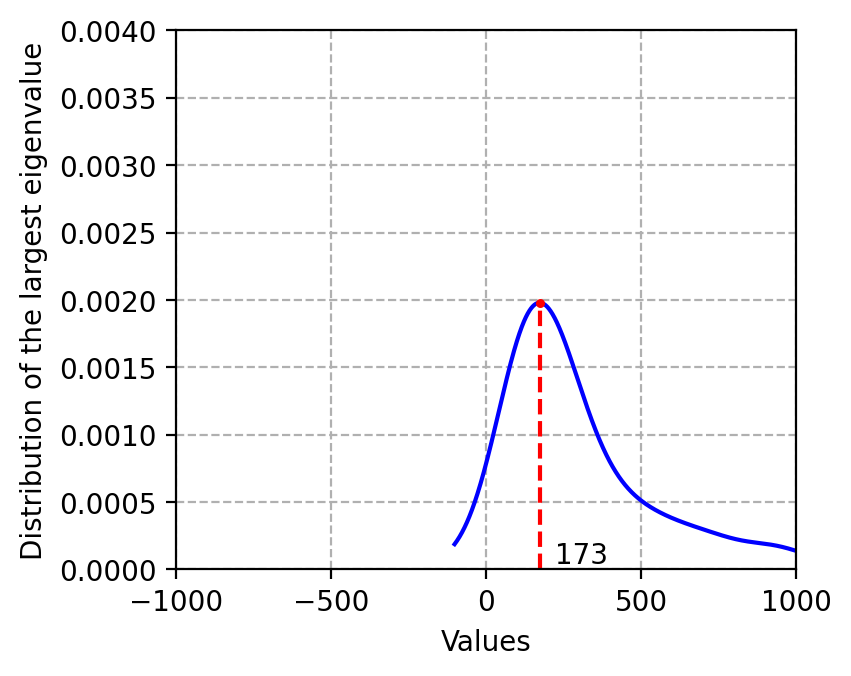}}
    \subfloat[]{\includegraphics[trim=0 0 0 0, clip, scale=0.4]
    {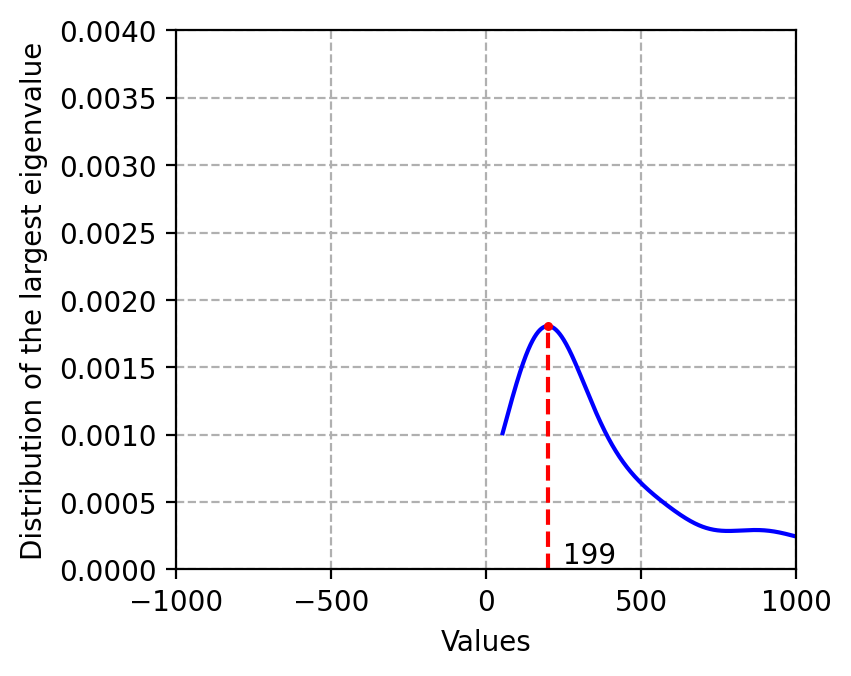}}
    \subfloat[]{\includegraphics[trim=0 0 0 0, clip, scale=0.4]
    {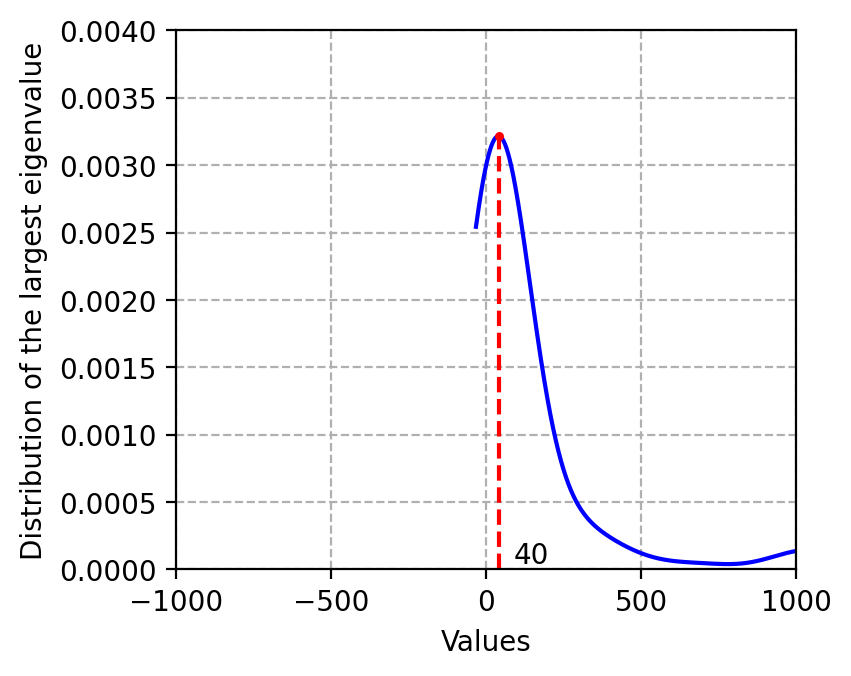}}
    \subfloat[]{\includegraphics[trim=0 0 0 0, clip, scale=0.4]
    {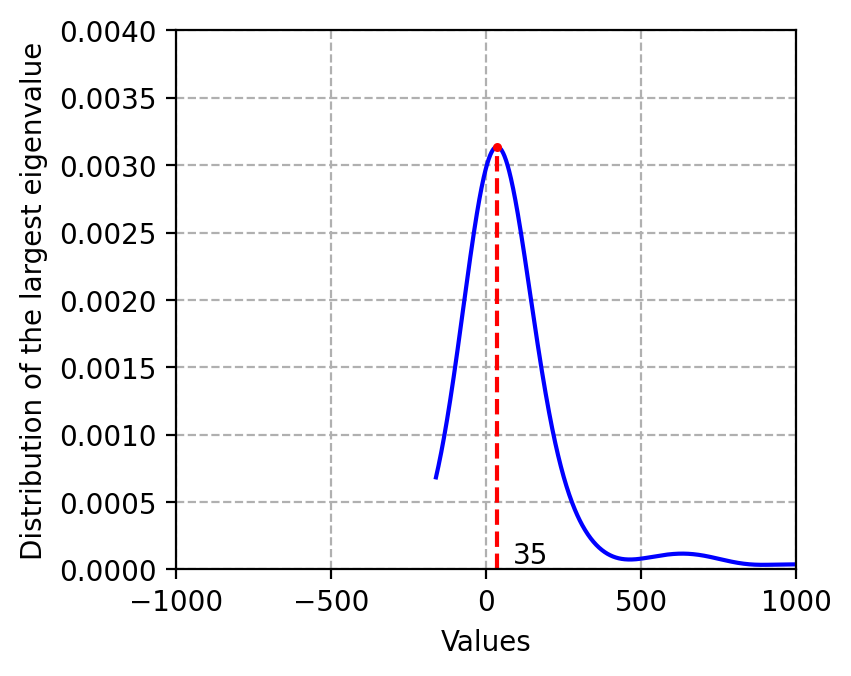}}
    \par 

\caption{Largest eigenvalue of the Hessian based on Houston 2013 dataset. (a) CNN3D. (b) DFFN. (c) M3D-DCNN. (d) RSSAN. (e) SpectralFormer. (f) SSFTT. (g) GroupTransformer. (h) Proposed CNN-mixer. (i) Proposed SSA-mixer. (j) Proposed CSA-mixer. (k) Proposed SSA+CNN-mixer. (l) Proposed CSA+CNN-mixer.}
\label{fig:hu_Hessian}
\end{figure*}

\begin{figure*}
    \centering
    \subfloat[]{\includegraphics[trim=0 0 0 0, clip, scale=0.4]{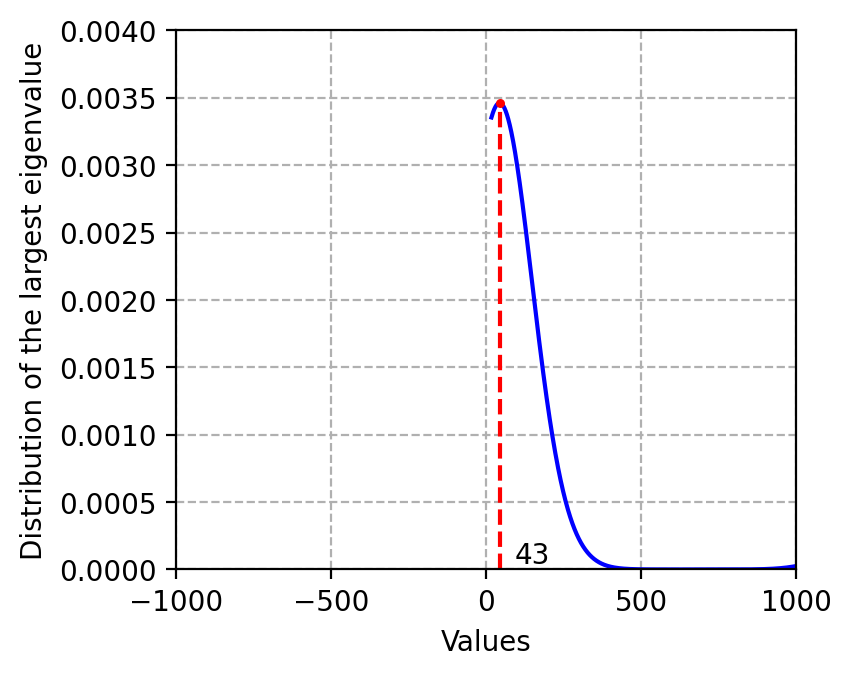}}
    \subfloat[]{\includegraphics[trim=0 0 0 0, clip, scale=0.4]{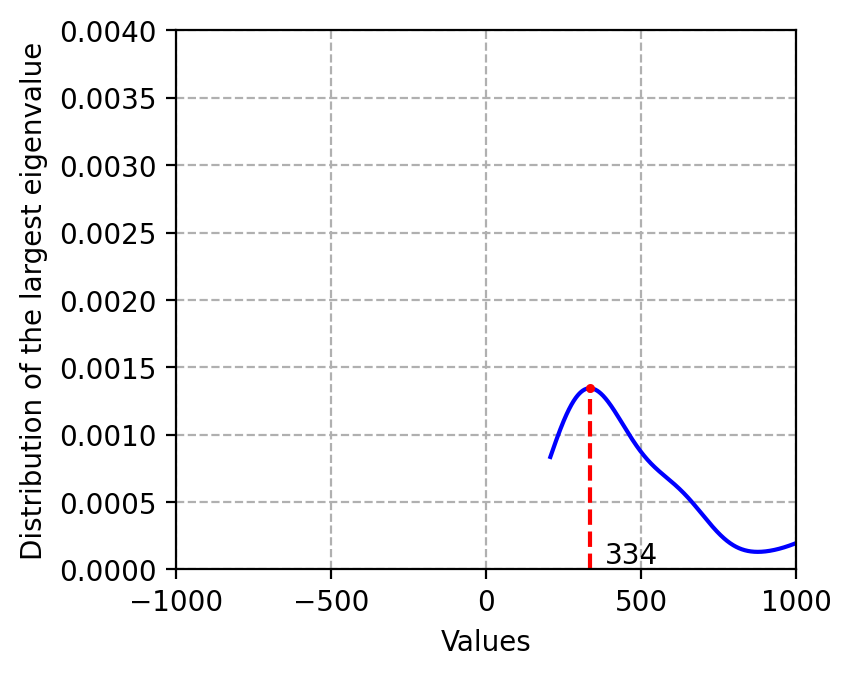}}
    \subfloat[]{\includegraphics[trim=0 0 0 0, clip, scale=0.4]{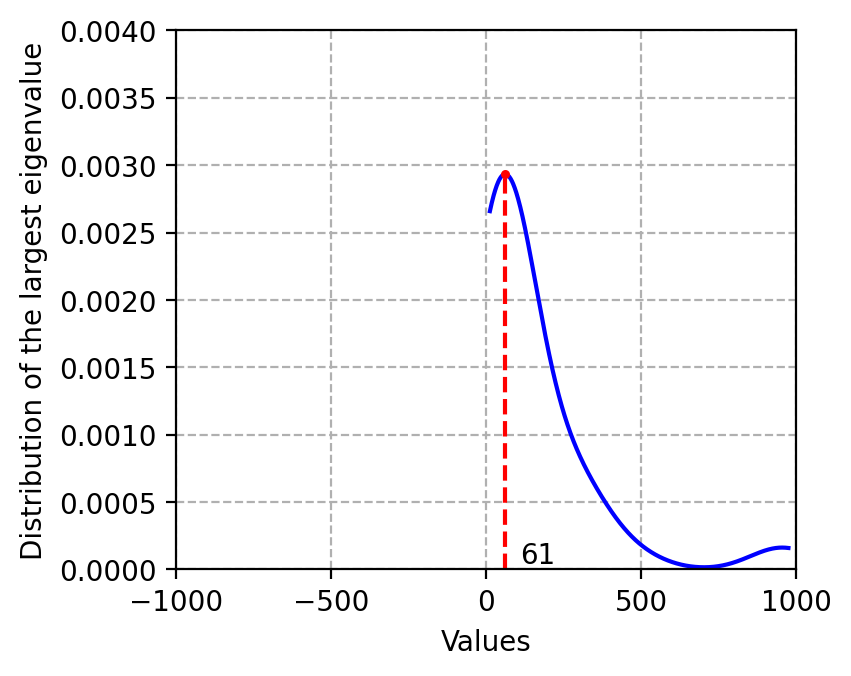}}
    \subfloat[]{\includegraphics[trim=0 0 0 0, clip, scale=0.4]{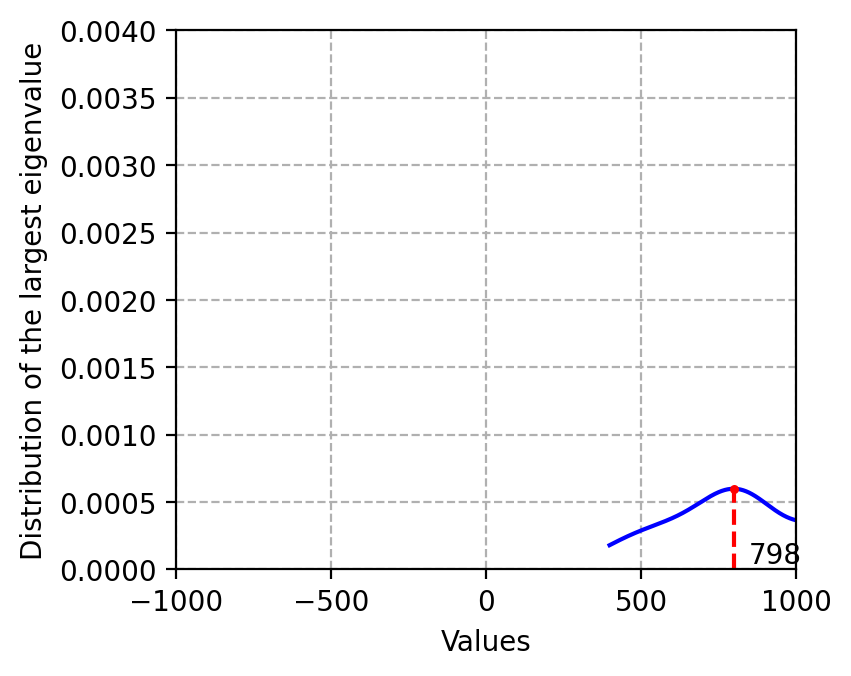}}\par
    \vspace{-0.38cm} 
    \subfloat[]{\includegraphics[trim=0 0 0 0, clip, scale=0.4]{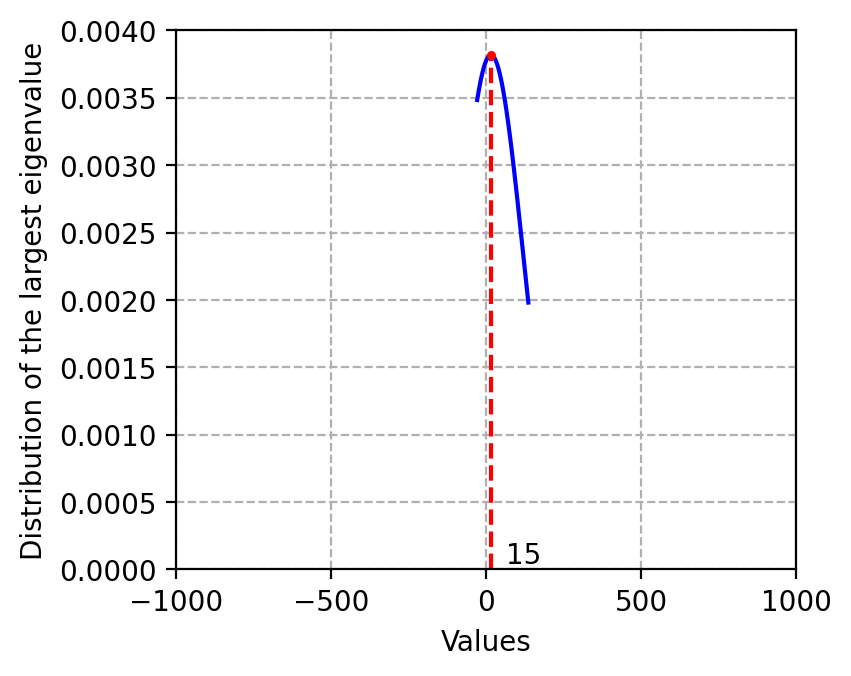}}
    \subfloat[]{\includegraphics[trim=0 0 0 0, clip, scale=0.4]{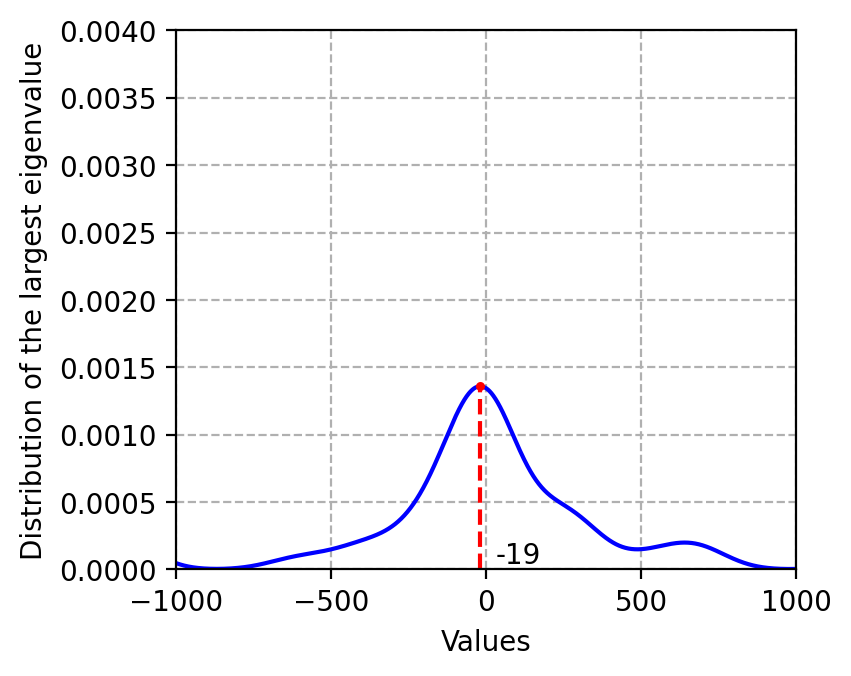}}
    \subfloat[]{\includegraphics[trim=0 0 0 0, clip, scale=0.4]{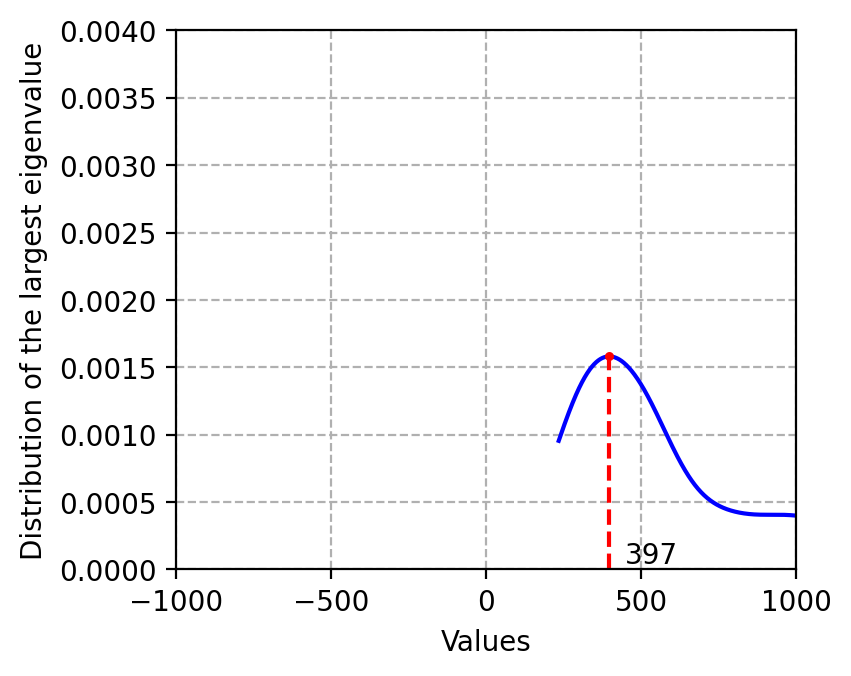}}
    \subfloat[]{\includegraphics[trim=0 0 0 0, clip, scale=0.4]
    {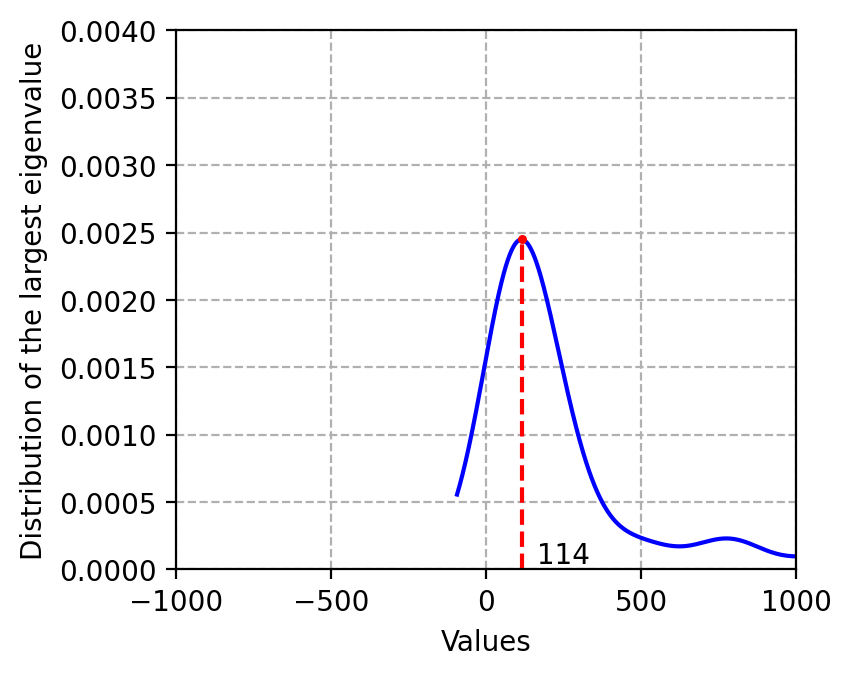}}\par
    \vspace{-0.38cm}  
    \subfloat[]{\includegraphics[trim=0 0 0 0, clip, scale=0.4]
    {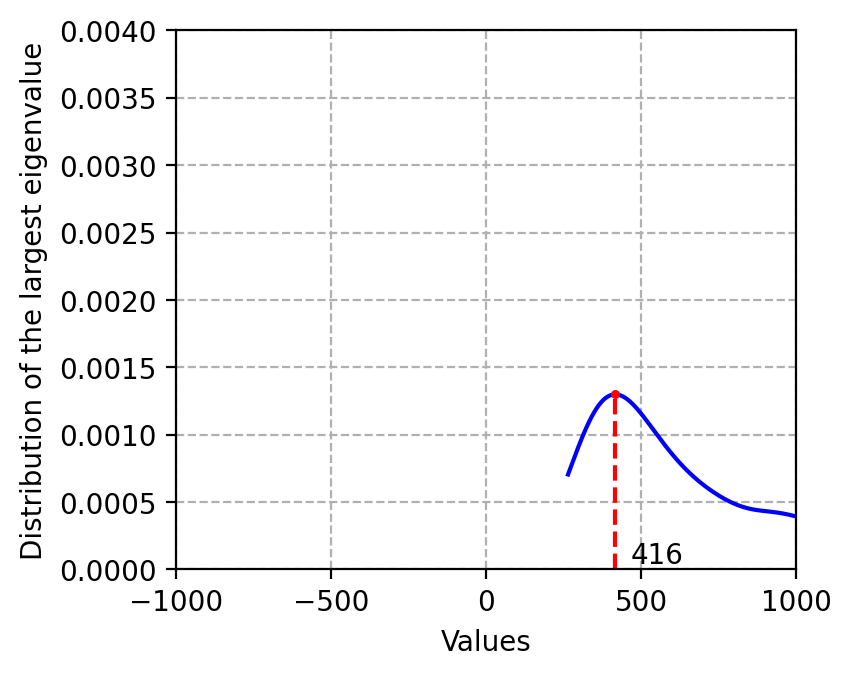}}
    \subfloat[]{\includegraphics[trim=0 0 0 0, clip, scale=0.4]
    {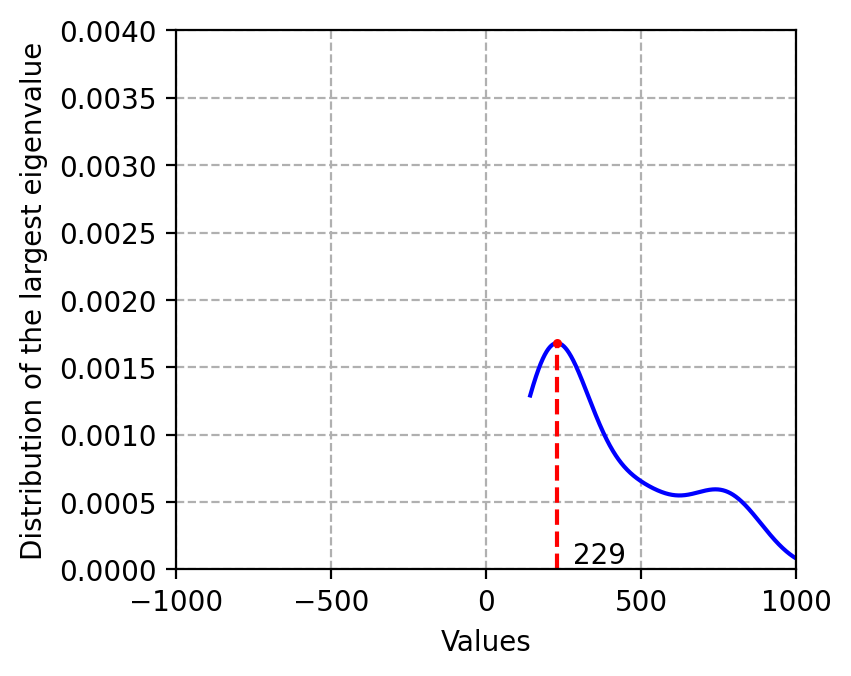}}
    \subfloat[]{\includegraphics[trim=0 0 0 0, clip, scale=0.4]
    {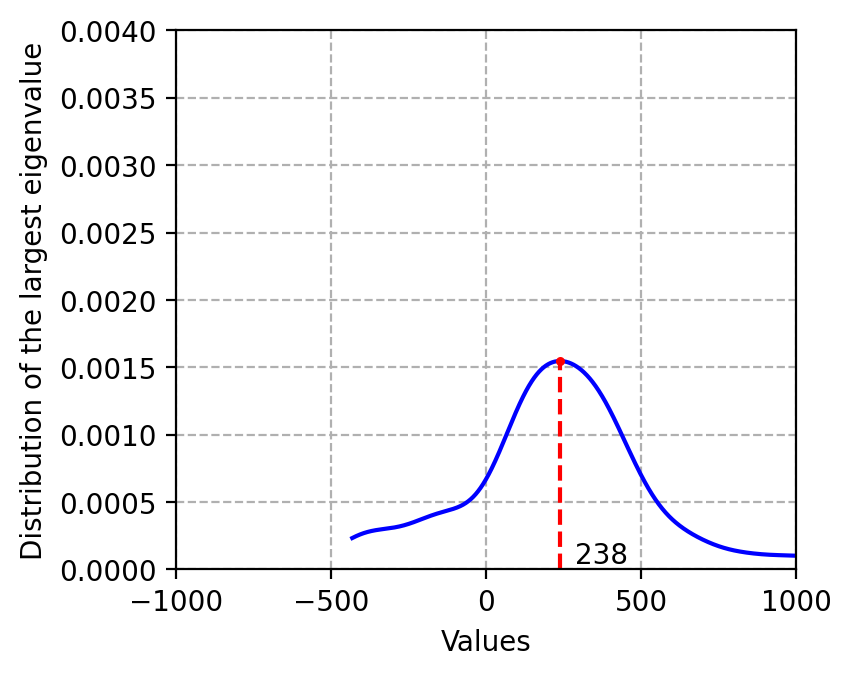}}
    \subfloat[]{\includegraphics[trim=0 0 0 0, clip, scale=0.4]
    {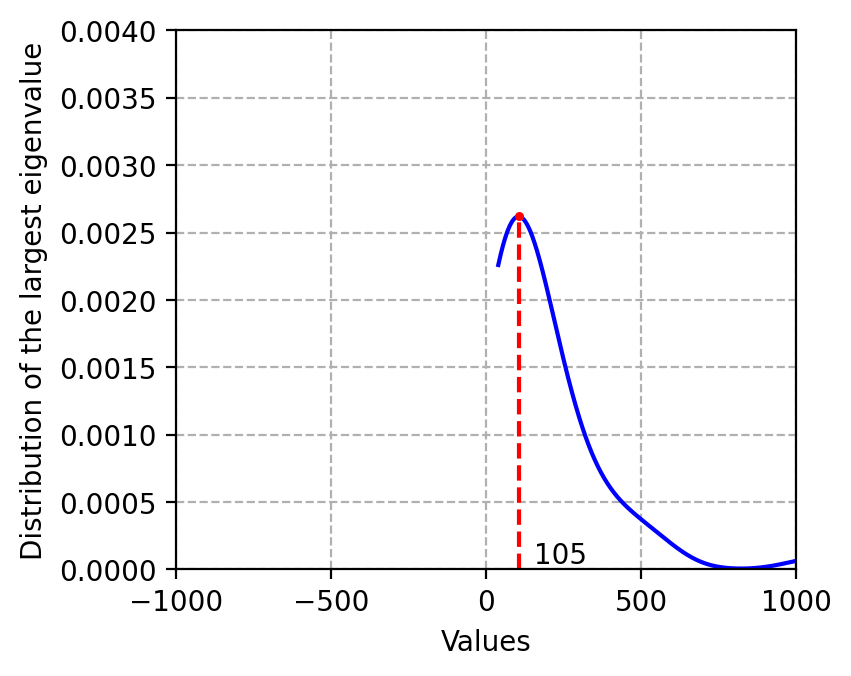}}
    \par 

\caption{Largest eigenvalue of the Hessian based on Botswana dataset. (a) CNN3D. (b) DFFN. (c) M3D-DCNN. (d) RSSAN. (e) SpectralFormer. (f) SSFTT. (g) GroupTransformer. (h) Proposed CNN-mixer. (i) Proposed SSA-mixer. (j) Proposed CSA-mixer. (k) Proposed SSA+CNN-mixer. (l) Proposed CSA+CNN-mixer.}
\label{fig:bot_Hessian}
\end{figure*}

\begin{figure*}
    \centering
    \subfloat[]{\includegraphics[trim=0 0 0 0, clip, scale=0.4]{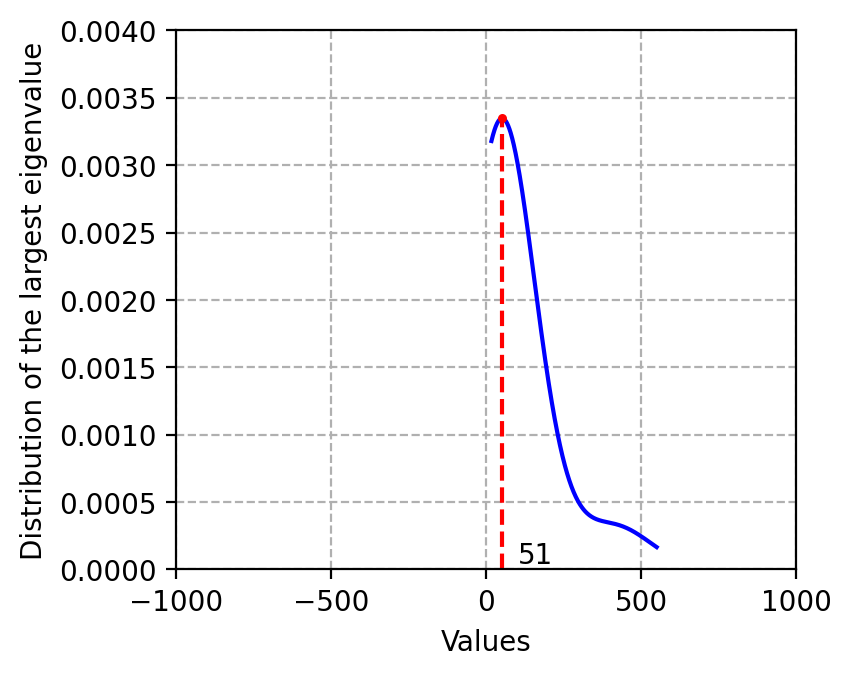}}
    \subfloat[]{\includegraphics[trim=0 0 0 0, clip, scale=0.4]{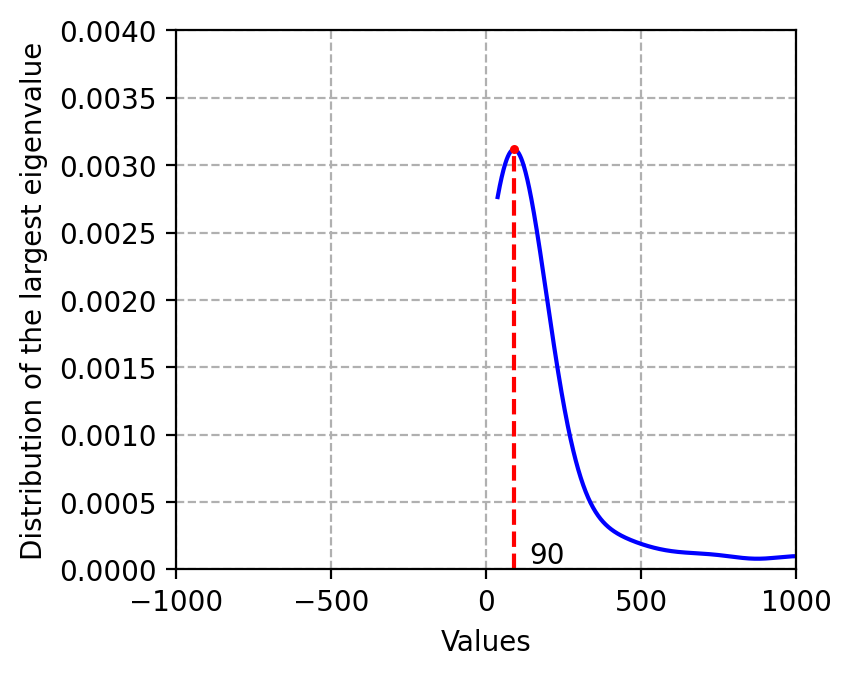}}
    \subfloat[]{\includegraphics[trim=0 0 0 0, clip, scale=0.4]
    {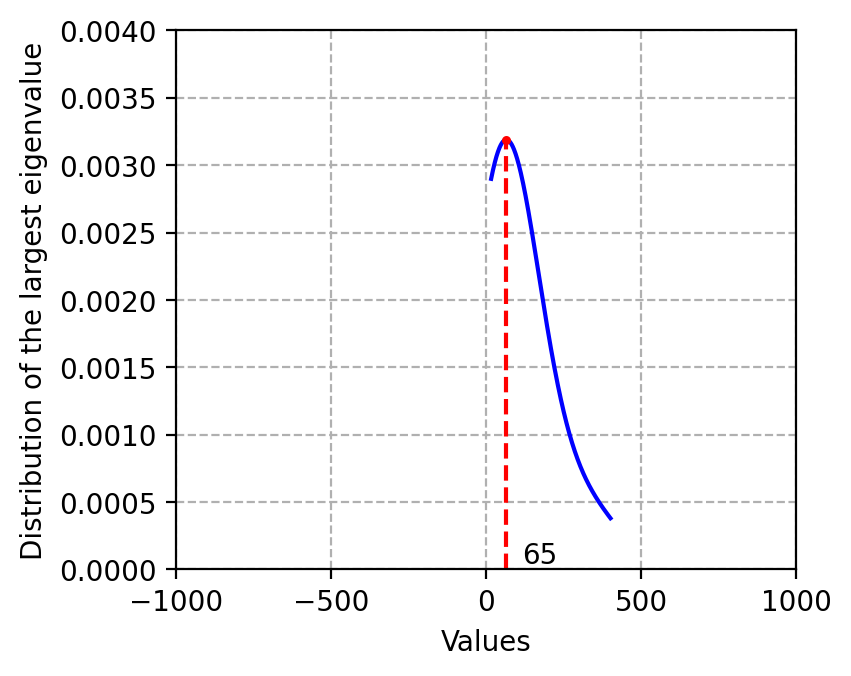}}
    \subfloat[]{\includegraphics[trim=0 0 0 0, clip, scale=0.4]{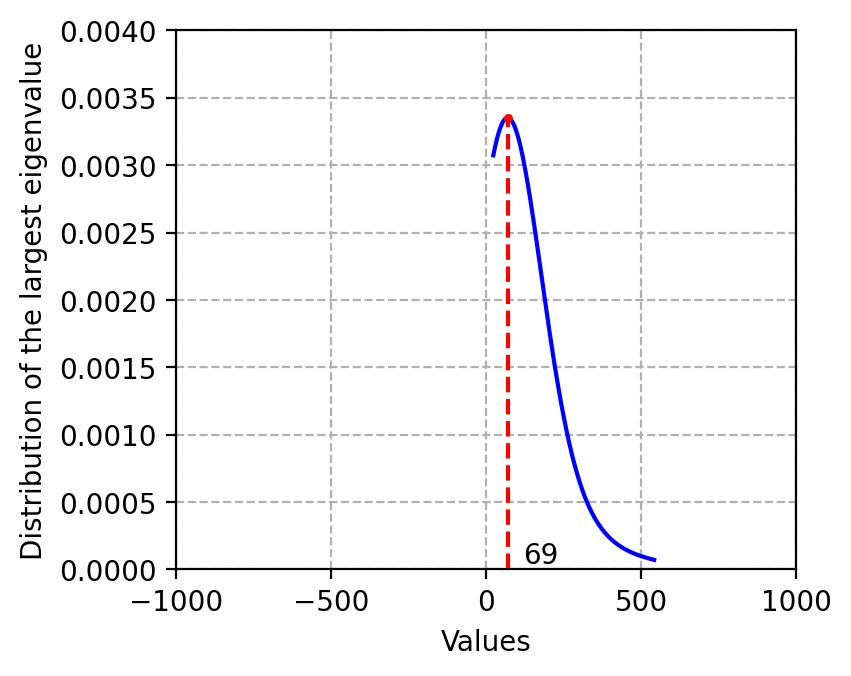}}\par
    \vspace{-0.38cm}  
    \subfloat[]{\includegraphics[trim=0 0 0 0, clip, scale=0.4]{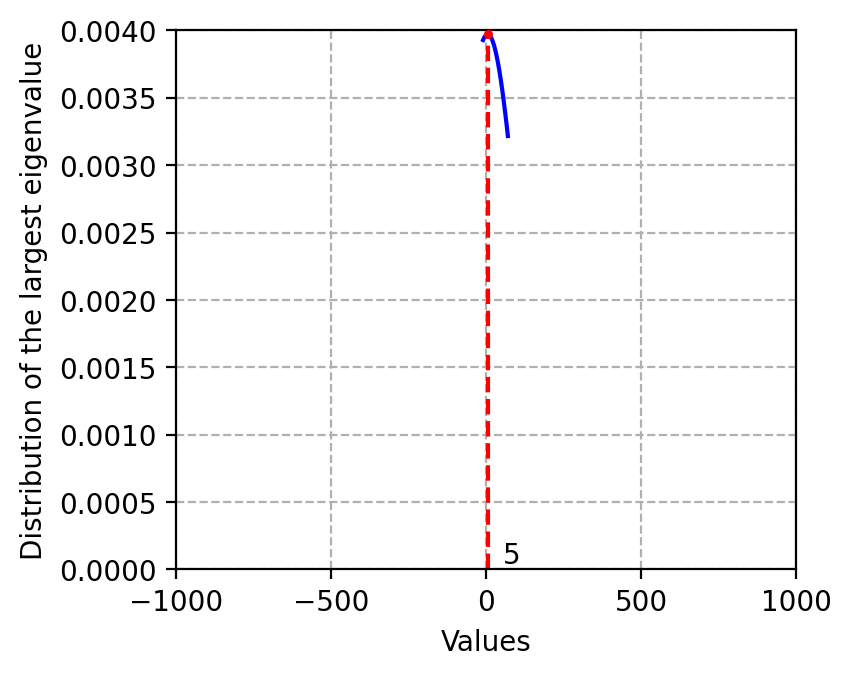}}
    \subfloat[]{\includegraphics[trim=0 0 0 0, clip, scale=0.4]{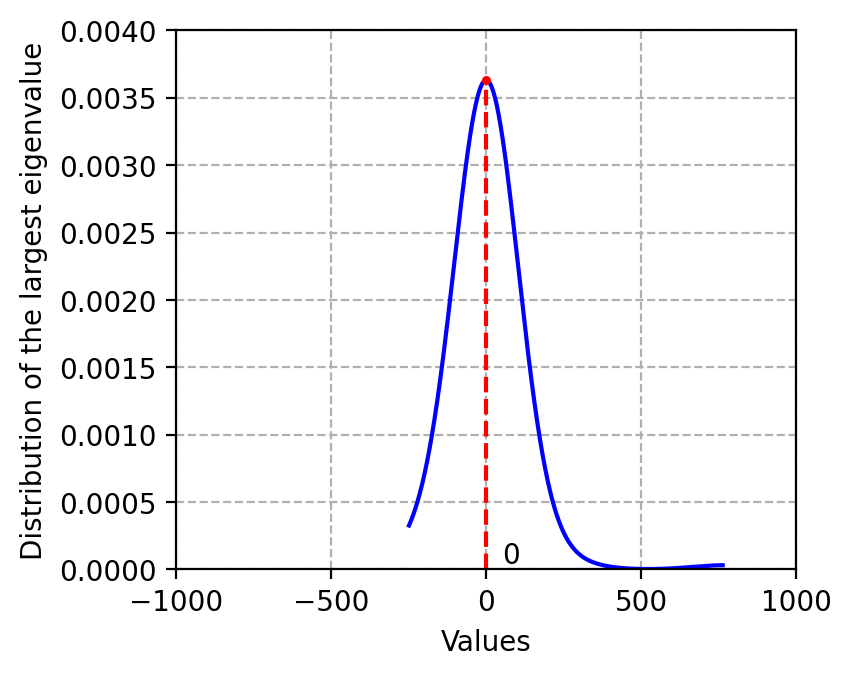}}
    \subfloat[]{\includegraphics[trim=0 0 0 0, clip, scale=0.4]{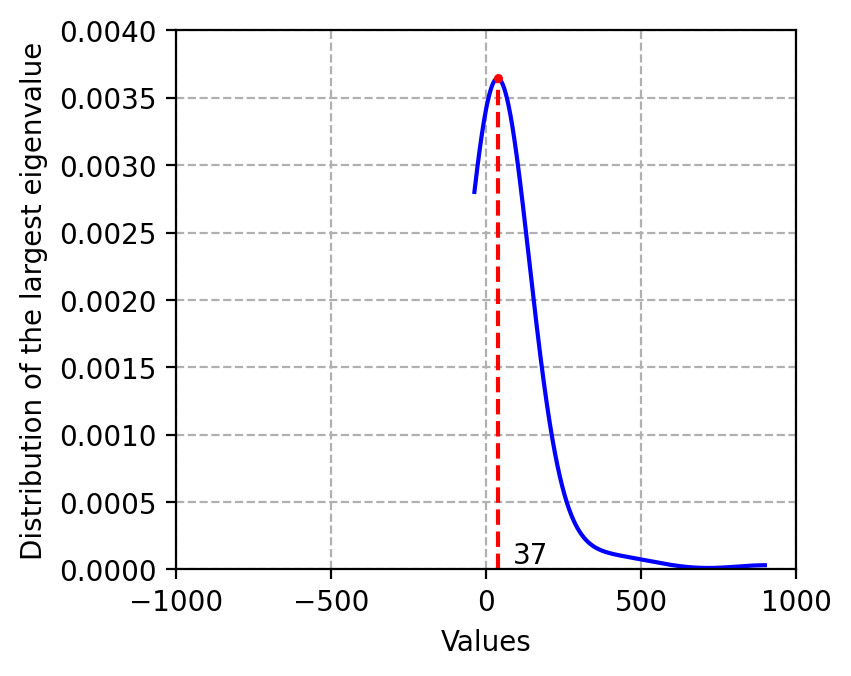}}
    \subfloat[]{\includegraphics[trim=0 0 0 0, clip, scale=0.4]
    {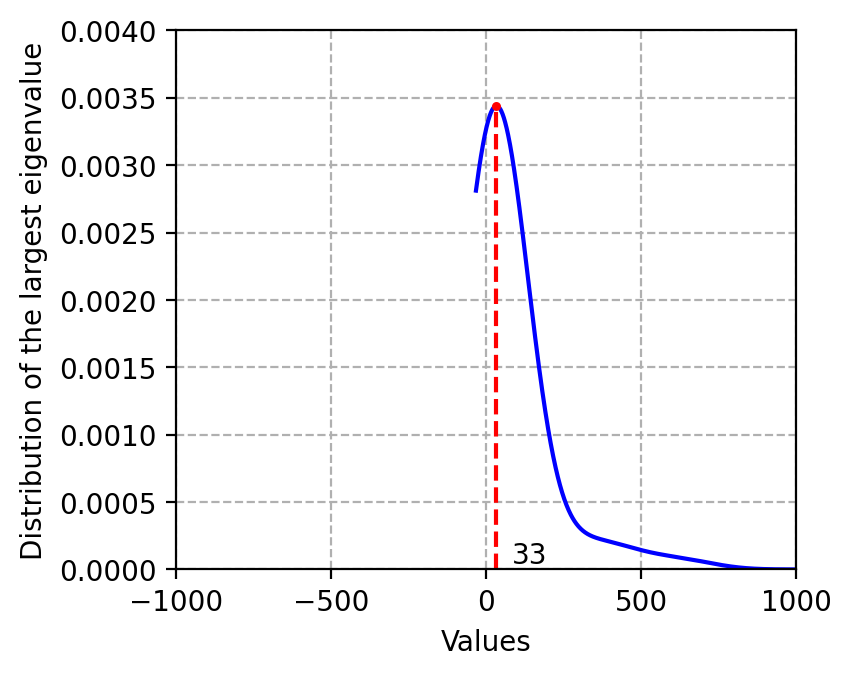}}\par
    \vspace{-0.38cm}  
    \subfloat[]{\includegraphics[trim=0 0 0 0, clip, scale=0.4]
    {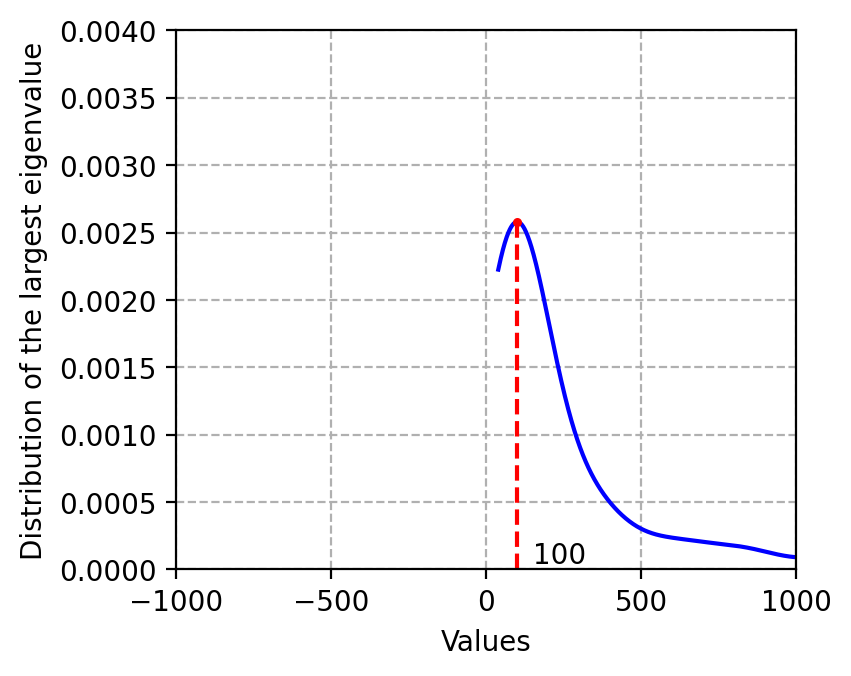}}
    \subfloat[]{\includegraphics[trim=0 0 0 0, clip, scale=0.4]
    {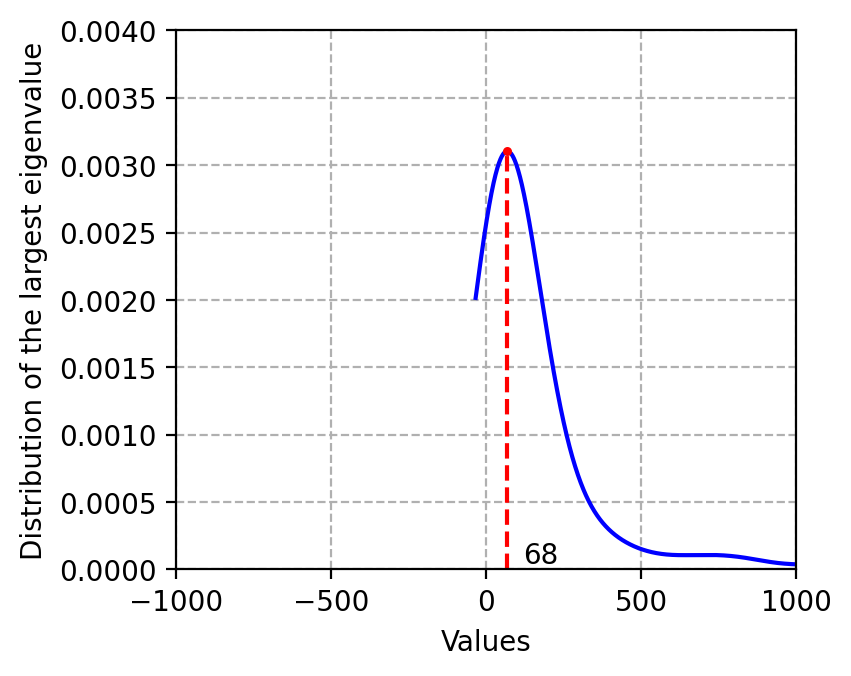}}
    \subfloat[]{\includegraphics[trim=0 0 0 0, clip, scale=0.4]
    {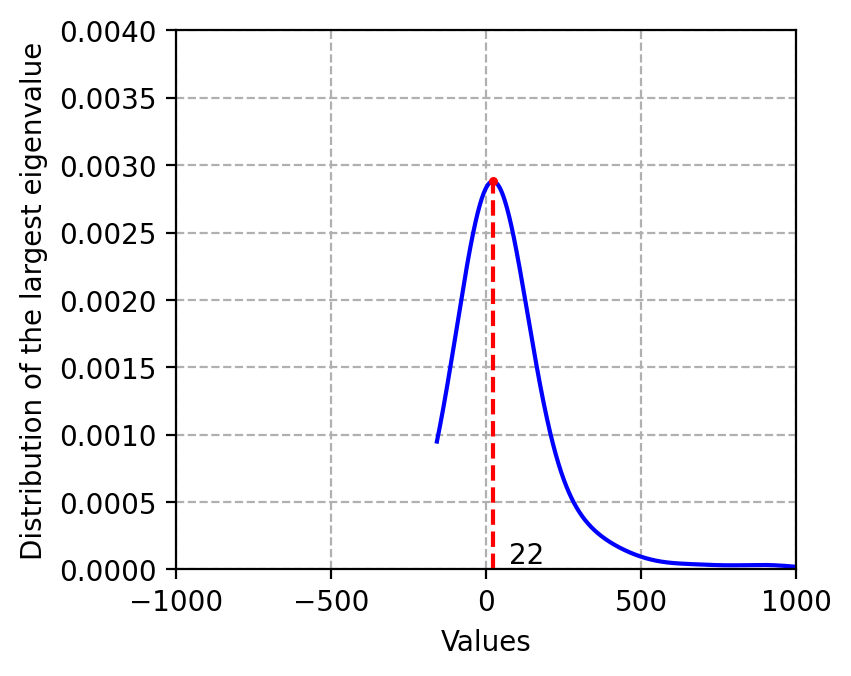}}
    \subfloat[]{\includegraphics[trim=0 0 0 0, clip, scale=0.4]
    {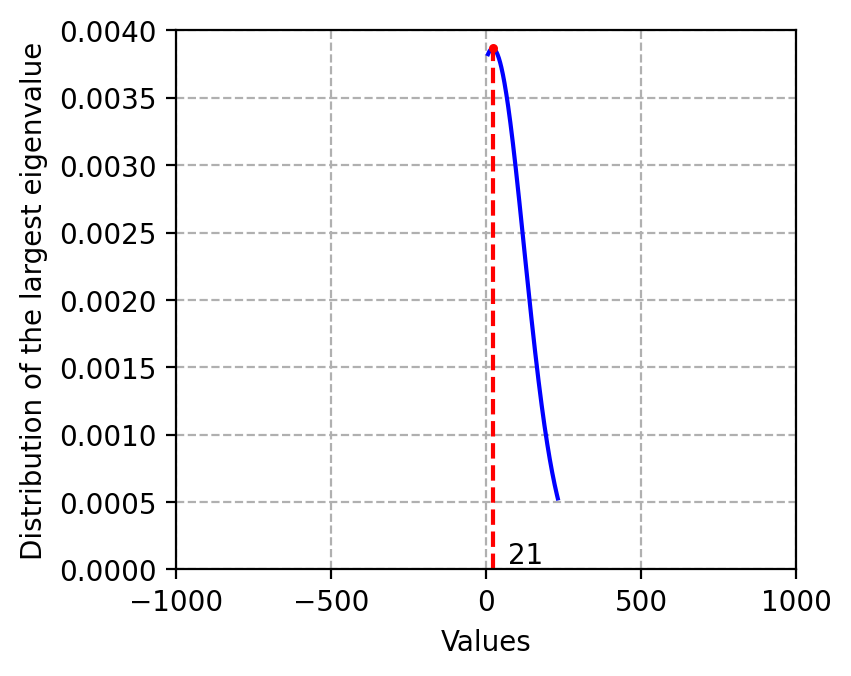}}
    \par 

\caption{Largest eigenvalue of the Hessian based on Pavia dataset. (a) CNN3D. (b) DFFN. (c) M3D-DCNN. (d) RSSAN. (e) SpectralFormer. (f) SSFTT. (g) GroupTransformer. (h) Proposed CNN-mixer. (i) Proposed SSA-mixer. (j) Proposed CSA-mixer. (k) Proposed SSA+CNN-mixer. (l) Proposed CSA+CNN-mixer.}
\label{fig:pu_Hessian}
\end{figure*}

Through the aforementioned loss surface contours, the overall global features of different models after disturbances can be intuitively visualized. However, these contours do not offer a microscopic analysis of the model's local characteristics in the neighborhood of the 'best' pretrained weight. To address this gap, this paper introduces the distribution of the maximum eigenvalue of the Hessian, aiming to quantitatively analyze the model's gradient properties from a local perspective in the vicinity of the 'best' pretrained weight. Fig. \ref{fig:hu_Hessian}, \ref{fig:bot_Hessian}, and \ref{fig:pu_Hessian} depict the distribution of the maximum eigenvalue of the Hessian for different models across the three datasets. Among the four typical CNN models, CNN3D and M3D-DCNN exhibit similar curves for the maximum eigenvalue of the Hessian. The magnitude of the maximum eigenvalue of the Hessian approaches zero, and negative values are virtually absent. In contrast, for DFFN, the magnitude of the maximum eigenvalue of the Hessian is relatively larger, especially on the Houston 2013 and Botswana datasets. Consequently, in general, CNN3D and M3D-DCNN, exhibit smoother local behavior around the 'best' pretrained weights among the four classical CNN-based models. Interestingly, on the Pavia University dataset, the magnitude of the maximum eigenvalue of the Hessian for all four CNN algorithms is similar, and none of them exhibit negative values. This implies that all four CNN algorithms demonstrate remarkably smooth local behavior around the optimal points as they approach the end of model training. (e), (f), and (g) represent three classical vision Transformer models. Among these models, it can be observed that SpectralFormer's distribution of the maximum eigenvalue of the Hessian approaches the $x = 0$ axis. This indicates that the model exhibits a highly flat behavior in the vicinity of optimal weights, resulting in minimal responsiveness to local perturbations. Conversely, for SSFTT, the maximum eigenvalue of the Hessian is partly situated on the $x < 0$ side across all three datasets. This implies non-convexity in the model's local behavior near this point, potentially hindering its ability to search for optimal weights during training. Within the GroupTransformer algorithm, the magnitude of the maximum eigenvalue of the Hessian is notably higher on the Botswana dataset compared to Houston 2013 and Pavia University. This leads to a sharper high-dimensional loss surface near the optimal point. This discrepancy may be attributed to the relatively smaller number of training samples per class in Botswana, indicating a need for improved model generalization. Upon comparing the five vision Transformer models proposed, it is evident that the horizontal coordinate of the peak value in the maximum eigenvalue distribution of the Hessian consistently exceeds 0. This signifies their capacity to maintain convexity in the vicinity of optimal points. Moreover, when compared to the other three pure (CNN-mixer, SSA-mixer, and CSA-mixer) models, the magnitude of the Hessian eigenvalue for the two hybrid-mixer (SSA+CNN-mixer and CSA+CNN-mixer) models approaches 0, indicating that hybrid-mixer models tend to exhibit smoother behavior near the optimal point. However, as shown in Fig. \ref{fig:hu_disturbance_robustness}, \ref{fig:bot_disturbance_robustness}, and \ref{fig:pu_disturbance_robustness}, the smoothness of loss values after disturbance is already relatively high. Consequently, while hybrid-mixer models can further enhance local smoothness, they do not wield a decisive influence on optimizing the entire model, given the closely matched performance of all five models across the three datasets.

\begin{figure*}
    \centering
    \subfloat[]{\includegraphics[scale=0.6]{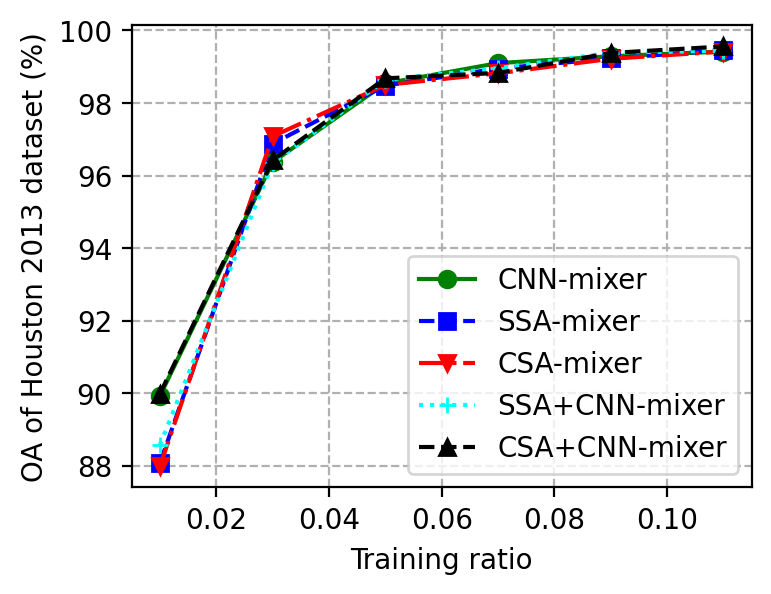}}
    \subfloat[]{\includegraphics[scale=0.6]{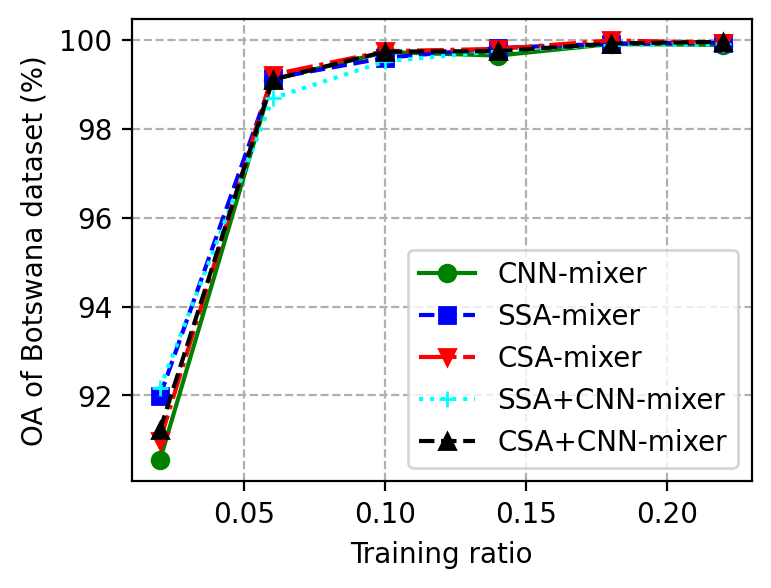}}
    \subfloat[]{\includegraphics[scale=0.6]{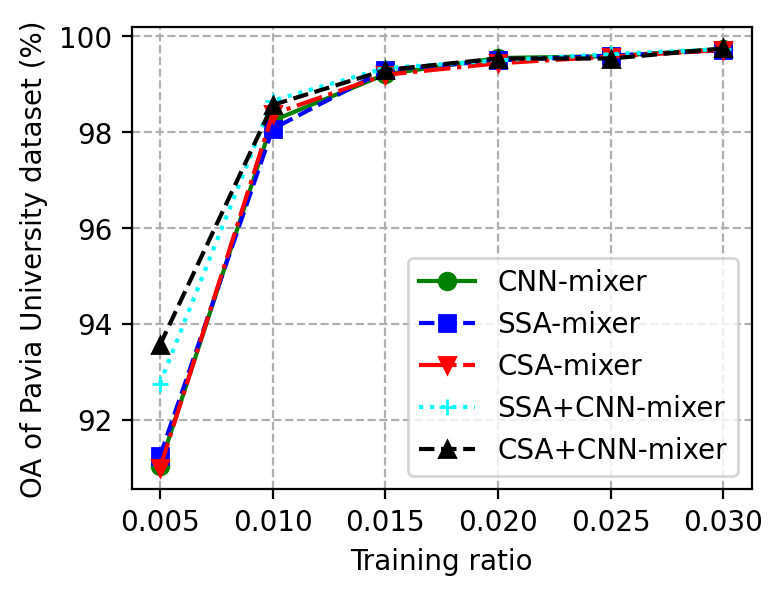}}

\caption{Training ratio effect on the overall accuracy. (a) Houston 2013 dataset. (b) Botswana dataset. (c) Pavia University dataset.}
\label{fig:ratio_effect}
\end{figure*}

3) Impact of training ratio on OA: To investigate the impact of the number of training samples on the overall accuracy of HSI classification, experiments were conducted using different numbers of training samples in the three datasets with the proposed five approaches based on the unified hierarchical vision Transformer architecture. For the Houston 2013 dataset, proportions of 1\%, 3\%, 5\%, 7\%, 9\%, and 11\% of the data were selected for training using a stratified sampling strategy. As shown in Fig.~\ref{fig:ratio_effect}(a), when only 1\% of the sample points were used as the training dataset, the OA ranges from 87.98\% to 89.98\%. At this stage, the model based on the CSA-mixer exhibits an OA of 87.98\%, while the model using CSA+CNN-mixer achieves an OA of 89.98\%, the highest accuracy among the models. As the number of training samples increases, the OA accuracy of all the five mixer models exhibits an increasing trend. When the training sample proportion reaches 11\%, the model accuracy approaches saturation. At this point, the CSA+CNN mixer model reaches a notable accuracy of 99.56\%, demonstrating a slight enhancement in performance with an accuracy improvement of less than 0.14\% relative to four other models. For the Botswana dataset, a stratified sampling strategy was also employed to select training data in proportions of 2\%, 6\%, 10\%, 14\%, 18\%, and 22\%. As illustrated in Fig.~\ref{fig:ratio_effect}(b), when only 2\% of the data was utilized as training samples, the CNN-mixer-based model records an OA of 90.54\%. It is at least 0.44\% less than the accuracies achieved by the other four algorithms. Additionally, as the number of training samples grows, all the five models initially exhibit significant improvements in OA accuracy, which gradually plateau as they approach saturation. When the training sample proportion reaches 22\% of the total dataset, the accuracy of the five models ranges from 99.88\% to 99.95\%, with an OA accuracy fluctuation of no more than 0.07\% among them. For the Pavia University dataset, only 0.5\%, 1\%, 1.5\%, 2\%, 2.5\%, and 3\% of the data were selected as training samples. This is because the number of sample points in this dataset is significantly greater than that of the previous two datasets. As shown in Fig.~\ref{fig:ratio_effect}(c), the OA accuracy curves for the five mixer algorithms exhibit similar trends to those observed in the previous two datasets. For example, when the training samples increase from 0.5\% to 3\%, the OA accuracy improves dramatically from 90.99\% to 99.75\%, showing an impressive growth of 8.76\%. With 2\% of the data used as the training set, the OA accuracy of the five mixer models ranges from 99.44\% to 99.55\%. Overall, under the constraint of limited annotated samples, the quantity of training samples has a particularly noticeable impact on the accuracy of HSI classification. Furthermore, based on the proposed unified hierarchical vision Transformer architecture, different HSI classification models constructed with various mixers exhibit comparable performance across different datasets. This provides additional empirical support that for HSI vision Transformer classification algorithms, performance primarily relies on the unified hierarchical vision Transformer architecture rather than specific MSA or other mixer modules, especially under conditions where the proportion of training data is sufficiently substantial.

\section{Conclusions} \label{conlucison}
A novel unified hierarchical vision Transformer architecture is developed for HSI classification. Five different vision Transformer models are constructed by configuring different mixers within the proposed unified architecture. Experiments on three commonly analyzed hyperspectral benchmark data sets with different characteristics reveal that the proposed methods outperform traditional CNN-based or vision Transformer-based HSI classification methods. Furthermore, an in-depth analysis conducted from two perspectives, disturbance robustness and the distribution of the maximum eigenvalue of the Hessian, implies that the effectiveness of vision Transformer-based HSI classification models primarily depends on the holistic unified architecture, rather than the commonly presumed MSA module. This paper provides insights into the design of vision Transformer-based neural networks for future research in HSI classification. Further work is warranted to incorporate self-supervised learning and analyze the frequency characteristics of feature space extraction in various mixer modules within the vision Transformer architecture through a self-supervised pre-training paradigm.

\section{Acknowledgments}
This work was supported by the NASA grant $\# 80NSSC22K1163$.

\clearpage
\twocolumn  % 
\bibliographystyle{IEEEtran}
% \bibliography{Refs}

% \end{IEEEbiography}
\end{document}